\documentclass[10pt,twocolumn,letterpaper]{article}

\usepackage{cvpr}
\usepackage{times}
\usepackage{epsfig}
\usepackage{graphicx}
\usepackage{subcaption}
\usepackage{amsmath}
\usepackage{amssymb}
\usepackage{lscape}
\usepackage{xspace}

\usepackage[pagebackref=true,breaklinks=true,letterpaper=true,colorlinks,bookmarks=false]{hyperref}

\cvprfinalcopy
\def\cvprPaperID{2327} 

\ifcvprfinal\fi
\begin{document}

\makeatletter
\renewcommand{\paragraph}{%
  \@startsection{paragraph}{4}%
  {\z@}{0.5ex \@plus 0.5ex \@minus .2ex}{-1em}%
  {\normalfont\normalsize\bfseries}%
}
\makeatother

\newcommand{\GW}{GuessWhat?!\xspace}
\newcommand{\Referit}{ReferIt\xspace}
\newcommand{\oracle}{oracle\xspace}
\newcommand{\Oracle}{Oracle\xspace}
\newcommand{\questioner}{questioner\xspace}
\newcommand{\Questioner}{Questioner\xspace}
\title{\GW Visual object discovery through multi-modal dialogue}


\author{Harm de Vries\\
University of Montreal\\
{\tt\small mail@harmdevries.com}
\and
Florian Strub\\
Univ. Lille, CNRS, Centrale Lille,\\
Inria, UMR 9189 CRIStAL\\
{\tt\small florian.strub@inria.fr}
\and
Sarath Chandar\\
University of Montreal\\
{\tt\small sarathcse2008@gmail.com}
\and
Olivier Pietquin\\
DeepMind\\   
{\tt\small pietquin@google.com}
\and
Hugo Larochelle\\
Twitter\\
{\tt\small hlarochelle@twitter.com}
\and
Aaron Courville\\
University of Montreal\\
{\tt\small aaron.courville@gmail.com}
}

\maketitle

\begin{abstract}
We introduce \GW, a two-player guessing game as a testbed for research on the interplay of computer vision and dialogue systems. The goal of the game is to locate an unknown object in a rich image scene by asking a sequence of questions. 
Higher-level image understanding, like spatial reasoning and language grounding, is required to solve the proposed task. Our key contribution is the collection of a large-scale dataset consisting of 150K human-played games with a total of 800K visual question-answer pairs on 66K images. We explain our design decisions in collecting the dataset and introduce the oracle and questioner tasks that are associated with the two players of the game. We prototyped deep learning models to establish initial baselines of the introduced tasks. 
\end{abstract}

\section{Introduction}
People use natural language as the most effective way to communicate, including when it comes to describe the visual world around them. They often need only a few words to refer to a specific object in a rich scene. Whenever such expressions \emph{unambiguously} point to one object, we speak of a referring expression~\cite{krahmer2012computational}. However, uniquely identifying the referred object is not always possible, as it depends on the listener's state of mind and the context of the scene. Many real life situations, therefore, require multiple exchanges before it is clear what object is referred to:
\begin{quote}
- Did you see that dog?\\
* You mean the one in the corner?\\
- No, the one that's running.\\
* Yes, what's up with that?
\end{quote}

\vspace{-0.3em}
\begin{figure}[t]
\begin{center}
\centering
\includegraphics[width=0.98\linewidth]{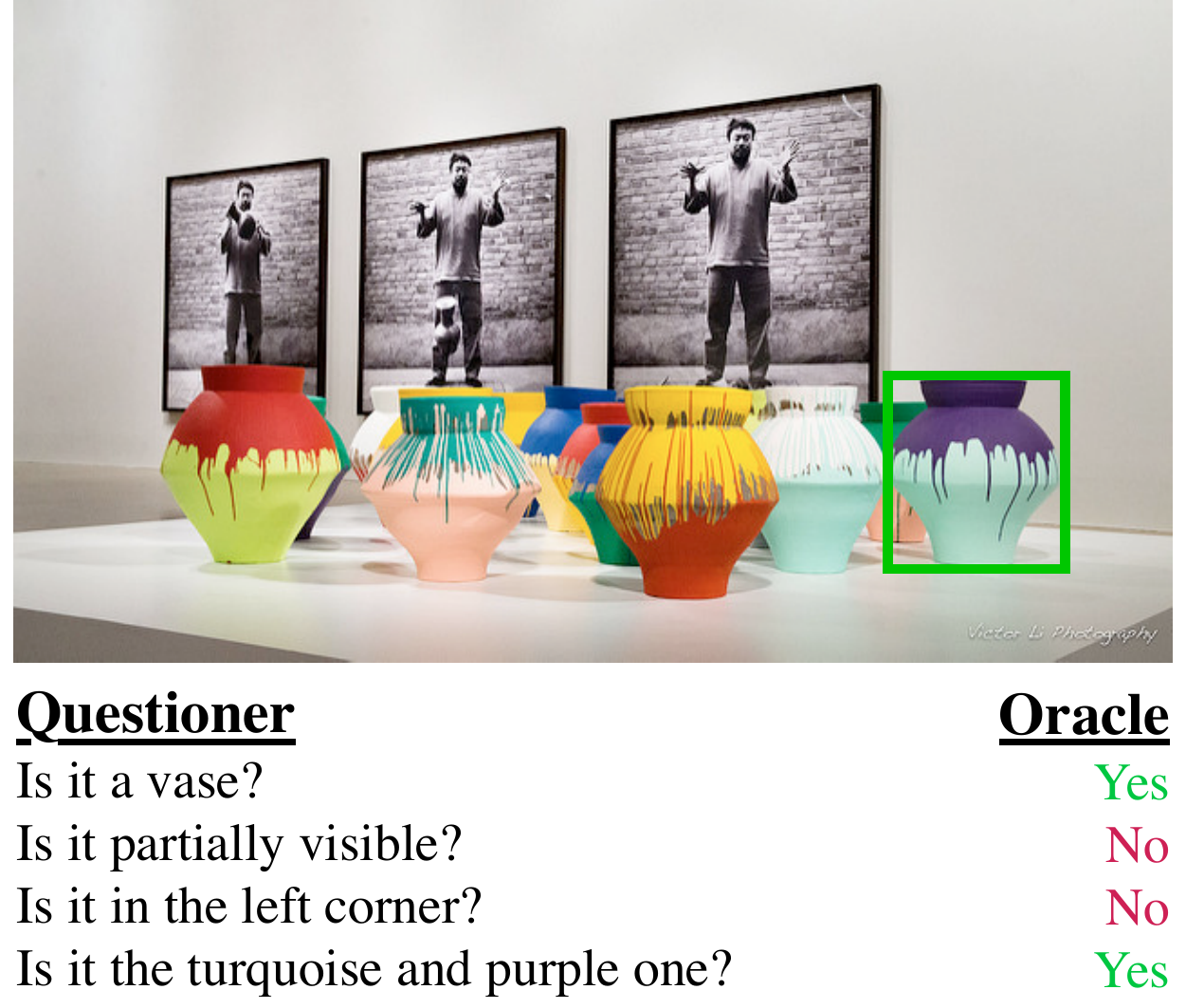}
\end{center}
\vskip -1em
\caption{An example game. After a sequence of four questions, it becomes possible to locate the object (highlighted by a green bounding box). }
\label{fig:front}
\vskip -0.7em
\end{figure}
A computer vision system able to hold conversations about what it \textit{sees} would be an important step towards intelligent scene understanding. Such systems would be more transparent and interpretable because humans may naturally interact with them, for example by asking clarifying questions about what it perceives. Still, a fundamental challenge remains: how to create models that understand natural language descriptions and ground them in the visual world. 

The last few years has seen an increasing interest from the computer vision community in tasks towards this goal. Thanks to advances in training deep neural networks~\cite{Goodfellow-et-al-2016-Book} and the availability of large-scale classification datasets~\cite{lin2014microsoft,ILSVRC15,zhou2014learning}, automatic object recognition has now reached human-level performance~\cite{lecun2015deep}. As a result,  attention has been shifted toward tasks involving higher-level image understanding. One prominent example is image captioning~\cite{lin2014microsoft}, the task of automatically producing natural language descriptions of an image. Visual Question Answering (VQA)~\cite{antol2015vqa} is another popular task that involves answering single open-ended questions concerning an image. Closer to our work, the \Referit game~\cite{kazemzadeh2014referitgame} aims to generate a single expression that refers to one object in the image. 

\begin{figure}[t]
\centering
\includegraphics[width=1\linewidth]{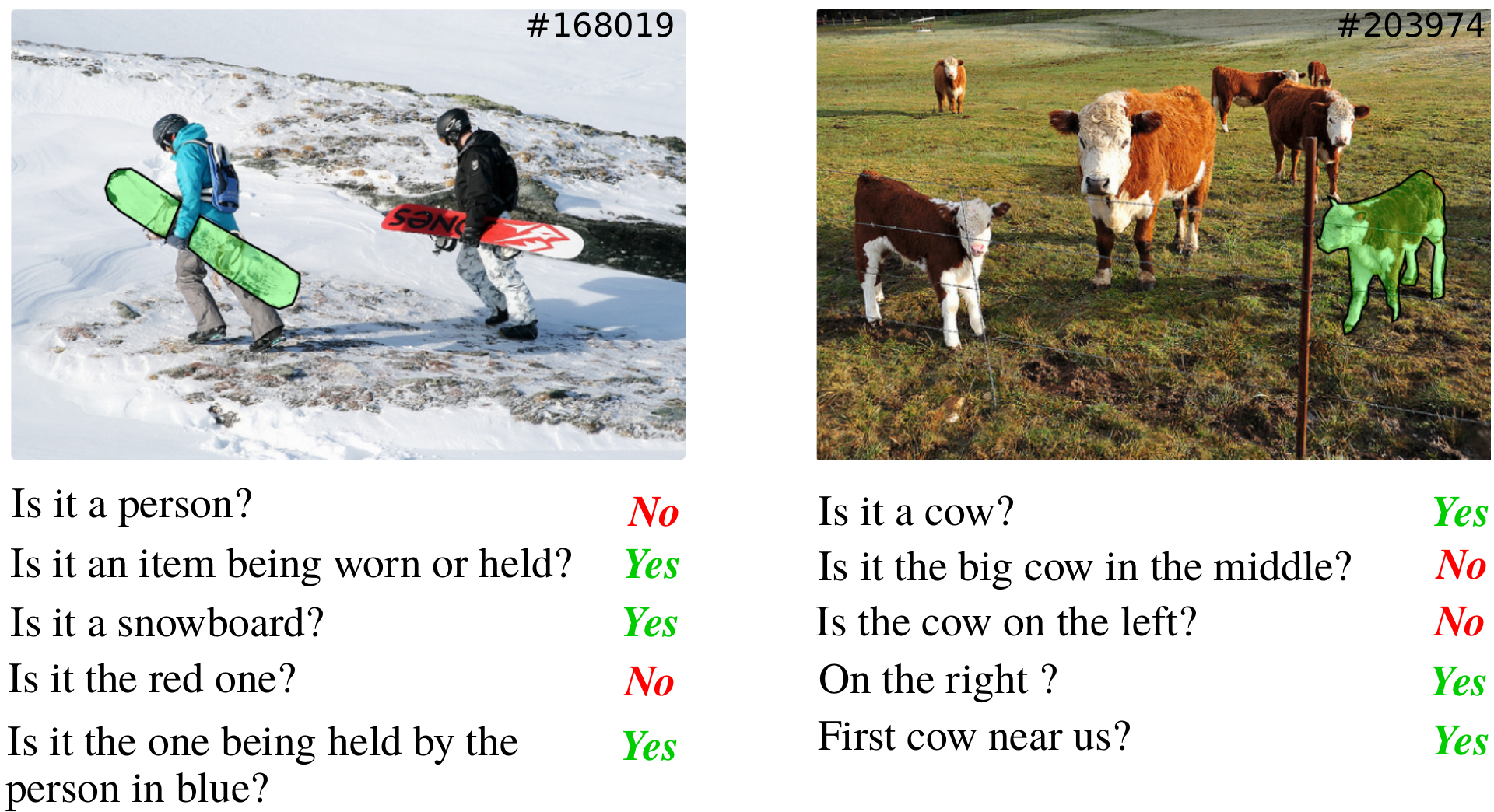}
\caption{Two example games in the dataset. After a sequence of five questions we are able to locate the object (highlighted by a green mask).}
\label{fig:examples}
\vskip -1em
\end{figure}

On the other hand, there has been a renewed interest in dialogue systems~\cite{lemon2012conversational,serban2015survey}, inspired by the success of data-driven approaches in other areas of natural language processing~\cite{cho2014learning}. Traditionally, dialogue systems have been built through heavy engineering and hand-crafted expert knowledge, despite machine learning attempts for almost two decades~\cite{levin1997stochastic,singh1999reinforcement}.
One of the difficulties comes from the lack of automatic evaluation as -- contrary to machine translation -- there is no evaluation metric that correlates well with human evaluation~\cite{liu2016not}. A promising alternative is goal-directed dialogue tasks~\cite{lemon2012conversational,singh1999reinforcement,weston2015towards,wen2016network} where agents converse to pursue a goal rather than casually chit-chat. The agent's success rate in completing the task can then be used as an automatic evaluation metric. Many tasks have recently been introduced, including the bAbI tasks~\cite{weston2015towards} for testing an agent's ability to answer questions about a short story, the movie dialog dataset~\cite{dodge2015evaluating} to assess an agent's capabilities regarding personal movie recommendation and a Wizard-of-Oz framework~\cite{wen2016network} to evaluate an agent's performance for assisting users in finding restaurants.

In this paper, we bring these two fields together and propose a novel goal-directed task for multi-modal dialogue. The two-player game, called \GW, extends the ReferIt game~\cite{kazemzadeh2014referitgame} to a dialogue setting. To succeed, both players must understand the relations between objects and how they are expressed in natural language. From a machine learning point of view, the \GW challenge is the following: learn to acquire natural language by interaction on a visual task. Previous attempts in that direction~\cite{akinator,wen2016network} do not ground natural language to their immediate environment; instead they rely on an external database through which a conversational agent searches.  

The key contribution of this paper is the introduction of the \GW dataset that contains 155,280 dialogues composed of 831,889 question/answer pairs on 66,537 images extracted from the MS COCO dataset~\cite{lin2014microsoft}. We define three sub-tasks that are based on the \GW dataset and prototype deep learning baselines to establish their difficulty. 
The paper is organized as follows. First, we explain the rules of the \GW game in Sec.~\ref{sec:game}. Then, Sec.~\ref{sec:related_work} describes how \GW relates to previous work. In Sec.~\ref{sec:data_collection} we highlight our design decisions in collecting the  dataset, while Sec.~\ref{sec:data_analysis} analyses many aspects of the dataset. Sec.~\ref{sec:baselines} introduces the questioner and oracle tasks and their baseline models. Finally, Sec.~\ref{sec:discussion} provides a final discussion of the \GW game.

\section{\GW game}
\label{sec:game}

    \GW is a cooperative two-player game in which both players see the picture of a rich visual scene with several objects. One player -- the \textbf{oracle} -- is randomly assigned an object (which could be a person) in the scene.  
    This object is not known by the other player -- the \textbf{questioner} -- whose goal it is to locate the hidden object. To do so, the questioner can ask a series of yes-no questions which are answered by the oracle as shown in  Fig~\ref{fig:front}~and~\ref{fig:examples}. Note that the questioner is not aware of the list of objects, they can only see the whole picture.
Once the questioner has gathered enough evidence to locate the object, they notify the oracle that they are ready to guess the object. We then reveal the list of objects, and if the questioner picks the right object, we consider the game successful. Otherwise, the game ends unsuccessfully. We also include a small penalty for every question to encourage the questioner to ask informative questions. Fig~\ref{fig:oracle}~and~\ref{fig:questioner} in Appendix~\ref{ap:website} display a full game from the perspective of the oracle and questioner, respectively. 

The oracle role is a form of visual question answering where the answers are limited to \emph{Yes}, \emph{No} and \emph{N/A} (not applicable). The \textit{N/A} option is included to respond even when the question being asked is ambiguous or an answer simply cannot be determined. For instance, one cannot answer the question "\textit{Is he wearing glasses?}" if the face of the selected person is not visible. 
The role of the questioner is much harder. They need to generate questions that progressively narrow down the list of possible objects. Ideally, they would like to minimize the number of questions necessary to locate the object. The optimal policy for doing so involves a binary search: eliminate half of the remaining objects with each question. Natural language is often very effective at grouping objects in an image scene. Such strategies depend on the picture, but we distinguish the following types:
\begin{description}
\item[Spatial reasoning] We group objects spatially within the image scene. One may use absolute spatial information -- \textit{Is it on the bottom left of the picture?} -- or relative spatial location -- \textit{Is it to the left of the blue car?}.
\item[Visual properties] We group objects by their size -- \textit{Is it big?}, shape -- \textit{Is it square?} -- or color -- \textit{Is it blue?}.
\item[Object taxonomy] We can use the hierarchical structure of object categories, i.e.  taxonomy, to group objects e.g. \textit{Is it a vehicle?} to refer to both cars and trucks. 
\item[Interaction] We group objects by how we interact with them -- \textit{Can you drive it?}. 
\end{description}

The goal of the \GW task is to enable machines to understand natural descriptions and ground them into the visual world. Note that such higher-level reasoning only occurs when the scene is rich enough i.e. when there are enough objects in the scene. People otherwise tend to fall back to a linear search strategy by simply enumerating objects (often by their category names).

\section{Related work}
\label{sec:related_work}
The \GW game and the data collected from it present opportunities for the extension of current research on image captioning, visual question answering and dialogue systems. In the following, we describe previous work in these areas and relate them to the open challenges offered by \GW. We also mention other relevant work on dataset collection.

\paragraph{Image captioning}
Our work builds on top of the MS COCO dataset~\cite{lin2014microsoft} which consists of 120k images with more than 800k object segmentations. In addition, the dataset provides $5$ captions per image which initiated an explosion of interest from the research community into generating natural language descriptions of images. Several methods have been proposed \cite{karpathy2015deep,vinyals2015show,DBLP:journals/corr/XuBKCCSZB15}, all inspired by the encoder-decoder approach \cite{cho2014learning,sutskever2014sequence} that has proven successful for machine translation. Image captioning research uncovered successful approaches to automatically generate coherent, factual statements about images. Modeling the interactions in \GW requires instead to model the process of asking useful questions about images.

\paragraph{VQA datasets} Visual Question Answering (VQA) tasks form another well known extension of the captioning task. They instead require answering a question given a picture (\emph{e.g. "How many zebras are there in the picture?", "Is it raining outside?"} ). Recently, the VQA challenge \cite{antol2015vqa} has provided a new dataset far bigger than previous attempts \cite{geman2015visual,malinowski2014multi} where, much like in \GW, questions are free-form. 
An extensive body of work has followed from this publication, largely building on the image captioning literature \cite{agrawal2016analyzing,lu2016hierarchical,shih2015look,yang2015stacked}. Unfortunately, many of these advanced methods were shown to marginally improve on simple baselines~\cite{DBLP:journals/corr/JabriJM16}. Recent work~\cite{agrawal2016analyzing} also reports that trained models often report the same answer to a question irrespective of the image, suggesting that they largely exploit predictive correlations between questions and answers present in the dataset. 
The \GW game and dataset attempt to circumvent these issues. Because of the questioner's aim to locate the hidden object, the generated questions are different in nature: they naturally favour spatial understanding of the scene and the attributes of the objects within it, making it more valuable to consult the image. Besides, it only contains binary questions, whose answers we find to be balanced and has twice more questions on average per picture. 


\paragraph{Goal-directed dialogue} \GW is also relevant to the goal-directed dialogue research community. Such systems are aimed at collaboratively achieving a goal with a user, such as retrieving information or solving a problem. 
Although goal-directed dialogue systems are appealing, they remain hard to design. Thus, they are usually restricted to specific domains such as train ticket sales, tourist information or call routing~\cite{pietquin2006probabilistic,singh1999reinforcement,young2013pomdp}. 
Besides, existing dialogue datasets are either limited to fewer than 100k example dialogues~\cite{dodge2015evaluating}, unless they are generated with template formats~\cite{dodge2015evaluating,wen2016network,weston2015towards} or simulation~\cite{pietquin2013survey,schatzmann2006survey} in which case they don't reflect the free-form of natural conversations.
Finally, recent work on end-to-end dialogue systems fail to handle dynamical contexts. For instance,~\cite{wen2016network} intersects a dialogue with an external database to recommend restaurants. Well-known game-based dialogue systems~\cite{20questions,akinator} also rely on static databases.
In contrast, \GW dialogues are heavily grounded by the images. The resulting dialogue is highly contextual and must be based on the content of the current picture rather than an external database. Thus, to the best of our knowledge, the \GW dataset marks an important step for dialogue research, as it is the first large scale dataset that is both goal-oriented and multi-modal.

\paragraph{Human computation games} 
\GW is in line with Von Ahn's seminal work on human computation games~\cite{von2004labeling,von2006peekaboom} who showed that games are an effective way to gather labeled data. The first ESP game~\cite{von2004labeling} was developed to collect image tags, and was later extended to Peekaboom~\cite{von2006peekaboom} to gather object segmentations. These games were developed more than a decade ago, when object recognition was in its infancy and served a different purpose than \GW. 

\paragraph{ReferIt} Probably closest to our work is the \Referit game~\cite{kazemzadeh2014referitgame,mao2015generation,DBLP:journals/corr/YuPYBB16}. In this game, one player observes an annotated object in a scene, for which they need to generate an expression that refers to it (\emph{e.g.\ \"the man wearing the white t-shirt\"}). The other player then receives this expression and subsequently clicks on the location of the object within the image. The original dataset~\cite{kazemzadeh2014referitgame} uses the IMAGEClef dataset~\cite{escalante2010segmented}, while three recent extensions~\cite{mao2015generation,DBLP:journals/corr/YuPYBB16} were built on top of MS COCO. All three databases select images with only $2-4$ objects of the \emph{same} category. In contrast, \GW picks images with $3-20$ objects without further restrictions on the object class, and thus contains three times more images than the \Referit datasets. To further investigate the difference between \Referit and \GW, we compare three samples for the \emph{same} selected object in Fig~\ref{fig:referit} in Appendix~\ref{ap:gw_samples}. While \Referit directly locates the object with a single expression, \GW iteratively narrows down the object by means of positive and \emph{negative} feedback on questions. We also observe that \GW dialogues favor more abstract concepts, such as "\textit{Is it edible?}" or "\textit{Is it on oval plate?}" than \Referit.

\begin{figure*}[t]
\centering
\begin{subfigure}{0.32\linewidth}
\includegraphics[width=\linewidth]{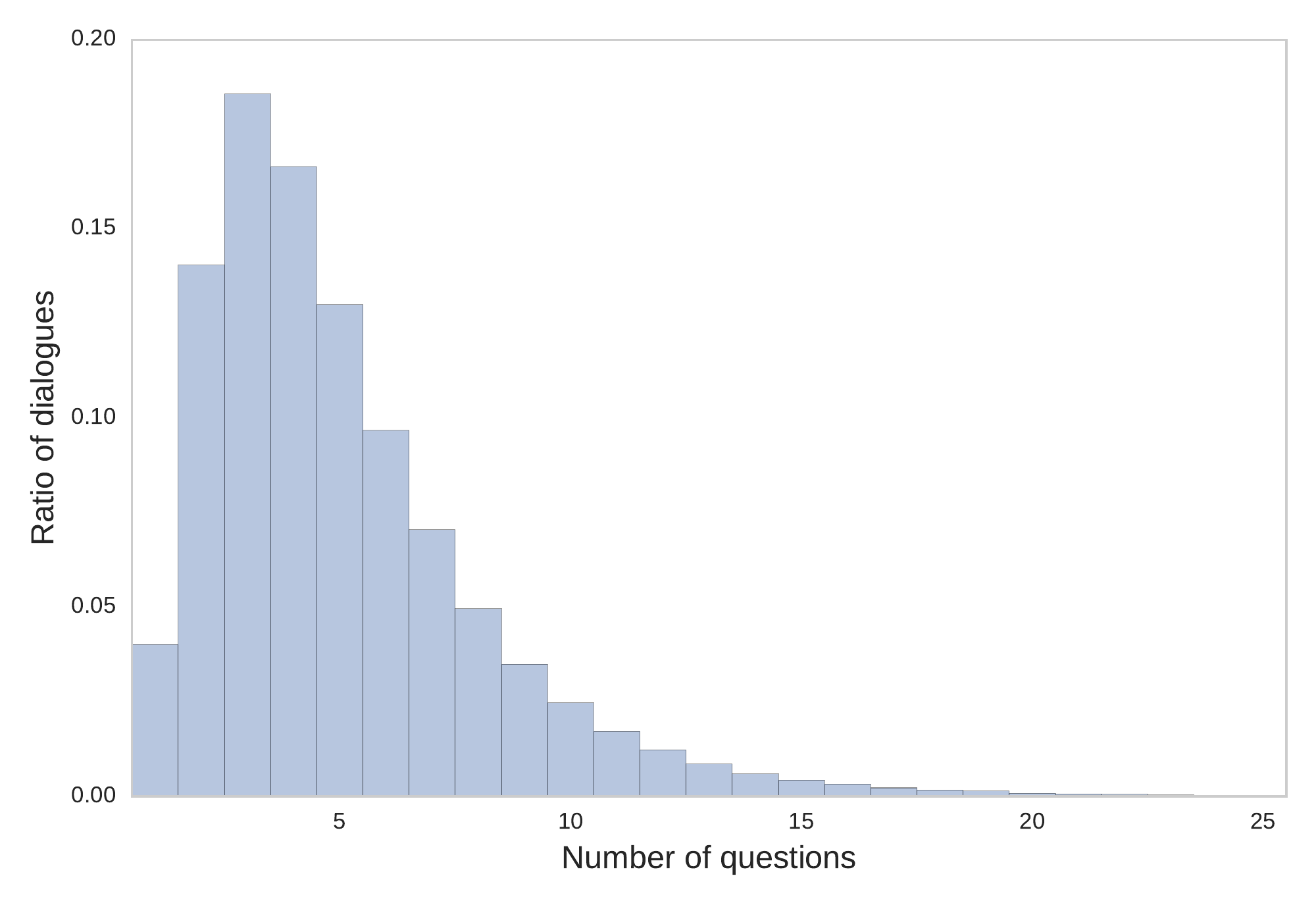}
\vskip -0.6em
\caption{}
\label{fig:stat:q_d}
\vskip -1.2em
\end{subfigure}
\begin{subfigure}{0.32\linewidth}
\includegraphics[width=\linewidth]{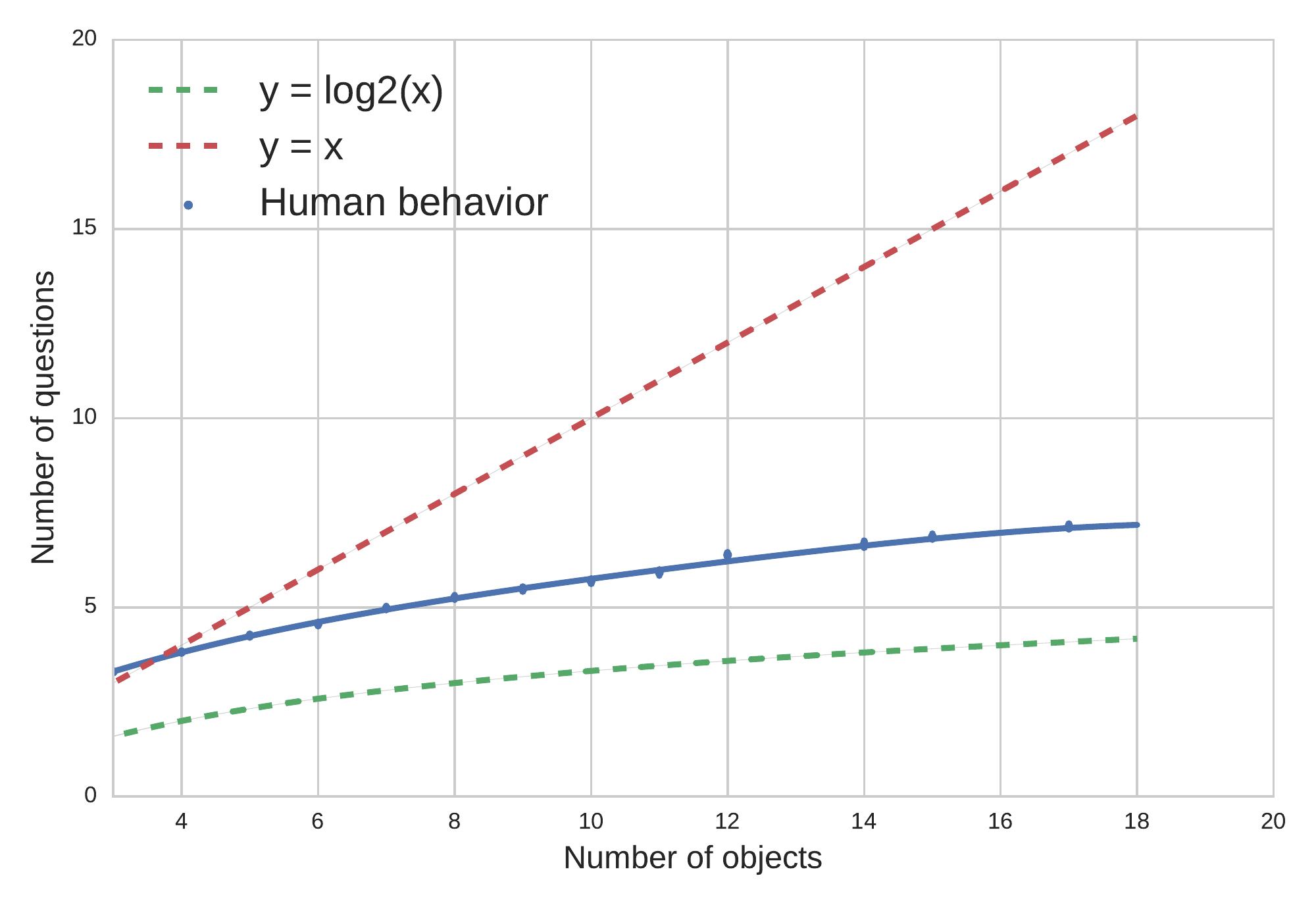}
\vskip -0.6em
\caption{}
\label{fig:stat:o_q}
\vskip -1.2em
\end{subfigure}
\begin{subfigure}{0.32\linewidth}
\includegraphics[width=\linewidth]{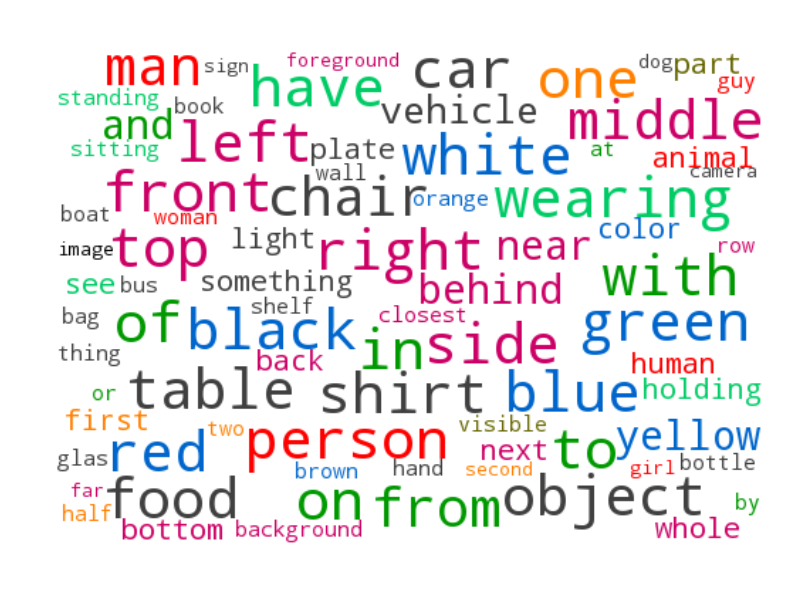}
\vskip -1.3em
\caption{}
\label{fig:cloud}
\vskip -1.2em
\end{subfigure}
\vskip -0.5em
\caption{(a) Number of questions per dialogue (b) Number of questions per dialogue vs the number of objects within the picture (c) Word cloud of \GW vocabulary with each word proportional to its frequency. Words are colored based on a hand-crafted clustering. Uninformative words such as \emph{"it"}, \emph{"is"} are manually removed.}
\label{fig:statistic}
\vskip -1em
\end{figure*}

\section{\GW Dataset}
\subsection{Data collection}
\label{sec:data_collection}

\paragraph{Images} We use a subset of the training and validation images and objects of the MS COCO dataset~\cite{lin2014microsoft}. We first discard objects that are too small ($\text{area} <  500\text{px}^2$) to be decently located by a human observer. Then, we only keep images containing three to twenty objects, to avoid trivial or overly complicated images. In total, we keep 77,973 images with 609,543 objects. We verified that this selection does not significantly alter the original dataset distribution.

\paragraph{Amazon Mechanical Turk} The data collection was crowd-sourced on Amazon Mechanical Turk (AMT)~\cite{buhrmester2011amazon}. We created two separate tasks -- known as HITs on AMT -- for the questioner and oracle roles, and rewarded the questioner slightly more than the oracle. 
We ensured the quality of the data collection by several means. First, the workers had to go through a qualification round which consisted of successfully completing $10$ games while producing fewer than $4$ mistakes or disconnects.  After qualification, HITs continue to consist of a batch of $10$ successful games. We incentivize the worker to produce as many successful dialogues in a row by providing bonuses for making fewer mistakes. Secondly, players could report on each other and players were banned after a certain number of reports.  Thus, players were incentivized to cooperate. In the end, we only kept dialogues from qualified people and successful dialogues from the qualification round. 
In contrast to traditional dataset collection, our game requires an interactive session between two players. Fortunately, we found that the \GW game was highly engaging. A total of more than 10K people participated in our HITs, and our top ten participants played over $2,000$ games each. 
Since questions were manually typed, they could contain spelling mistakes. Thus, we retrieved all questions containing words that do not occur in an English dictionary and manually corrected the $1000$ most common words. For the remaining 30k questions, we created two HITs that to correct the spelling mistakes. See Figure~\ref{fig:mistake_fixer} in Appendix~\ref{ap:website} for further details. 


\subsection{Data analysis}
\label{sec:data_analysis}
In the following, we explore properties of the data we collected using the \GW game. We provide global statistics, examine the vocabulary used by the questioners and highlight the relationship between properties of objects to guess and the odds of having a successful dialogue.

\paragraph{Dataset statistics}
The raw \GW dataset is composed of 155,280 dialogues containing 821,889 question/answer pairs on 66,537 unique images and 134,073 unique objects. The answers are respectively 52.2\% \emph{no}, 45.6\% \emph{yes} and 2.2\% \emph{N/A}. On average, there are 5.2 questions per dialogue and 2.3 dialogues per image. The dialogues contain 3,986,192 word tokens in total, making up 11,465 different words with at least one occurrence and 5,444 words with at least 3 occurrences. Moreover, 84.6\% of the dialogues are successful, 8.4\% are unsuccessful and 7.0\% are not completed (disconnection, timeout etc.). Thus, different subsets co-exist in the \GW dataset, we will refer to the dataset as full, finished and successful when we include all the dialogues, all finished dialogues (successful and unsuccessful) or only successful dialogues, respectively. For more details, the previous statistics are broken down into dataset types in Tab~\ref{tab:dataset_statistics}.

\begin{table}
\centering
\scalebox{0.79}{
\begin{tabular}{|l|c|c|c|}
\hline
 & Full & Finished & Success\\
\hline
\# dialogues         & 155,280   & 144,434    & 131,394  \\
\# questions         & 821,889   & 732,081    & 648,493  \\
\# words             & 3,986,192 & 3,540,497  & 3,125,219\\
\# voc. size         & 11,465    & 10,985     & 10,469   \\
\# voc. size (3+)    & 5,444     & 5,179      & 4,919    \\
\# images          & 66,537    & 65,112     & 62,954   \\
\# objects           & 134,073   & 125,349    & 114,271  \\
\hline
\end{tabular}}
\caption{\GW statistics split by dataset types.}
\label{tab:dataset_statistics}
\vskip -1em
\end{table}

\paragraph{Question distributions} To get a better understanding of the \GW games, we show the number of questions within a dialogue and the average number of questions given the number of objects within a image in Fig~\ref{fig:statistic}. First, the number of questions within a dialogue decreases exponentially, as players tend to shorten their dialogues to speed up the game (and therefore maximize their gains). More interestingly, we observe that the average number of questions given the number of objects within an image appears to follow a function that grows at a rate between logarithmically and linearly. A questioning strategy of simply listing objects (e.g. "\emph{is it the chair}", etc.) would imply linear growth in the number of questions, while the optimal binary search strategy would imply logarithmic growth. Thus the human questioners seem to imply a strategy that is somewhere in between. We conjecture three reasons why humans do not achieve the optimal search strategy. First, the questioner does not have access to the ground truth list of objects in the picture, and might, therefore, overestimate the number of objects. Second, some humans tend to favor a linear search strategy. Finally, the questioner may ask additional questions to confirm that he has located the right object. This can be important in the presence of possible oracle errors.

\paragraph{Vocabulary} To gain insight into the vocabulary used by the questioner, we compute the frequency of words in the \GW corpus and display the most frequent words as a word cloud in Fig~\ref{fig:cloud}. Several key words clearly stand out. As explained in Sec.~\ref{sec:game}, some of those key words refer to abstract object properties such as \emph{person} or \emph{object}, spatial locations such as \emph{right/left} or \emph{side} and visual features such as \emph{red/black/white}. Furthermore, prepositions are also heavily used to express relationships between objects. 
To better understand the sequential aspect of the questions, we study the evolution of the vocabulary at each question round.
We observe that questioners use abstract object properties such as \emph{human/object/furniture} only at the beginning of the dialogues, and quickly switch to either spatial or visual terms such as \emph{left/right}, \emph{white/red} or \emph{table,chair}. This can be highlighted by applying a Dynamic Topic Model~\cite{blei2006dynamic} to study the evolution of topics over the course of the dialogue as shown in Fig~\ref{fig:DTM} in Appendix~\ref{ap:statistics}. 

\begin{figure*}[t]
\centering
\begin{subfigure}{0.30\linewidth}
\includegraphics[width=\linewidth]{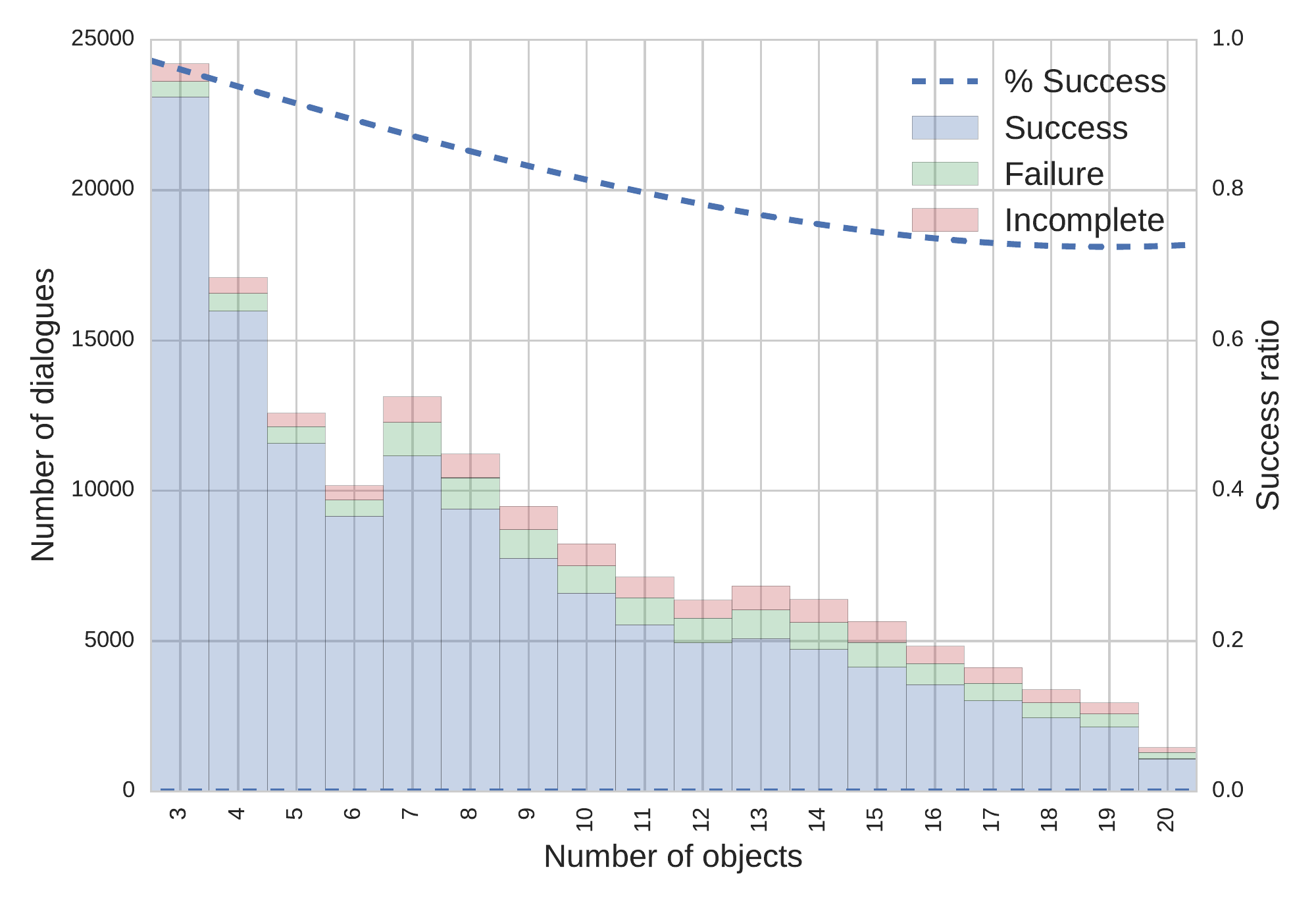}
\vskip -0.6em
\caption{}
\label{fig:success_objects}
\vskip -1.2em
\end{subfigure}
\begin{subfigure}{0.33\linewidth}
\includegraphics[width=\linewidth]{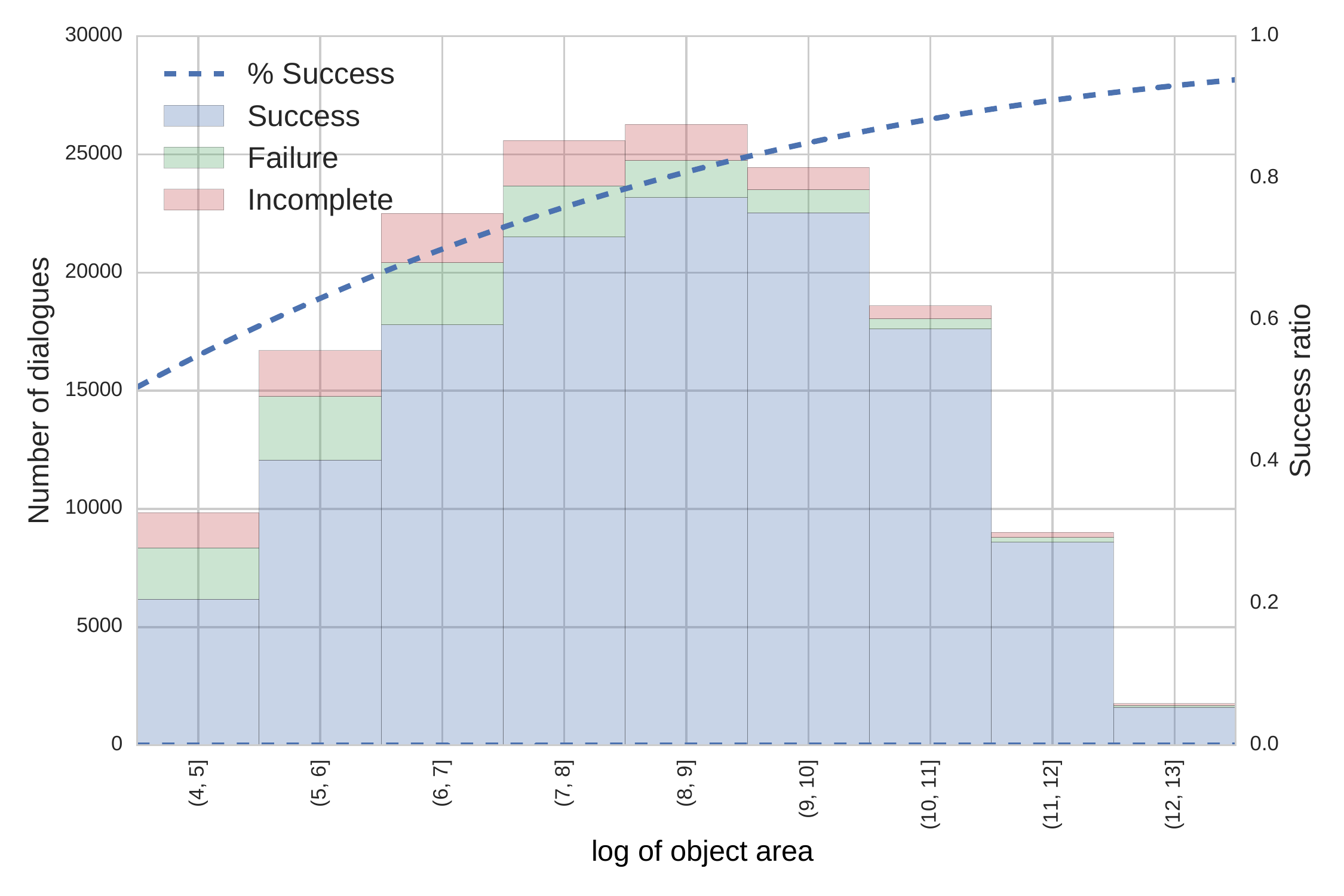}
\vskip -0.6em
\caption{}
\label{fig:success_area}
\vskip -1.2em
\end{subfigure}
\begin{subfigure}{0.32\linewidth}
\includegraphics[width=\linewidth]{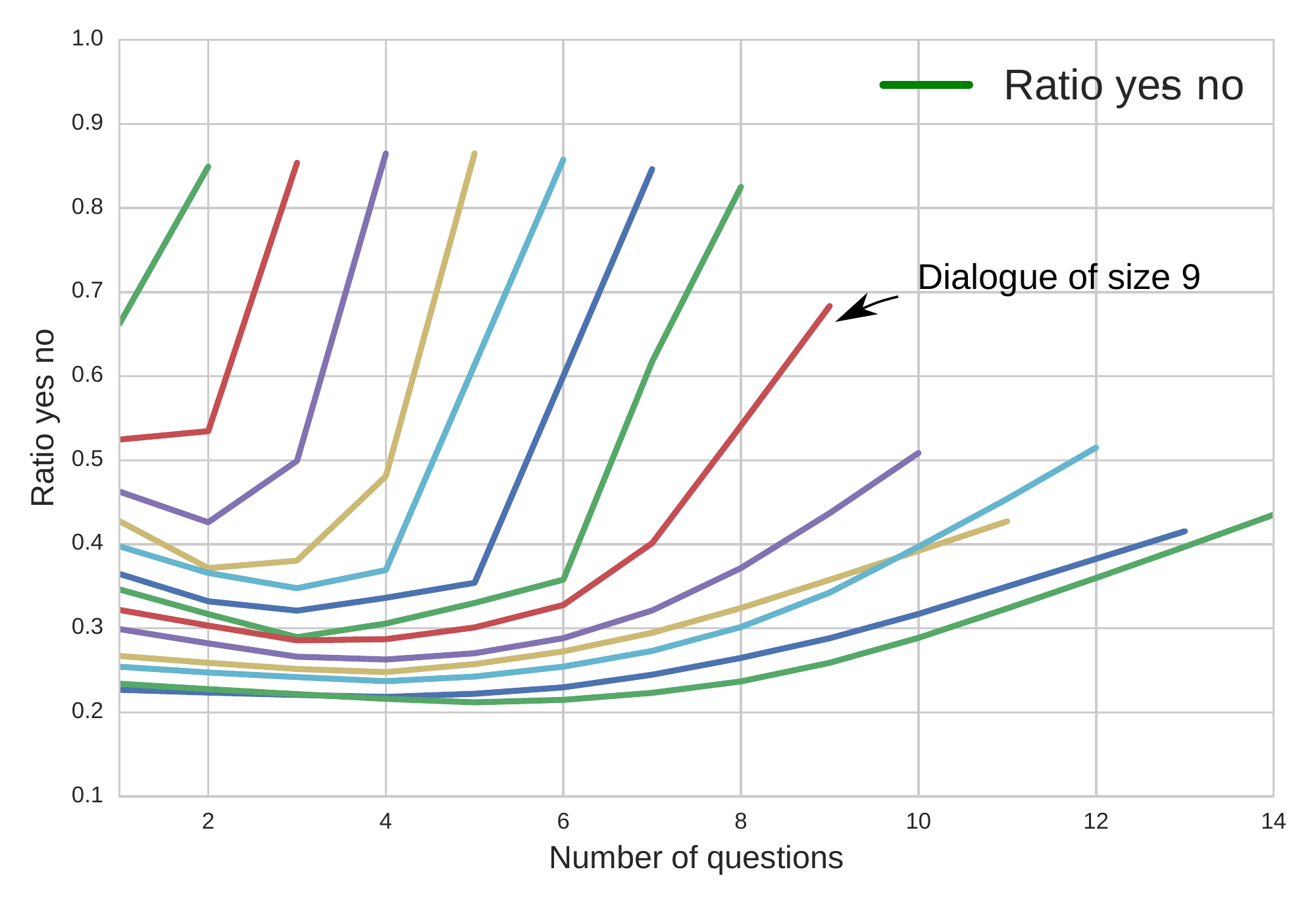}
\vskip -0.6em
\caption{}
\label{fig:yes_no}
\vskip -1.2em
\end{subfigure}
\vskip -0.5em
\caption{(a-b) Histogram of absolute/relative successful dialogues with respect to the number of objects and the size of the objects, respectively. (c) Evolution of answer distribution clustered by the dialogue length}
\label{fig:success}
\vskip -1em
\end{figure*}

\paragraph{Elements of success} To investigate whether certain object properties favour success, we compute the success ratio of dialogues relative to: the size of the unknown objects in Fig~\ref{fig:success_area}, the number of objects within the images in Fig~\ref{fig:success_objects}, the object category, the location of objects within the images and the size of the dialogues in Fig~\ref{fig:success_category}, Fig~\ref{fig:success_spatial} and Fig~\ref{fig:success_length} in Appendix ~\ref{ap:statistics}, respectively. As one may expect, the more complex the scene is, the lower the success rate is. When there are only 3 objects, the questioner has 95\% success rate, while this ratio drops to around 70\% with 20 objects. Similarly, big objects are almost always found while the smallest one are only found 60\% of the time. Questioners easily find objects in the middle of the picture but have more difficulties to find them on the border. Finally, objects from categories that are often grouped together, e.g. \emph{bananas} or \emph{books}, have a lower success rates.

\paragraph{Miscellaneous} In Fig~\ref{fig:yes_no} we break down the ratio of yes-no answers within the dialogues. While the first yes-no answers are balanced for small dialogues, they often terminate with a final \emph{yes}. In contrast, long dialogues often start with a higher proportion of negative answers which slowly decrease during the exchange. 

\subsection{Dataset release}
We split the \GW dataset by randomly assigning 70\%, 15\% and 15\% of the \emph{images} and its corresponding dialogues to the training, validation and test set. This way of dividing the data ensures that we evaluate performance on images not seen during training. The \GW dataset is available at \url{https://guesswhat.ai/download}.

\section{Baselines}\label{sec:baselines}
We now empirically investigate the difficulty of the oracle and questioner tasks. To do so, we trained reasonable baselines for each task and measured their performance. 

Formally, a \GW game revolves around an image $I \in \mathbb{R}^{M\times N}$ containing a set of $K$ segmented objects $\{O_1, \dots, O_K\}$. Each object $O_k$ is assigned an object category $c_k \in \{1, \dots, C\}$ and has a pixel-wise segmentation mask $S_k \in \{0, 1\}^{M\times N}$ to specify its location and size. The game further consists of a sequence of questions and answers $D = \{q_1, a_1, \hdots, q_J, a_J\}$, produced by the questioner and oracle. We will use $q_{<j}$ and $a_{<j}$ to refer to the first $j-1$ questions and answers, respectively. Each question $q_j$ contains a sequence of $N_j$ tokens, i.e.\ $q_j = \{w_{j1}, \dots, w_{jN_j}\}$, where $w_{ji}$ is taken from a vocabulary $V$ and represents the token at position $i$ in question $j$. Each answer is either \emph{Yes}, \emph{No} or \emph{N/A}, i.e.\ $a_j \in \{\text{Yes, No, N/A}\}$. Finally, the oracle has access to the identity of the correct object $O_{\text{correct}}$, and the prediction of the questioner will be denoted as $O_{\text{predict}}$. 

\subsection{Oracle baselines}
\begin{figure}[t]
\begin{center}
\includegraphics[width=0.95\linewidth]{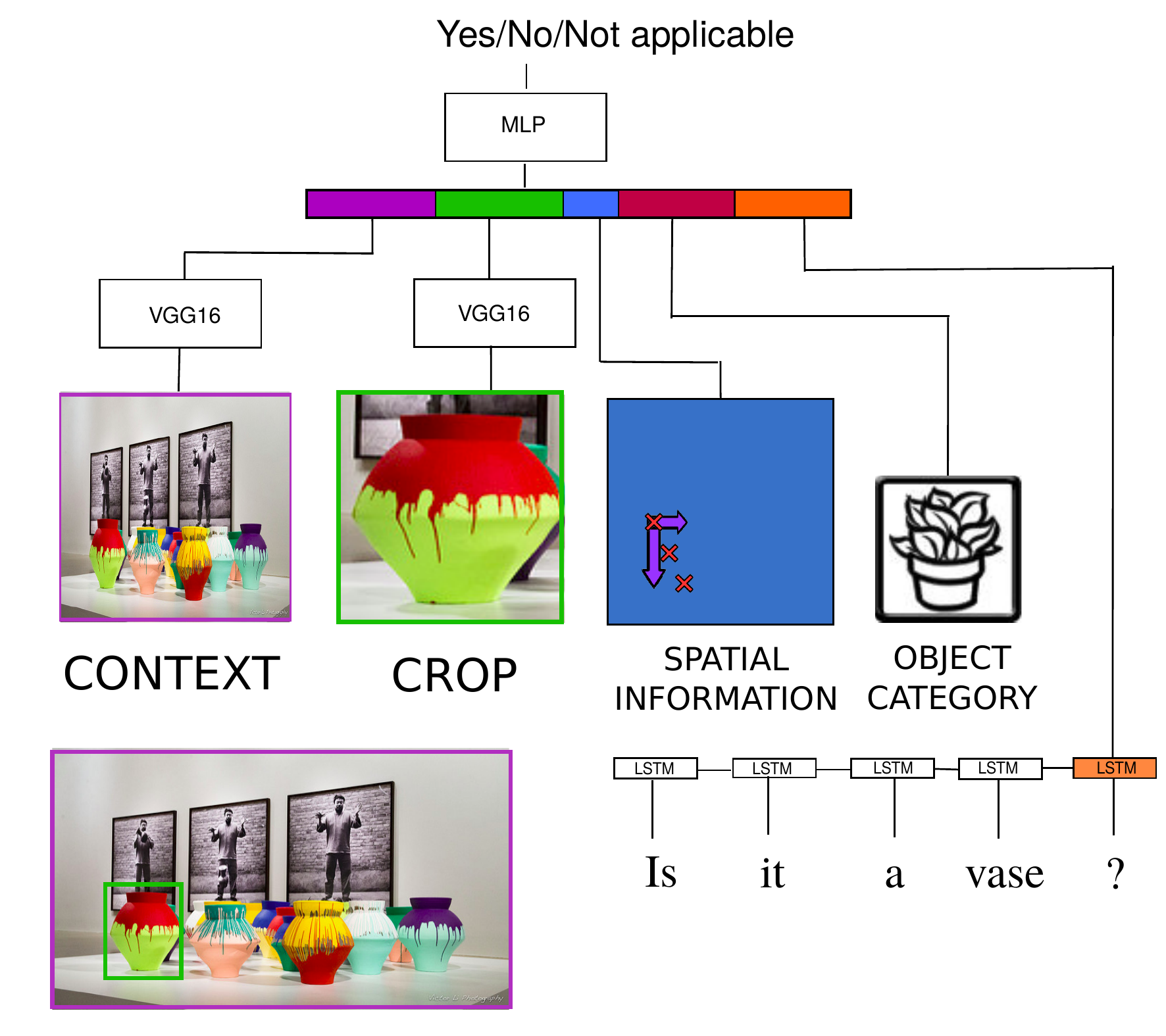}
\end{center}
   \caption{An schematic overview of the "Image + Question + Crop + Spatial + Category" oracle model. }
\label{fig:oracle_model}
\vskip -1em
\end{figure}

The oracle task requires to produce a yes-no answer for any object within a picture given a natural language question. We first introduce our model and then outline its results to get a better understanding of the \GW dataset. 

\paragraph{Model} 
We propose a simple neural network based approach to this model, illustrated in Fig~\ref{fig:oracle_model}. Specifically, we use an appropriate neural network architecture to embed each of the following information: the image $I$, the cropped object from $S$, its spatial information, its category $c$ and the current question $q$. These embeddings are then concatenated as a single vector and fed as input to a single hidden layer MLP that outputs the final answer distribution using a softmax layer. Finally, we minimize the cross-entropy error during the training and report the classification error at evaluation time.  

The details on how we compute the embeddings are as follows.
To embed the full image, it is rescaled to a $224$ by $224$ image and is passed through a pre-trained VGG network to obtain its FC8 features. As for the selected object, it is first cropped by finding the smallest rectangle that encapsulates it, based on its segmentation mask. We then rescale the crop to a $224$ by $224$ square, before obtaining its FC8 features from the pre-trained VGG network. Although we could use the mask to drop out pixels around the selected object, we keep the crop as is since pre-trained VGG networks are exposed to such background noise during their training. 
We also embed the spatial information of the crop, to help locate the cropped object within the whole image. To do so, we follow the approach of \cite{hu2016natural,DBLP:journals/corr/YuPYBB16} and extract an 8-dimensional vector of the location of the bounding box:
\begin{align}
x_{spatial} = [ &x_{min}, y_{min}, x_{max}, y_{max},\notag\\ &x_{center}, y_{center}, w_{box}, h_{box}]\label{eq:spatial}
\end{align}
where $w_{box}$ and $h_{box}$ denote the width and height of the bounding box, respectively. We normalize the image height and width such that coordinates range from $-1$ to $1$, and place the origin at the center of the image.
As for the object category, we convert its one-hot class vector into a dense category embedding using a learned look-up table.
Finally, the embedding of the current natural language question $q$ is computed using an Long Short-Term Memory (LSTM) network~\cite{hochreiter1997long} where questions are first tokenized by using the word punct tokenizer from the python nltk toolkit~\cite{bird2009natural}. For simplicity, we decided to ignore the question-answer pairs history $q_{<t}$ in our oracle baseline.

\paragraph{Training setting}
We train all oracle models on the full dataset. During training, we keep the parameters of the VGG network fixed, and optimize the LSTM, object category/word look-up tables and MLP parameters by minimizing the negative log-likelihood of the correct answer. We use ADAM~\cite{DBLP:journals/corr/KingmaB14} for optimization and train for at most $15$ epochs. We use early stopping on the validation set, and report the train, valid and test error.

\paragraph{Results} 
We report results for several oracle models using a different set of inputs in Table~\ref{tab:oracle_baseline}. We name the model after the input we feed to it. For instance, (Question+Category+Spatial+Image) refers to the network fed with the question $q$, the object category $c$, the spatial features $x_{spatial}$ and the full image $I$. The results of all subsets are reported in Table~\ref{tab:all_oracle_baseline} in Appendix~\ref{ap:statistics}. 

Because the \GW dataset is fairly balanced, simply outputting the most common answer in the training set -- No -- results in a high $50.8\%$ error rate. Solely providing the image or crop features barely improves upon this result. Only using the question slightly improves the error rate to $41.2\%$. We speculate that this small bias comes from questioners that refer to objects that are never segmented or overrepresented categories. As hoped, we observe that the error rate significantly drops ($< 31\%$) when we finally feed information on the object to guess (crop, spatial or category) to the model. We find that crop and category information are redundant: the (Question+Category) and (Question+Crop) model achieve respectively $29.2\%$ and $25.7\%$ error, while the combined model (Question+Category+Crop) achieves $24.7\%$. In general, we expect the object crop to contain additional information, such as color information, beside the object class. However, we find that the object category outperforms the object crop embedding. This might be partly due to the imperfect feature extraction from the crops. Finally, our best performing model combines object category and its spatial features along with the question.

\begin{table}
\scalebox{0.79}{
\begin{tabular}{|l|l|l|l|}
\hline
Model & Train err & Val err & Test err\\
\hline
\hline
Dominant class (no) & 47.4\% & 46.2\% & 50.9\%\\
Question & 40.2\% & 41.7\% & 41.2\%\\
Image & 45.7\% & 46.7\% & 46.7\%\\
Crop & 40.9\% & 42.7\% & 43.0\%\\
\hline
Question + Crop & 22.3\% & 29.1\% & 29.2\%\\
Question + Image & 37.9\% & 40.2\% & 39.8\%\\
Question + Category & 23.1\% & 25.8\% & 25.7\%\\
Question + Spatial & 28.0\% & 31.2\% & 31.3\%\\
\hline
Question + Category + Spatial & 17.2\% & 21.1\% & \textbf{21.5\%}\\
Question + Category + Crop & 20.4\% & 24.4\% & 24.7\%\\
Question + Spatial + Crop & 19.4\% & 26.0\% & 26.2\%\\
\hline
Question + Category + Spatial + Crop & 16.1\% & 21.7\% & 22.1\%\\
Question + Spatial + Crop + Image & 20.7\% & 27.7\% & 27.9\%\\
Question + Category + Spatial + Image & 19.2\% & 23.2\% & 23.5\%\\
\hline
\end{tabular}}
\caption{Classification errors for the oracle baselines on train, valid and test set. The best performing model is "Question + Category + Spatial" and refers to the MLP that takes the question, the selected object class and its spatial features as input.}
\label{tab:oracle_baseline}
\vskip -1em
\end{table}

\subsection{Questioner baselines}
Given an image, the questioner must ask a series of questions and guess the correct object. We separate the questioner task into two different sub-tasks that are trained independently:
\begin{description}
\item[Guesser] Given an image $I$ and a sequence of questions and answers $D_J$, predict the correct object $O_{\text{correct}}$ from the set of all objects $O$.
\item[Question Generator] Given an image $I$ and a sequence of $T$ questions and answers $D_{\leq T}$, produce a new question $q_{T+1}$. 
\end{description}
In general, one also needs a module to determine when to start guessing the object (and stop asking questions). In our baseline, we bypass this issue by fixing the number of questions to $5$ for the question generator model.

\paragraph{Guesser}
\begin{figure}[t]
    \centering
    \includegraphics[width=0.95\linewidth]{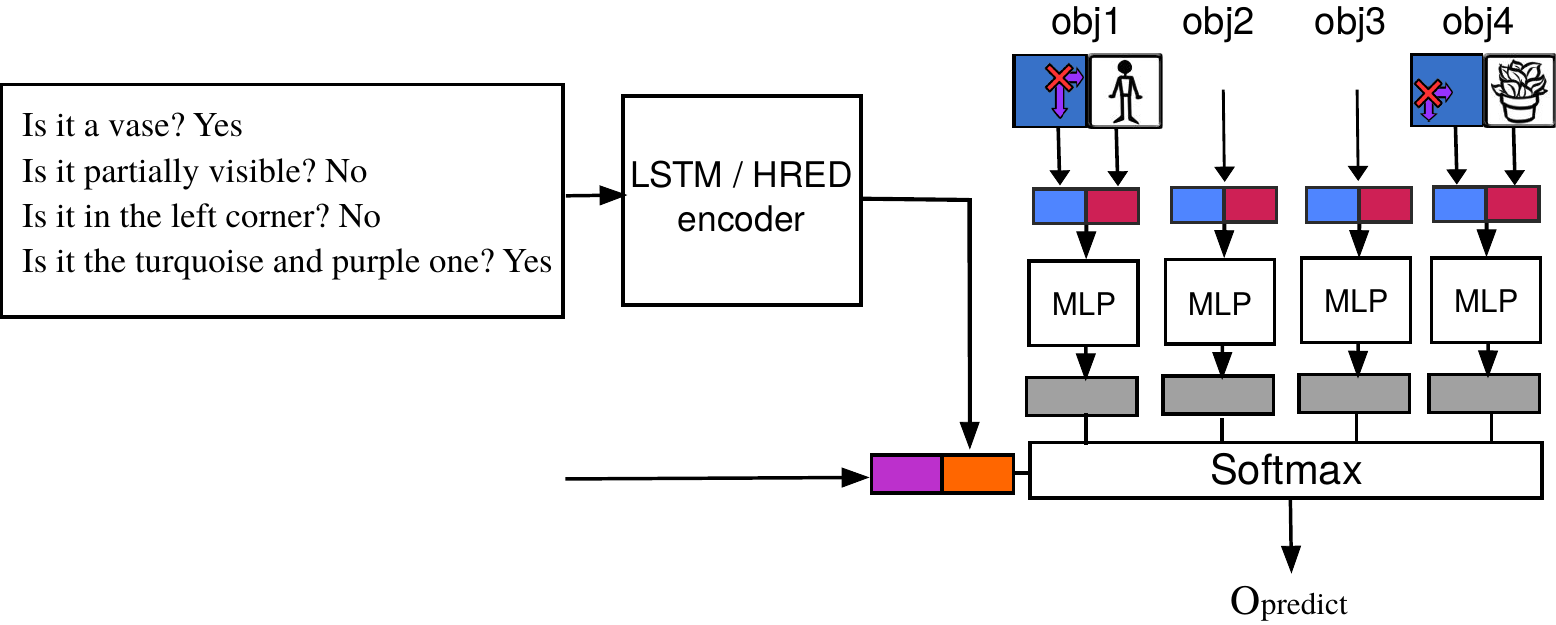}
    \caption{Overview of the guesser model for an image with 4 segmented objects. The weights are shared among the MLPs, this allows for an arbitrary number of objects. }
    \label{fig:guesser}
\end{figure}

\begin{table}[t]
\centering
\scalebox{0.79}{
\begin{tabular}{|l|l|l|l|}
\hline
Model & Train err & Val err & Test err\\
\hline
Human & 9.0\% & 9.2\% & 9.2\%\\
Random & 82.9\% & 82.9\% & 82.9\%\\
\hline
LSTM & 27.9\% & 37.9\% & \textbf{38.7}\%\\
HRED & 32.6\% & 38.2\% & 39.0\%\\
\hline
LSTM+VGG & 26.1\% & 38.5\% & 39.5\%\\
HRED+VGG & 27.4\% & 38.4\% & 39.6\%\\
\hline
\end{tabular}
}
\caption{Classification errors for the guesser baselines on train, valid and test set.}
\label{table:guesser_baseline}
\vskip -1em
\end{table}

The role of the guesser model is to predict the correct object. To do so, the guesser has access to the image, the dialogue and the list of objects in the image. We encode the image by extracting its FC8 features from VGG16 network. A dialogue of a \GW game is a sequence on two different levels: there is a variable number of question-answer pairs where each question in turn consists of a variable-length sequence of tokens. This can be encoded into a fixed size vector by using either an LSTM encoder~\cite{hochreiter1997long} or an HRED encoder~\cite{DBLP:journals/corr/SerbanSBCP15}. While the LSTM encoder considers the dialogue as one flat sequence, HRED explicitly models the hierarchy by two different Recurrent Neural Networks (RNN). First, an encoder RNN creates a fixed-size representation of a question or answer by reading in its tokens and taking the last hidden state of the RNN. This representation is then processed by the context RNN to obtain a representation of the current dialogue \emph{state}. For both models, we concatenate the image and dialogue features and do a dot-product with the embedding for all the objects in the image, followed by a softmax to obtain a prediction distribution over the objects. Given  the best performance of the "Question+Category+Spat" oracle model, we represent objects by their category and their spatial features. More precisely, we concatenate the 8-dimensional spatial representation (see Eq.~\ref{eq:spatial}) and the object category look-up and pass it through an MLP layer to get an embedding for the object. Note that the MLP parameters are shared to handle the variable number of objects in the image. See Fig~\ref{fig:guesser} for an overview of the guesser with HRED and LSTM.

Table~\ref{table:guesser_baseline} reports the results for the guesser baselines using human-generated dialogues. As a first baseline, we report the performance of a random guesser which does not use the dialogue information. We split the guesser results based on whether they use the VGG features or not. In general, we find that including VGG features does not improve the performance of the LSTM and HRED models. We hypothesize that the VGG features are a too coarse representation of the image scene, and that most of the visual information is already encoded in the question and the object features. Surprisingly, we find LSTMs to perform slightly better than the sophisticated HRED models. 

\paragraph{Question Generator}
\begin{figure}[t]
    \centering
    \includegraphics[width=0.95\linewidth]{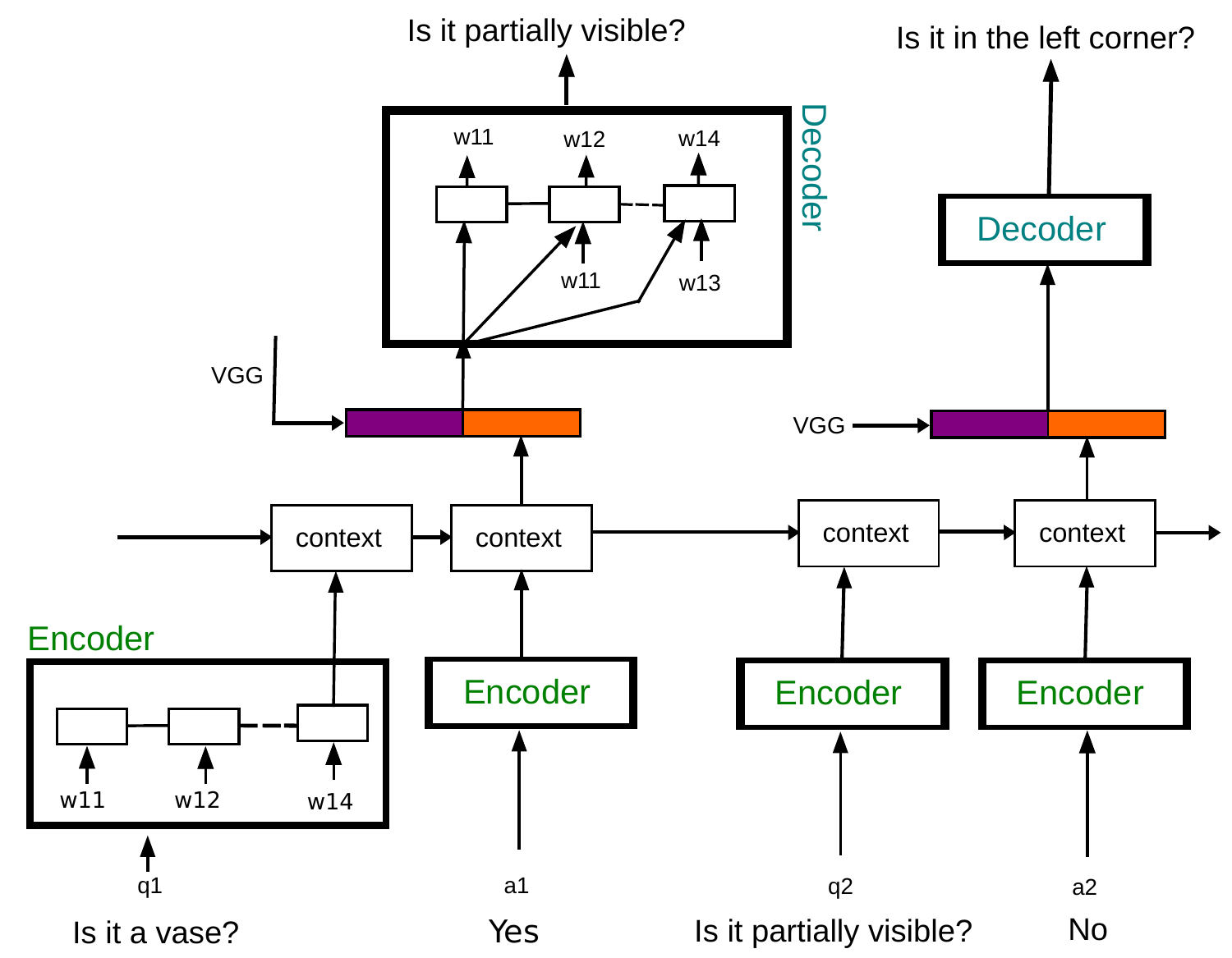}
    \caption{HRED model conditioned on the VGG features of the image. To avoid clutter, 
    we here only show the part of the model that defines a distribution over the third question given the first two questions, its answers and the image $P(q_2|q_{<2},a_{<2},I)$.
    The complete HRED model models the distribution over all questions.
    }
    \label{fig:HRED}
    \vskip -1em
\end{figure}

The question generation task is hard for several reasons. First, it requires high-level visual understanding to ask meaningful questions. Second, the generator should be able to handle long-term context to ask a sequence of relevant questions, which is one of the most challenging problems in dialogue systems. Additionally, we evaluate the question generator using the imperfect oracle and imperfect guesser, which introduces compounding errors. 

Hierarchical recurrent encoder decoder (HRED)~\cite{DBLP:journals/corr/SerbanSBCP15} is the current state of the art method for natural language generation tasks. We extend this model by conditioning on the VGG features of the image as illustrated in Fig~\ref{fig:HRED}. Finally, we train our proposed model by maximizing the conditional log-likelihood:
\begin{align}
   \log P(Q|A,I) = & \log \prod_{j=1}^J P(q_{j}| q_{<j}, a_{<j}, I)\\
    =&\log \prod_{j=1}^J \prod_{i=1}^{N_j} P(w_{ji} | w_{j <i}, a_{\leq j}, I)
\end{align}
with respect to the described parameters. At test time, we use a beam-search to approximately find the most probable question $q_j$. 
Evaluating the questioner model requires a pre-trained oracle and a pre-trained guesser model. We use our questioner model to first generate a question which is then answered by the oracle model. We repeat this procedure 5 times to obtain a dialogue. We then use the best performing guesser model to predict the object and report its error as the metric for the QGEN model. Since we use ground truth answers during the QGEN training while we use oracle answers at test time, there is a mismatch between the training and testing procedure. This can be avoided by using the oracle answers also during training time. We call these models QGEN+GT and QGEN+ORACLE respectively. 

Table~\ref{table:questioner_baseline} shows the results. A guesser based on human generated dialogues achieves $38.7\%$ error. The Question Generator models achieve reasonable performance which lies in between the random performance and the performance of the guesser on human dialogues. We observe that using the Oracle's answers while training the Question Generator introduces additional errors which significantly deteriorates performance. Some example dialogues generated by the QGen+GT model are shown in Fig. \ref{fig:correct_samples} and \ref{fig:incorrect_samples}. 

\begin{table}
\centering
\scalebox{0.79}{
\begin{tabular}{|l|l|}
\hline
Model & Error\\
\hline
Human generated dialogue & 38.7\%\\
QGen+GT & 53.2\%\\
QGen+ORACLE & 66.0\%\\
Random & 82.9\%\\
\hline
\end{tabular}
}
\caption{Test error for the question generator models (QGEN) based on VGG+HRED(FT) guesser model. We here report the accuracy error of the guesser model fed with the questions from the QGEN model.}
\label{table:questioner_baseline}
\vskip -1em
\end{table}

\section{Discussion}
\label{sec:discussion}
We introduced the \GW game, a novel framework for multi-modal dialogue. To the best of our knowledge, we present the first large-scale dataset involving images and dialogue. A wide range of challenges may arise from this union as they rely on different fields of machine learning such as natural language understanding, generative models or computer vision. 
\GW turns out to be an engaging game that greatly decreases the cost for collection of a big dataset required for modern algorithms. As a second contribution, we introduced three tasks based on the questioner and oracle role. In each case, we prototyped a neural architecture as a first baseline. We analyzed these results and presented a quantitative description of the \GW dataset.  

We believe \GW could allow for a myriad of other applications that may either be based on the game itself or extending the database to other tasks. For instance, it can be interesting to compute a confidence interval before proceeding to the final guess.  Differently, \GW could be a test bed for one-shot learning~\cite{fei2006one} of guessing new object categories, transfer learning on line-drawing images~\cite{castrejon2016learning} or using questions from another language. Thus, the \GW dataset offers an opportunity to develop original machine learning tasks upon it.

\paragraph{Acknowledgement} The authors would like to acknowledge the stimulating environment provided by the MILA and SequeL labs. We thank all members of the MILA lab who participated in a trial run of the data collection, and all workers of AMT who participated in our HITs. We thank Jake Snell, Mengye Ren, Laurent Dinh, Jeremie Mary and Bilal Piot for helpful discussions. We acknowledge the following agencies for research funding and computing support: NSERC, Calcul Qu\'{e}bec, Compute Canada, the Canada Research Chairs and CIFAR, CHISTERA IGLU and CPER Nord-Pas de Calais/FEDER DATA Advanced data science and technologies 2015-2020. SC is supported by a FQRNT-PBEEE scholarship.

{\small
\bibliographystyle{ieee}
\bibliography{egbib}

\begin{thebibliography}{10}\itemsep=-1pt

\bibitem{20questions}
{20 Questions}.
\newblock \url{http://www.20q.net/}.
\newblock Accessed: 2016-09.

\bibitem{akinator}
{Akinator}.
\newblock \url{en.akinator.com/}.
\newblock Accessed: 2016-09.

\bibitem{agrawal2016analyzing}
A.~Agrawal, D.~Batra, and D.~Parikh.
\newblock {Analyzing the Behavior of Visual Question Answering Models}.
\newblock {\em arXiv preprint arXiv:1606.07356}, 2016.

\bibitem{von2004labeling}
L.~V. Ahn and L.~Dabbish.
\newblock {Labeling images with a computer game}.
\newblock In {\em Proc. of the SIGCHI conference on Human factors in computing
  systems}. ACM, 2004.

\bibitem{von2006peekaboom}
L.~V. Ahn, R.~Liu, and M.~Blum.
\newblock {Peekaboom: a game for locating objects in images}.
\newblock In {\em Proc. of the SIGCHI conference on Human Factors in computing
  systems}. ACM, 2006.

\bibitem{antol2015vqa}
S.~Antol, A.~Agrawal, J.~Lu, M.~Mitchell, D.~Batra, Z.~Lawrence, and D.~Parikh.
\newblock {Vqa: Visual question answering}.
\newblock In {\em Proc. of ICCV}, 2015.

\bibitem{bird2009natural}
S.~Bird, E.~Klein, and E.~Loper.
\newblock {\em Natural language processing with Python}.
\newblock O'Reilly Media, Inc., 2009.

\bibitem{blei2006dynamic}
D.~Blei and J.~Lafferty.
\newblock {Dynamic topic models}.
\newblock In {\em Proc. ICML}, 2006.

\bibitem{buhrmester2011amazon}
M.~Buhrmester, T.~Kwang, and S.~Gosling.
\newblock {Amazon's Mechanical Turk a new source of inexpensive, yet
  high-quality, data?}
\newblock {\em Perspectives on psychological science}, 6(1):3--5, 2011.

\bibitem{castrejon2016learning}
L.~Castrejon, Y.~Aytar, C.~Vondrick, H.~Pirsiavash, and A.~Torralba.
\newblock {Learning Aligned Cross-Modal Representations from Weakly Aligned
  Data}.
\newblock In {\em Proc. CVPR}, 2016.

\bibitem{cho2014learning}
K.~Cho, B.~V. Merri{\"e}nboer, C.~Gulcehre, D.~Bahdanau, F.~Bougares,
  H.~Schwenk, and Y.~Bengio.
\newblock {Learning phrase representations using RNN encoder-decoder for
  statistical machine translation}.
\newblock In {\em Proc. of EMNLP}. Association for Computational Linguistics,
  2014.

\bibitem{dodge2015evaluating}
J.~Dodge, A.~Gane, X.~Zhang, A.~Bordes, S.~Chopra, A.~Miller, A.~Szlam, and
  J.~Weston.
\newblock {Evaluating prerequisite qualities for learning end-to-end dialog
  systems}.
\newblock In {\em Proc. of ICLR}, 2016.

\bibitem{escalante2010segmented}
H.~Escalante, C.~Hern{\'a}ndez, J.~Gonzalez, A.~L{\'o}pez-L{\'o}pez, M.~Montes,
  E.~Morales, E.~Sucar, L.~Villase{\~n}, and M.~Grubinger.
\newblock {The segmented and annotated IAPR TC-12 benchmark}.
\newblock {\em CVIU}, 2010.

\bibitem{fei2006one}
L.~Fei-Fei, R.~Fergus, and P.~Perona.
\newblock {One-shot learning of object categories}.
\newblock {\em IEEE transactions on pattern analysis and machine intelligence},
  2006.

\bibitem{geman2015visual}
D.~Geman, S.~Geman, N.~Hallonquist, and L.~Younes.
\newblock {Visual turing test for computer vision systems}.
\newblock {\em Proceedings of the National Academy of Sciences},
  112(12):3618--3623, 2015.

\bibitem{Goodfellow-et-al-2016-Book}
I.~Goodfellow, Y.~Bengio, and A.~Courville.
\newblock {Deep Learning}.
\newblock Book in preparation for MIT Press, 2016.

\bibitem{hochreiter1997long}
S.~Hochreiter and J.~Schmidhuber.
\newblock {Long short-term memory}.
\newblock {\em Neural computation}, 9(8):1735--1780, 1997.

\bibitem{hu2016natural}
R.~Hu, H.~Xu, M.~Rohrbach, J.~Feng, K.~Saenko, and T.~Darrell.
\newblock {Natural Language Object Retrieval}.
\newblock {\em Proc. of CVPR}, 2016.

\bibitem{DBLP:journals/corr/JabriJM16}
A.~Jabri, A.~Joulin, and L.~van~der Maaten.
\newblock {Revisiting Visual Question Answering Baselines}.
\newblock In {\em Proc of ECCV}, 2016.

\bibitem{karpathy2015deep}
A.~Karpathy and L.~Fei-Fei.
\newblock {Deep visual-semantic alignments for generating image descriptions}.
\newblock In {\em Proc. CVPR}, 2015.

\bibitem{kazemzadeh2014referitgame}
S.~Kazemzadeh, V.~Ordonez, M.~Matten, and T.~Berg.
\newblock {ReferItGame: Referring to Objects in Photographs of Natural Scenes}.
\newblock In {\em Proc. of EMNLP}, 2014.

\bibitem{DBLP:journals/corr/KingmaB14}
D.~P. Kingma and J.~Ba.
\newblock {Adam: {A} Method for Stochastic Optimization}.
\newblock {\em CoRR}, abs/1412.6980, 2014.

\bibitem{krahmer2012computational}
E.~Krahmer and K.~V. Deemter.
\newblock {Computational generation of referring expressions: A survey}.
\newblock {\em Computational Linguistics}, 38(1):173--218, 2012.

\bibitem{lecun2015deep}
Y.~LeCun, Y.~Bengio, and G.~Hinton.
\newblock {Deep learning}.
\newblock {\em Nature}, 521(7553):436--444, 2015.

\bibitem{levin1997stochastic}
E.~Levin and R.~Pieraccini.
\newblock A stochastic model of computer-human interaction for learning
  dialogue strategies.
\newblock In {\em Eurospeech}, volume~97, pages 1883--1886, 1997.

\bibitem{lin2014microsoft}
T.~Lin, M.~Maire, S.~Belongie, J.~Hays, P.~Perona, D.~Ramanan, P.~Doll{\'a}r,
  and L.~Zitnick.
\newblock {Microsoft coco: Common objects in context}.
\newblock In {\em Proc of ECCV}, 2014.

\bibitem{liu2016not}
C.~Liu, R.~Lowe, I.~Serban, M.~Noseworthy, L.~Charlin, and J.~Pineau.
\newblock {How NOT to evaluate your dialogue system: An empirical study of
  unsupervised evaluation metrics for dialogue response generation}.
\newblock {\em arXiv preprint arXiv:1603.08023}, 2016.

\bibitem{lu2016hierarchical}
J.~Lu, J.~Yang, D.~Batra, and D.~Parikh.
\newblock {Hierarchical Question-Image Co-Attention for Visual Question
  Answering}.
\newblock {\em arXiv preprint arXiv:1606.00061}, 2016.

\bibitem{malinowski2014multi}
M.~Malinowski and M.~Fritz.
\newblock {A multi-world approach to question answering about real-world scenes
  based on uncertain input}.
\newblock In {\em Proc. of NIPS}, pages 1682--1690, 2014.

\bibitem{mao2015generation}
J.~Mao, J.~Huang, A.~Toshev, O.~Camburu, A.~Yuille, and K.~Murphy.
\newblock Generation and comprehension of unambiguous object descriptions.
\newblock {\em arXiv preprint arXiv:1511.02283}, 2015.

\bibitem{lemon2012conversational}
{O. Lemon and O. Pietquin}, editor.
\newblock {\em {Data-Driven Methods for Adaptive Spoken Dialogue Systems}}.
\newblock Springer, 2012.

\bibitem{pietquin2006probabilistic}
O.~Pietquin and T.~Dutoit.
\newblock {A probabilistic framework for dialog simulation and optimal strategy
  learning}.
\newblock {\em IEEE Transactions on Audio, Speech, and Language Processing},
  2006.

\bibitem{pietquin2013survey}
O.~Pietquin and H.~Hastie.
\newblock {A survey on metrics for the evaluation of user simulations}.
\newblock {\em The knowledge engineering review}, 28(01):59--73, 2013.

\bibitem{rehurek_lrec}
R.~{\v R}eh{\r u}{\v r}ek and P.~Sojka.
\newblock {Software Framework for Topic Modelling with Large Corpora}.
\newblock In {\em Proc. LREC 2010 Workshop on New Challenges for NLP
  Frameworks}, 2010.

\bibitem{ILSVRC15}
O.~Russakovsky, J.~Deng, H.~Su, J.~Krause, S.~Satheesh, S.~Ma, Z.~Huang,
  A.~Karpathy, A.~Khosla, M.~Bernstein, et~al.
\newblock {Imagenet large scale visual recognition challenge}.
\newblock {\em International Journal of Computer Vision}, 115(3):211--252,
  2015.

\bibitem{schatzmann2006survey}
J.~Schatzmann, K.~Weilhammer, M.~Stuttle, and S.~Young.
\newblock {A survey of statistical user simulation techniques for
  reinforcement-learning of dialogue management strategies}.
\newblock {\em The knowledge engineering review}, 21(02):97--126, 2006.

\bibitem{serban2015survey}
I.~Serban, R.~Lowe, L.~Charlin, and J.~Pineau.
\newblock {A survey of available corpora for building data-driven dialogue
  systems}.
\newblock {\em arXiv preprint arXiv:1512.05742}, 2015.

\bibitem{DBLP:journals/corr/SerbanSBCP15}
I.~Serban, A.~Sordoni, Y.~Bengio, A.~Courville, and J.~Pineau.
\newblock {Hierarchical neural network generative models for movie dialogues}.
\newblock {\em arXiv preprint arXiv:1507.04808}, 2015.

\bibitem{shih2015look}
K.~Shih, S.~Singh, and D.~Hoiem.
\newblock {Where to look: Focus regions for visual question answering}.
\newblock In {\em Proc. of CVPR}, 2016.

\bibitem{singh1999reinforcement}
S.~Singh, M.~Kearns, D.~Litman, and M.~Walker.
\newblock {Reinforcement Learning for Spoken Dialogue Systems}.
\newblock In {\em Proc. of NIPS}, 1999.

\bibitem{sutskever2014sequence}
I.~Sutskever, O.~Vinyals, and Q.~Le.
\newblock {Sequence to sequence learning with neural networks}.
\newblock In {\em Proc of NIPS}, 2014.

\bibitem{vinyals2015show}
O.~Vinyals, A.~Toshev, S.~Bengio, and D.~Erhan.
\newblock {Show and tell: A neural image caption generator}.
\newblock In {\em Proc. of CVPR}, 2015.

\bibitem{wen2016network}
T.~Wen, M.~Gasic, N.~Mrksic, L.~Rojas-Barahona, P.~Su, S.~Ultes, D.~Vandyke,
  and S.~Young.
\newblock {A Network-based End-to-End Trainable Task-oriented Dialogue System}.
\newblock {\em arXiv preprint arXiv:1604.04562}, 2016.

\bibitem{weston2015towards}
J.~Weston, A.~Bordes, S.~Chopra, A.~Rush, B.~van Merri{\"e}nboer, A.~Joulin,
  and T.~Mikolov.
\newblock {Towards ai-complete question answering: A set of prerequisite toy
  tasks}.
\newblock In {\em Proc. of ICLR}, 2016.

\bibitem{DBLP:journals/corr/XuBKCCSZB15}
K.~Xu, J.~Ba, R.~Kiros, K.~Cho, A.~Courville, R.~Salakhutdinov, R.~Zemel, and
  Y.~Bengio.
\newblock {Show, attend and tell: Neural image caption generation with visual
  attention}.
\newblock 2015.

\bibitem{yang2015stacked}
Z.~Yang, X.~He, J.~Gao, L.~Deng, and A.~Smola.
\newblock {Stacked attention networks for image question answering}.
\newblock In {\em Proc. of CVPR}, 2016.

\bibitem{young2013pomdp}
S.~Young, M.~Ga{\v{s}}i{\'c}, B.~Thomson, and J.~Williams.
\newblock {POMDP-based statistical spoken dialog systems: A review}.
\newblock {\em Proc. of the IEEE}, 101(5):1160--1179, 2013.

\bibitem{DBLP:journals/corr/YuPYBB16}
L.~Yu, P.~Poirson, S.~Yang, A.~Berg, and T.~Berg.
\newblock {Modeling context in referring expressions}.
\newblock In {\em Proc. in ECCV}. Springer, 2016.

\bibitem{zhou2014learning}
B.~Zhou, A.~Lapedriza, J.~Xiao, A.~Torralba, and A.~Oliva.
\newblock {Learning deep features for scene recognition using places database}.
\newblock In {\em Proc of NIPS}, 2014.

\end{thebibliography}
}

\newpage

\appendix

\onecolumn

\section{User interface}
\label{ap:website}
\begin{figure}[h!]
\begin{subfigure}{0.32\linewidth}
\includegraphics[width=\linewidth]{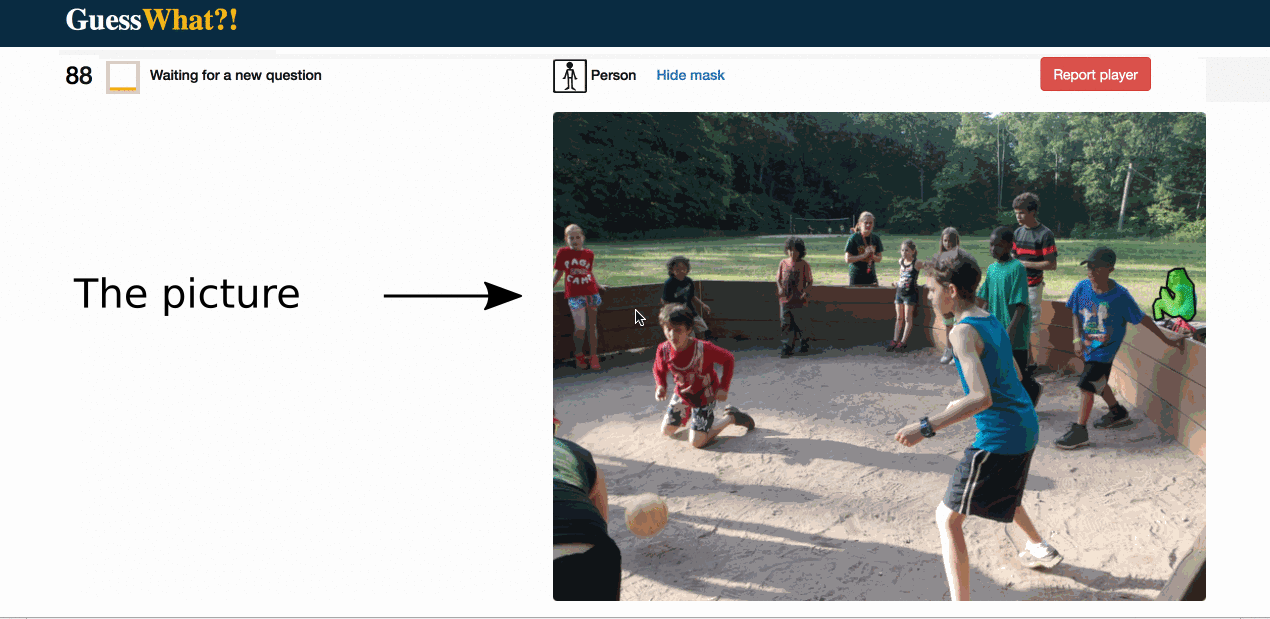}
\caption{}
\end{subfigure}
\begin{subfigure}{0.32\linewidth}
\includegraphics[width=\linewidth]{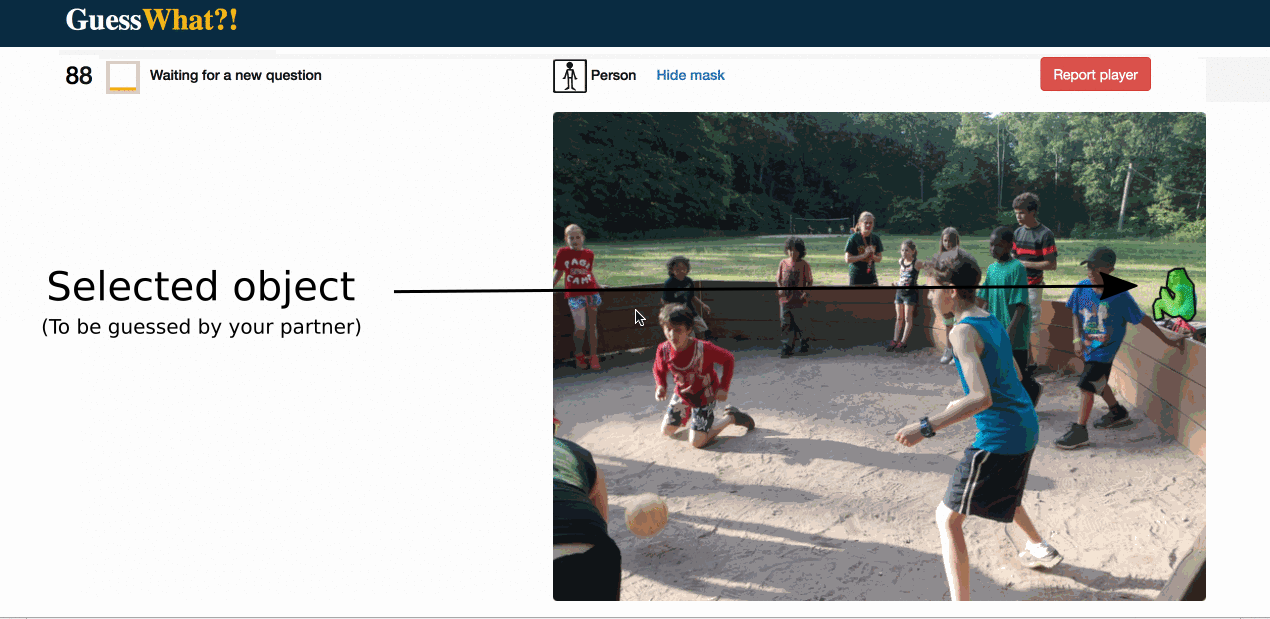}
\caption{}
\end{subfigure}
\begin{subfigure}{0.32\linewidth}
\includegraphics[width=\linewidth]{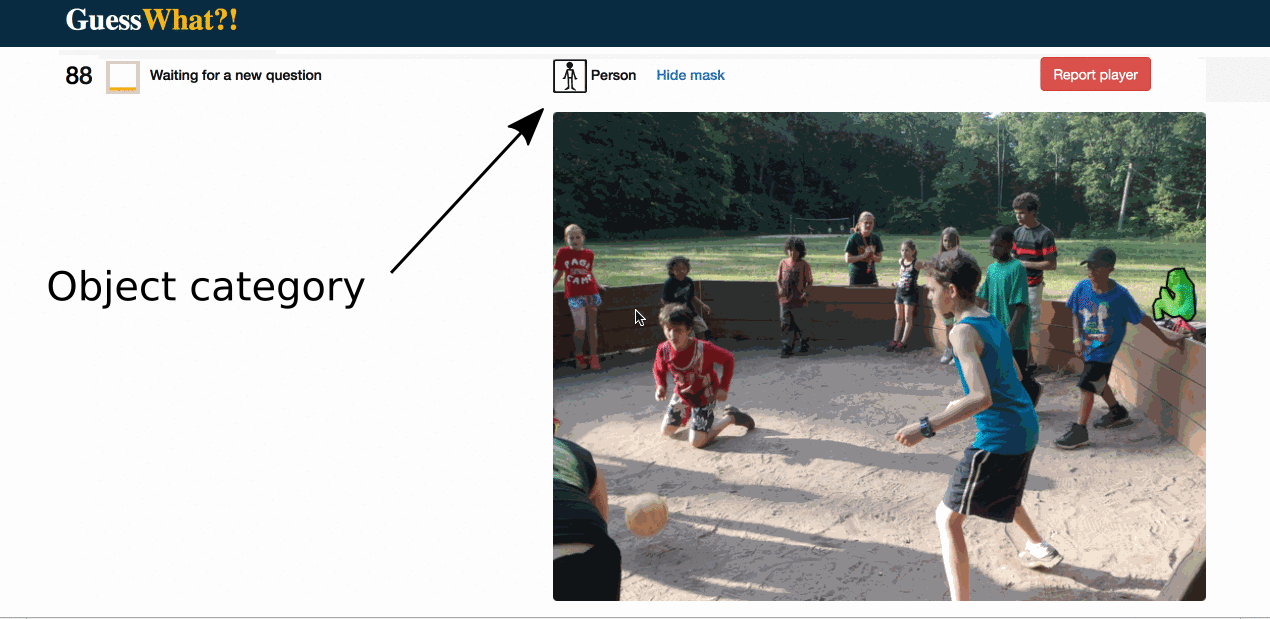}
\caption{}
\end{subfigure}
\begin{subfigure}{0.32\linewidth}
\includegraphics[width=\linewidth]{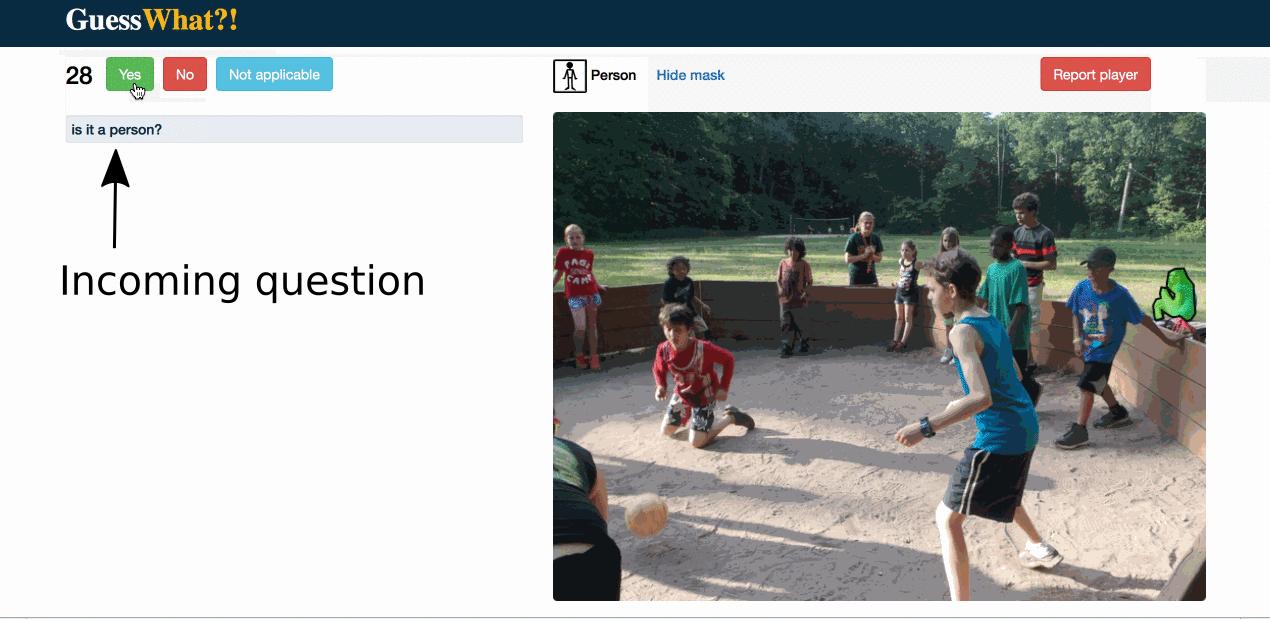}
\caption{}
\end{subfigure}
\begin{subfigure}{0.32\linewidth}
\includegraphics[width=\linewidth]{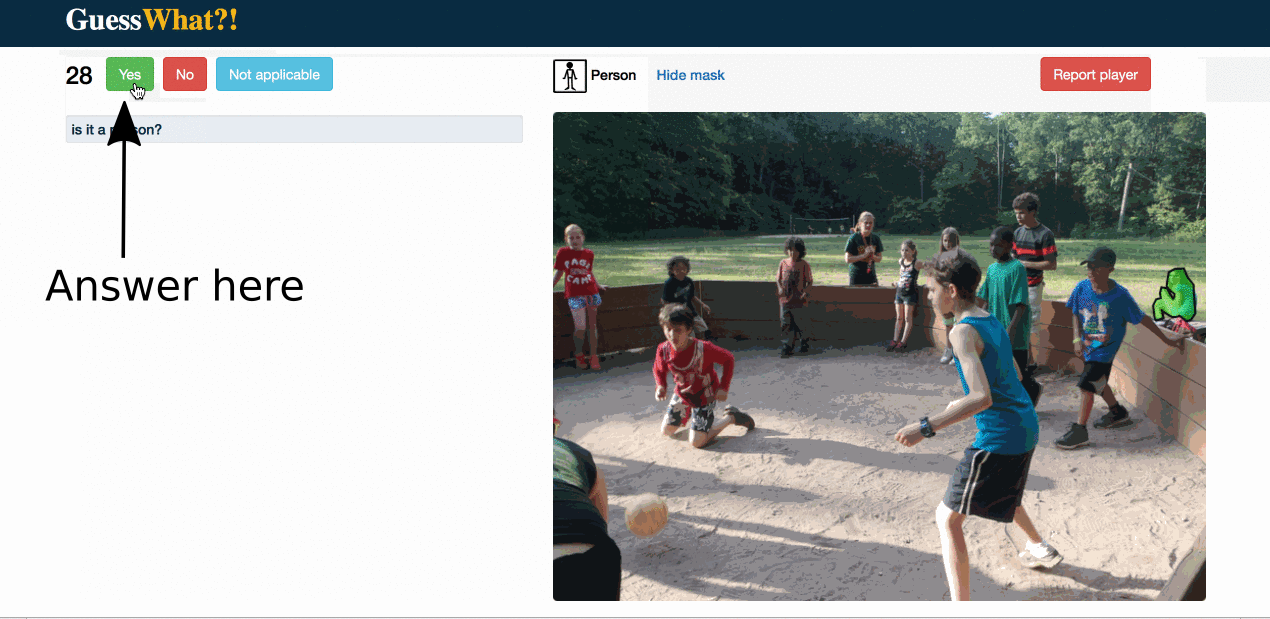}
\caption{}
\end{subfigure}
\begin{subfigure}{0.32\linewidth}
\includegraphics[width=\linewidth]{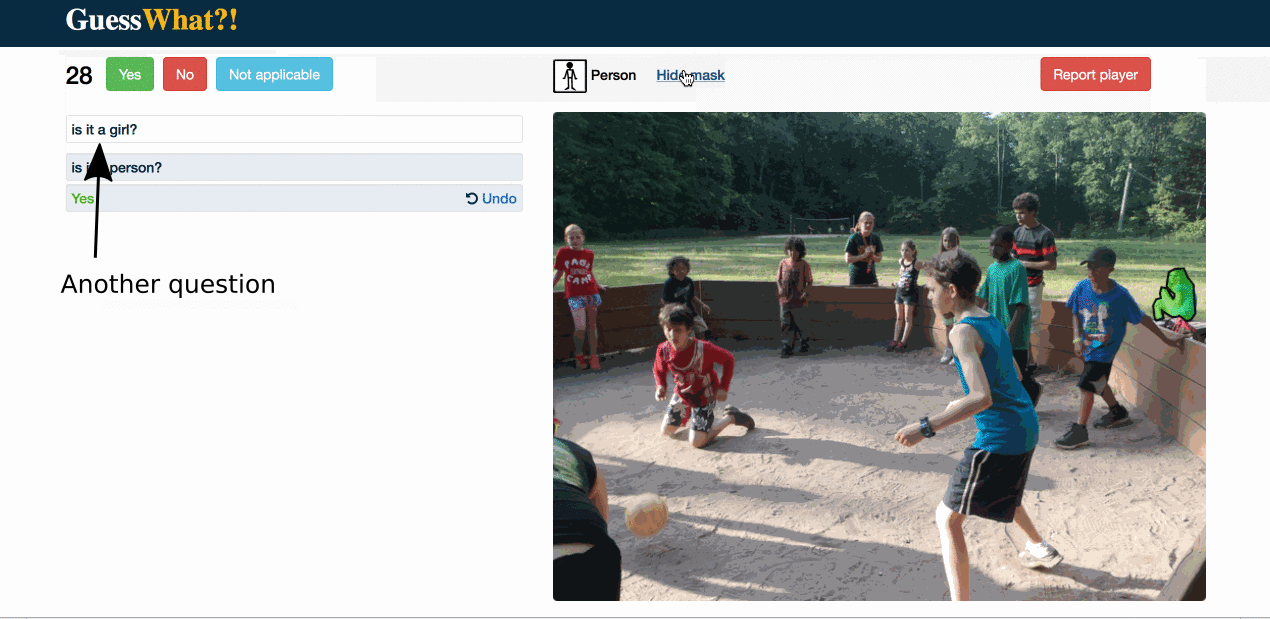}
\caption{}
\end{subfigure}
\begin{subfigure}{0.32\linewidth}
\includegraphics[width=\linewidth]{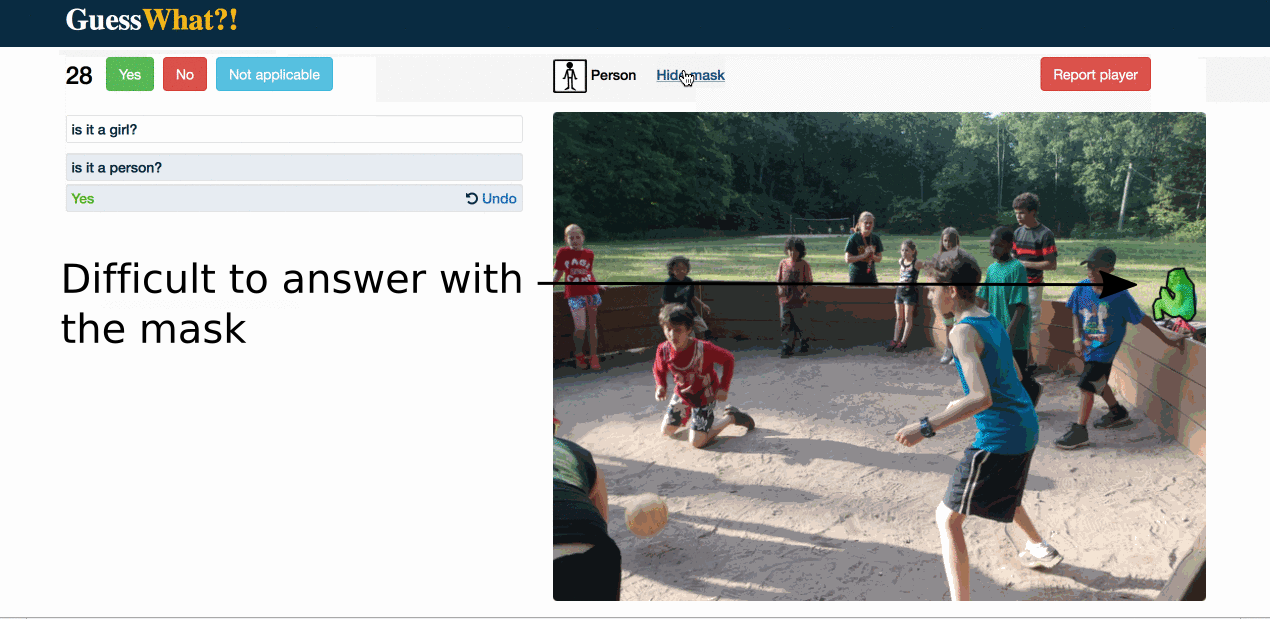}
\caption{}
\end{subfigure}
\begin{subfigure}{0.32\linewidth}
\includegraphics[width=\linewidth]{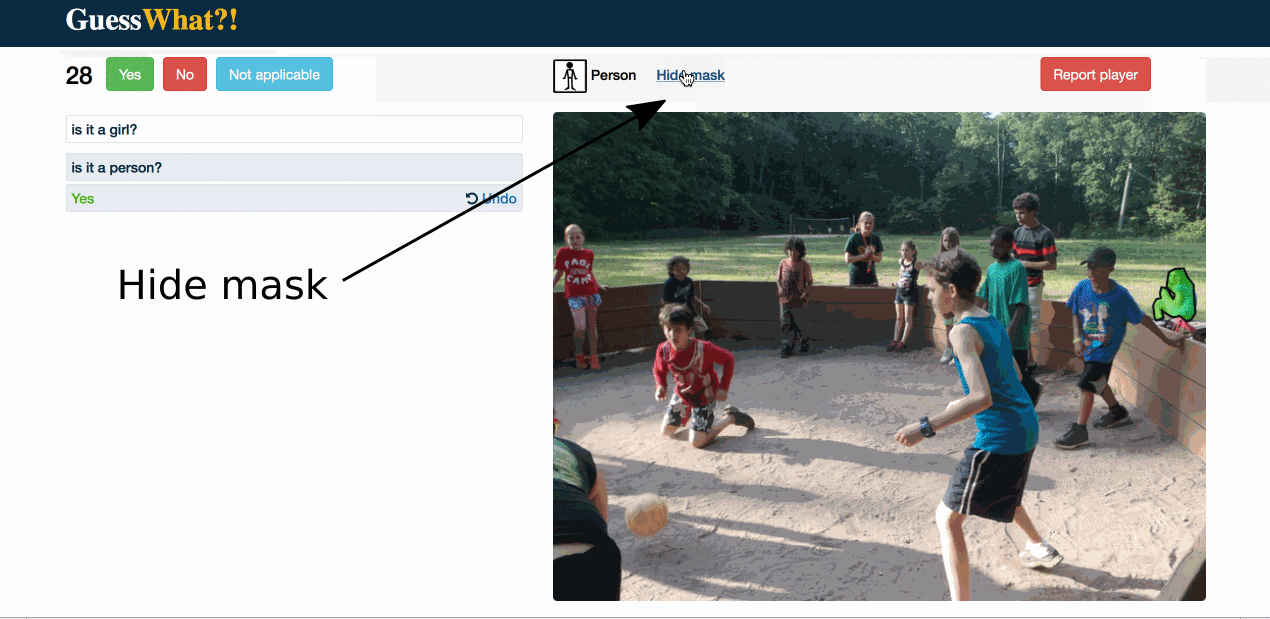}
\caption{}
\end{subfigure}
\begin{subfigure}{0.32\linewidth}
\includegraphics[width=\linewidth]{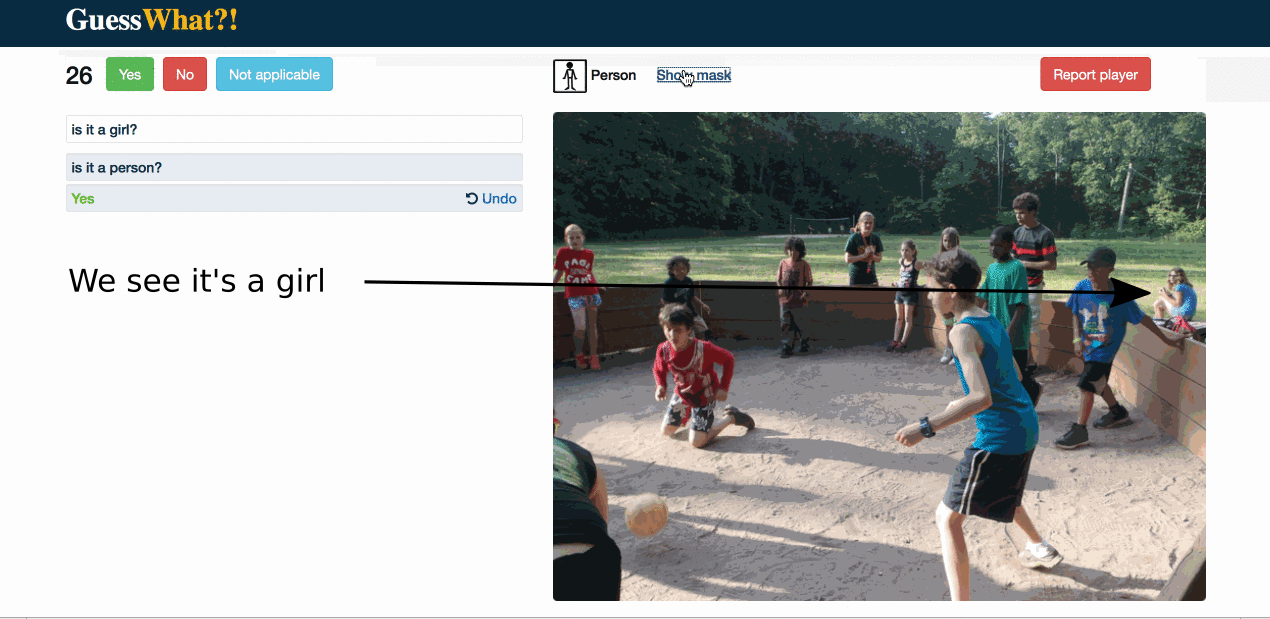}
\caption{}
\end{subfigure}
\begin{subfigure}{0.32\linewidth}
\includegraphics[width=\linewidth]{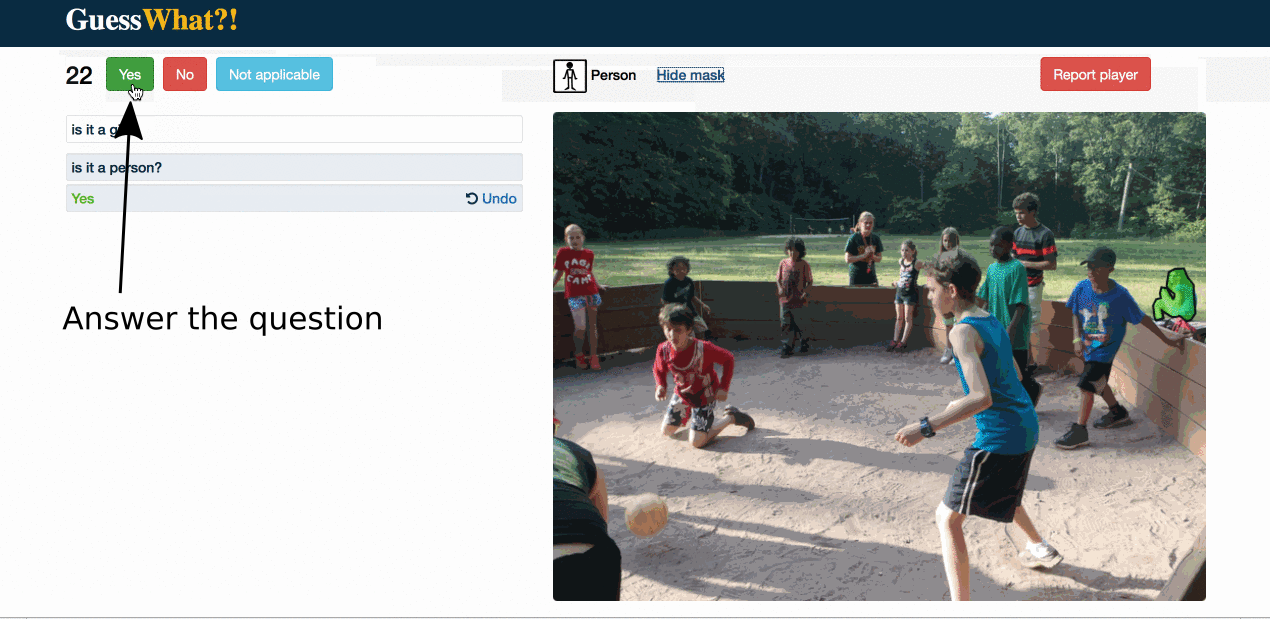}
\caption{}
\end{subfigure}
\begin{subfigure}{0.32\linewidth}
\includegraphics[width=\linewidth]{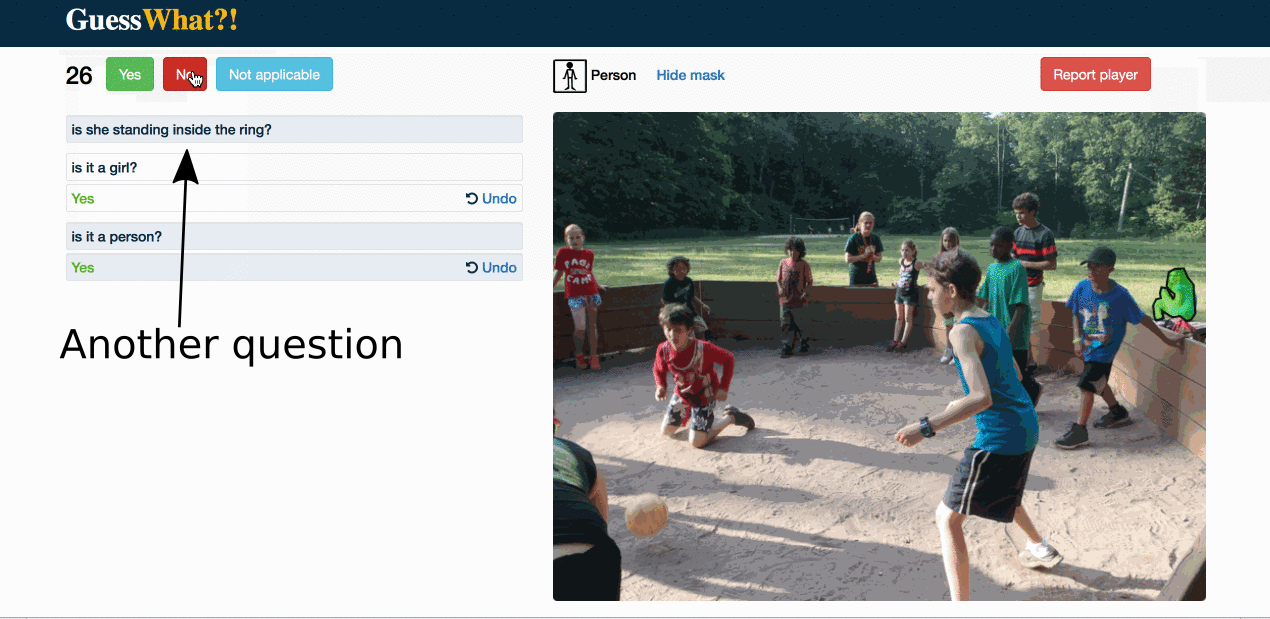}
\caption{}
\end{subfigure}
\begin{subfigure}{0.32\linewidth}
\includegraphics[width=\linewidth]{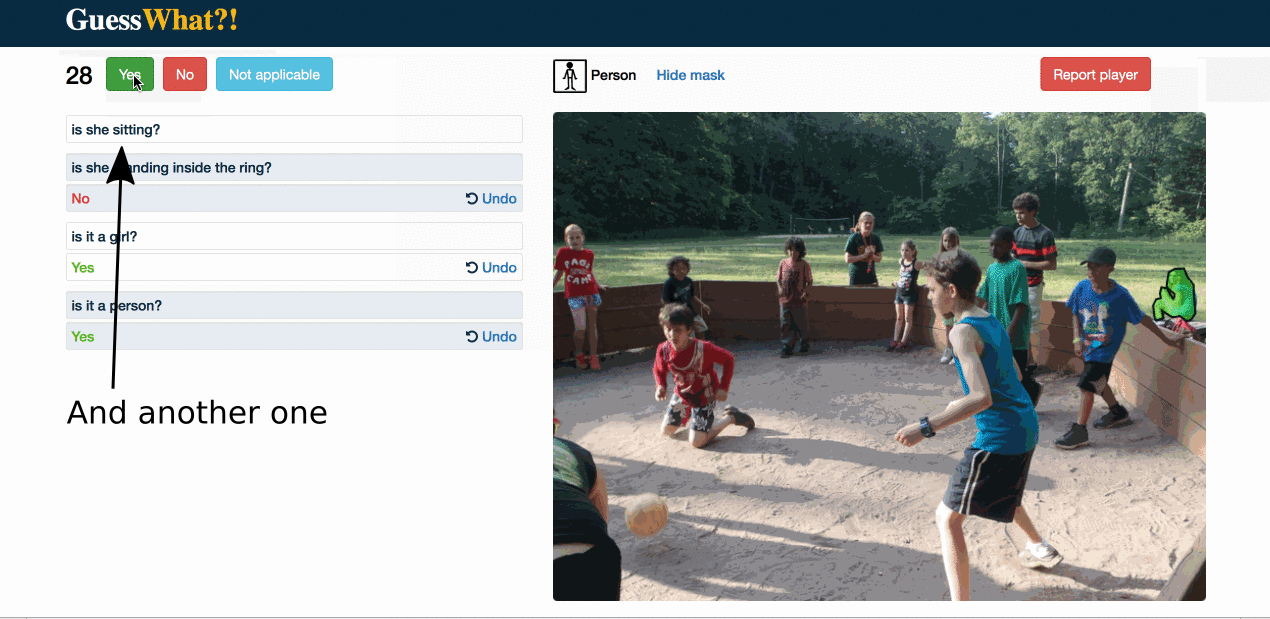}
\caption{}
\end{subfigure}
\begin{subfigure}{0.32\linewidth}
\includegraphics[width=\linewidth]{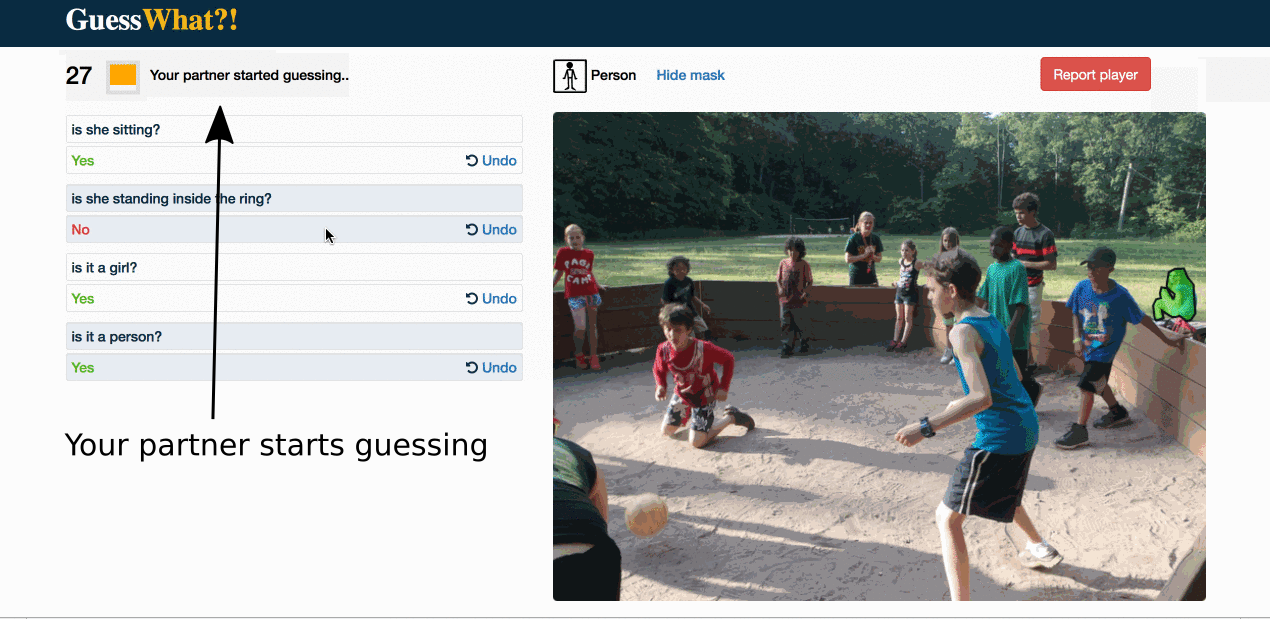}
\caption{}
\end{subfigure}
\begin{subfigure}{0.32\linewidth}
\includegraphics[width=\linewidth]{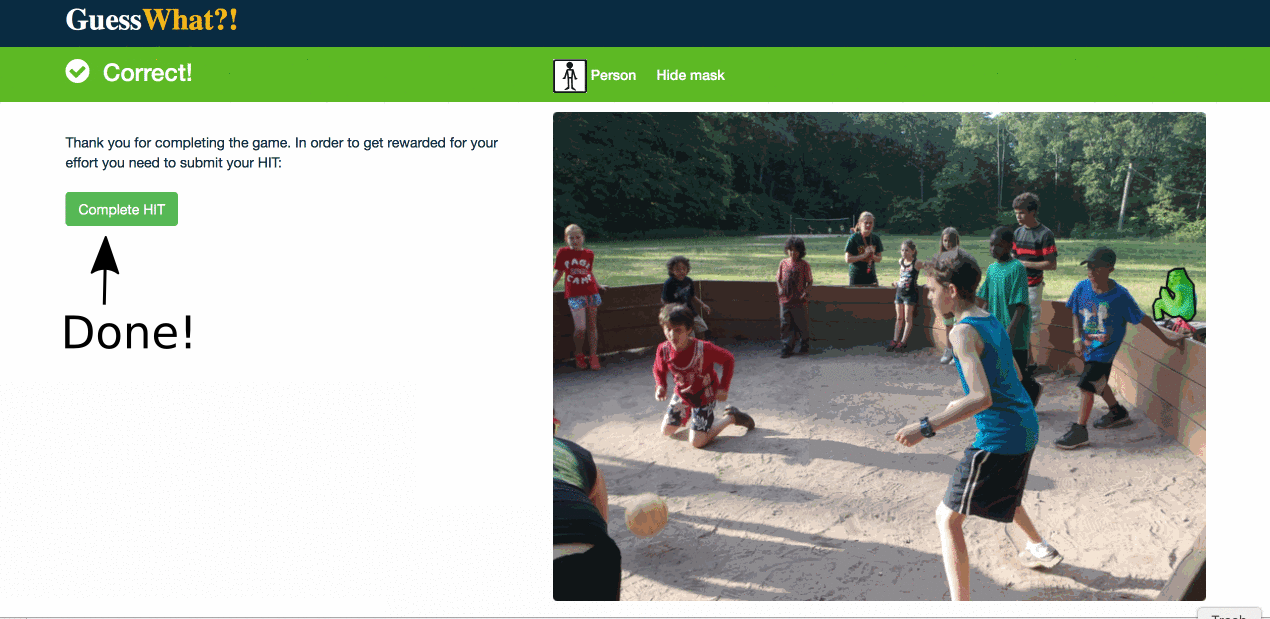}
\caption{}
\end{subfigure}
\caption{An example game from the perspective of the oracle. Shown from left to right and top to bottom. }
\label{fig:oracle}
\end{figure}

\begin{figure}[h!]
\begin{subfigure}{0.32\linewidth}
\includegraphics[width=\linewidth]{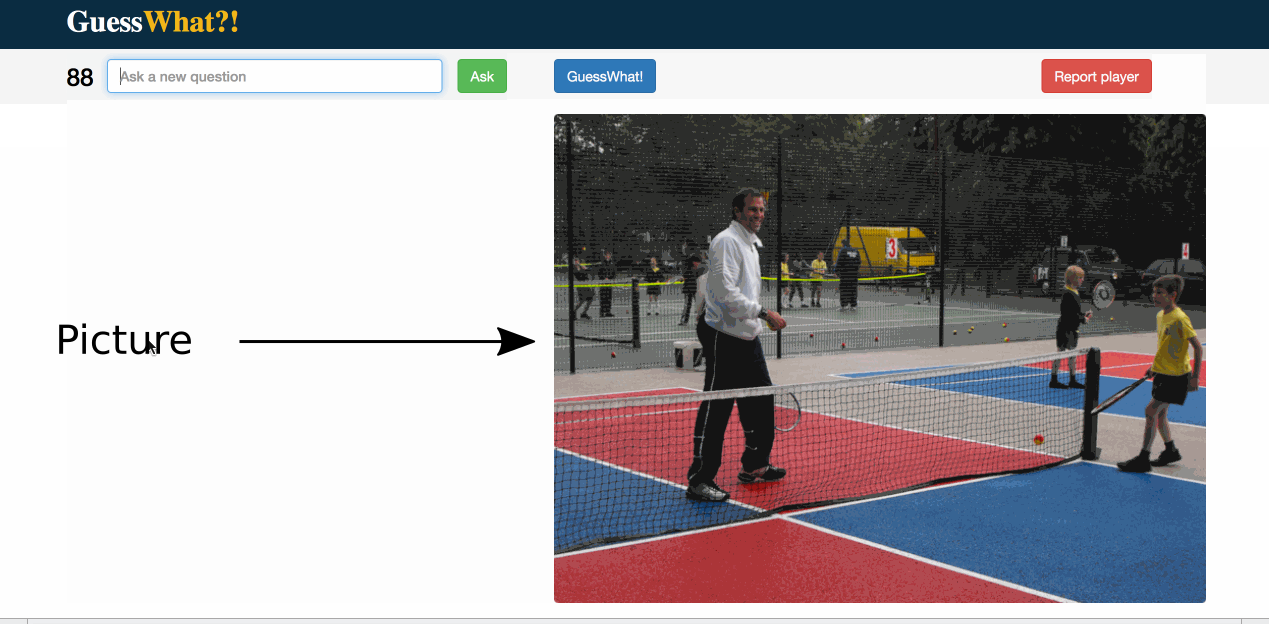}
\caption{}
\end{subfigure}
\begin{subfigure}{0.32\linewidth}
\includegraphics[width=\linewidth]{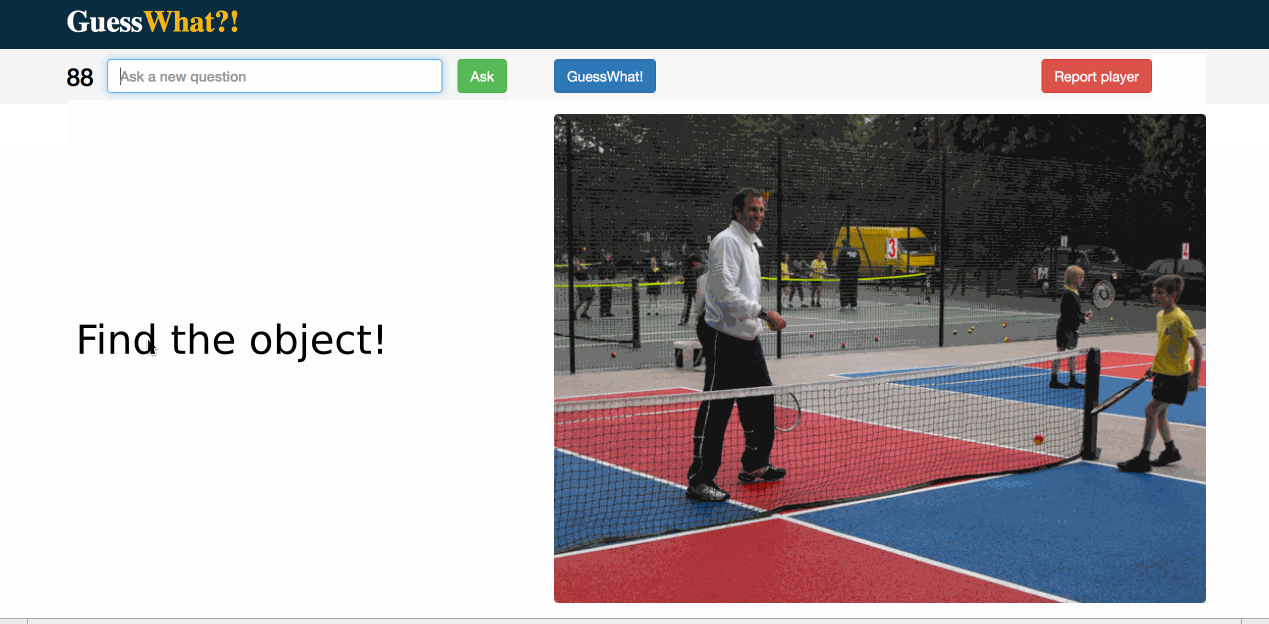}
\caption{}
\end{subfigure}
\begin{subfigure}{0.32\linewidth}
\includegraphics[width=\linewidth]{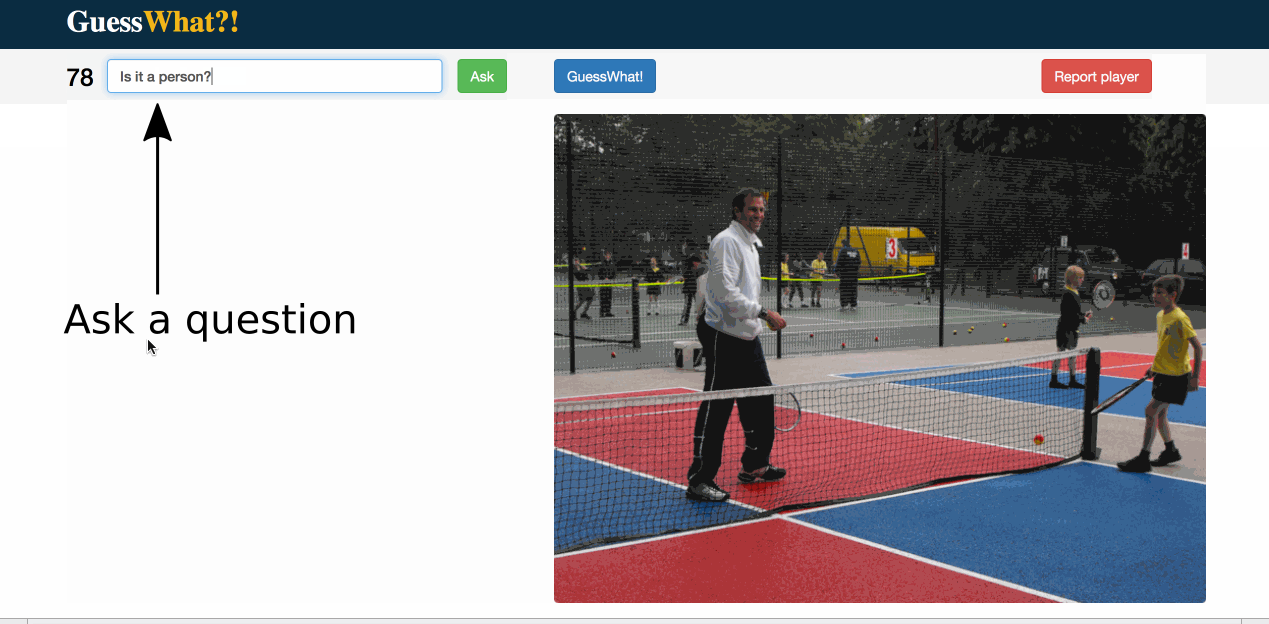}
\caption{}
\end{subfigure}
\begin{subfigure}{0.32\linewidth}
\includegraphics[width=\linewidth]{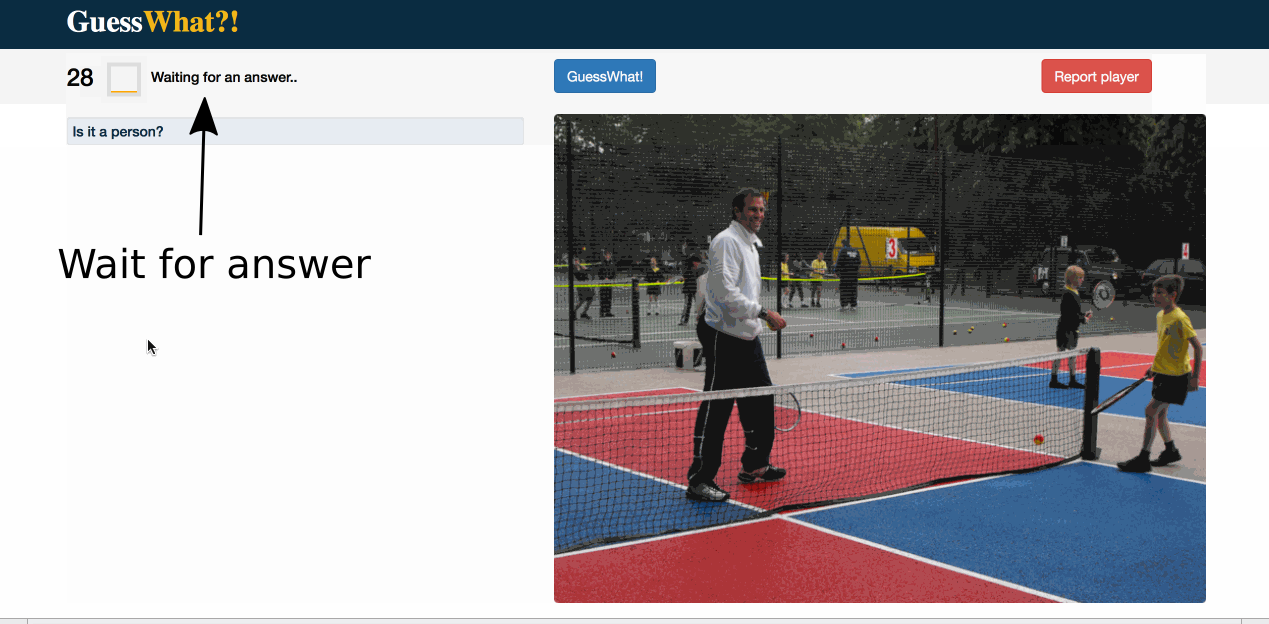}
\caption{}
\end{subfigure}
\begin{subfigure}{0.32\linewidth}
\includegraphics[width=\linewidth]{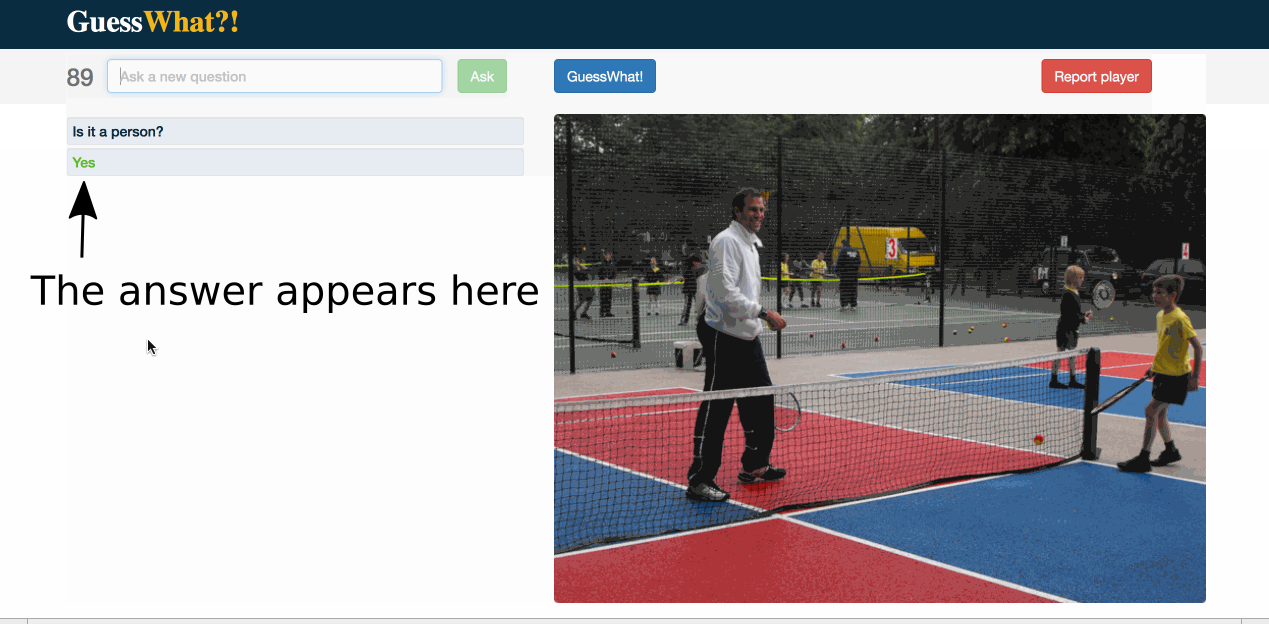}
\caption{}
\end{subfigure}
\begin{subfigure}{0.32\linewidth}
\includegraphics[width=\linewidth]{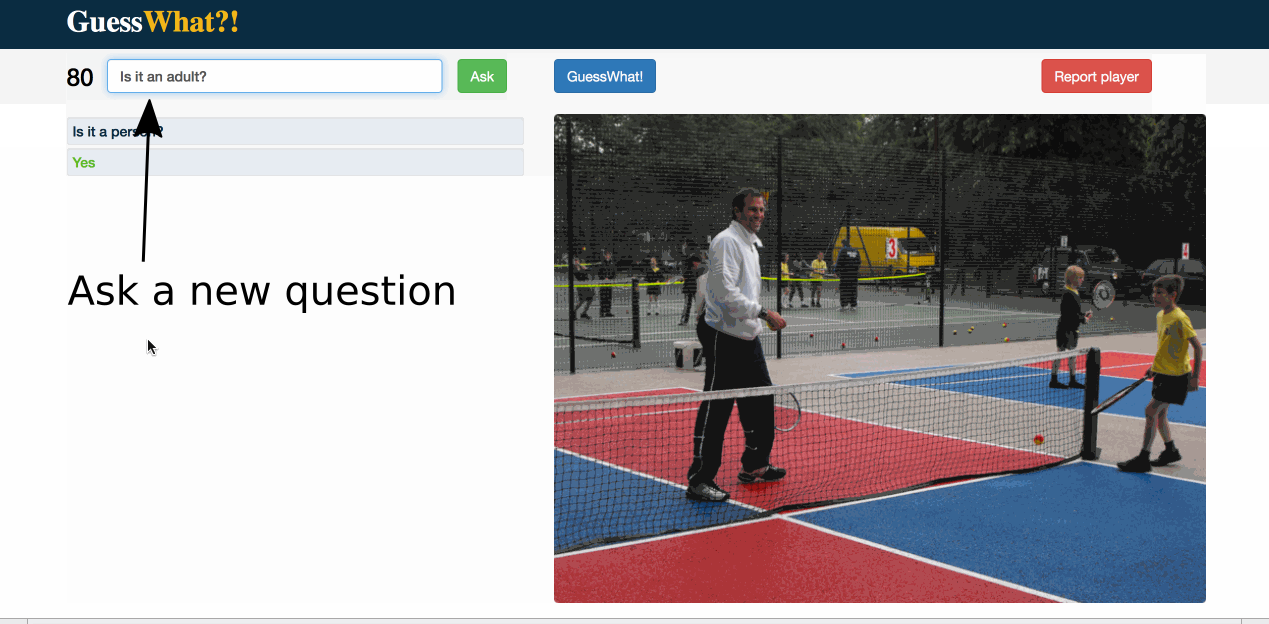}
\caption{}
\end{subfigure}
\begin{subfigure}{0.32\linewidth}
\includegraphics[width=\linewidth]{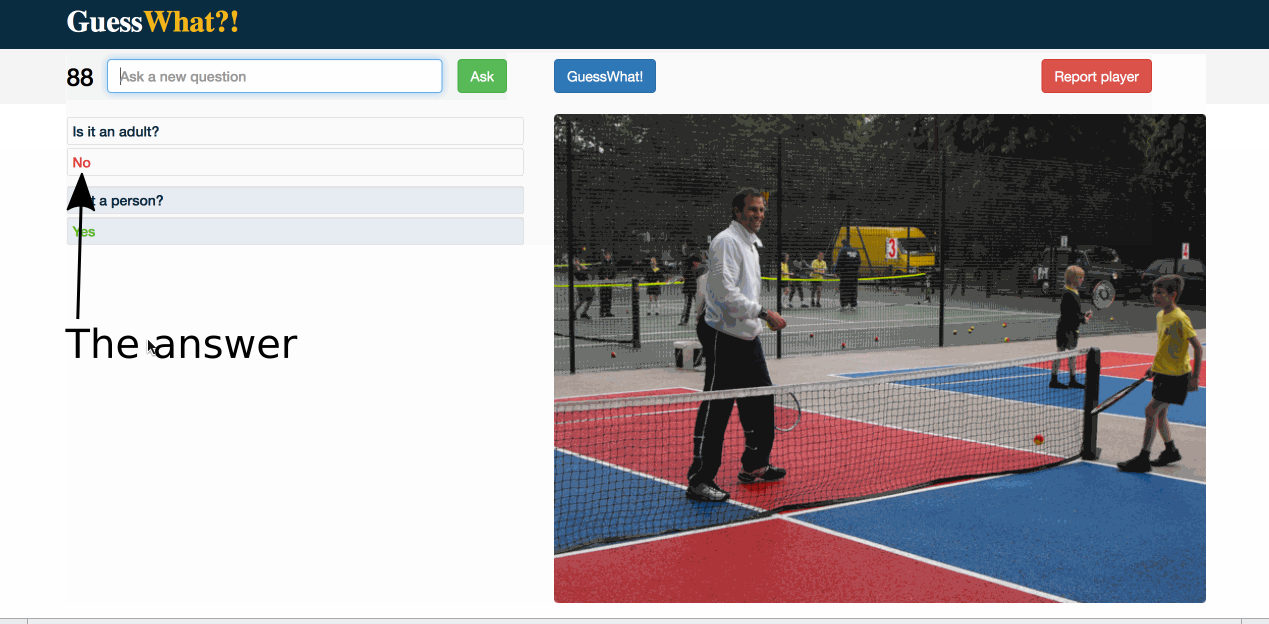}
\caption{}
\end{subfigure}
\begin{subfigure}{0.32\linewidth}
\includegraphics[width=\linewidth]{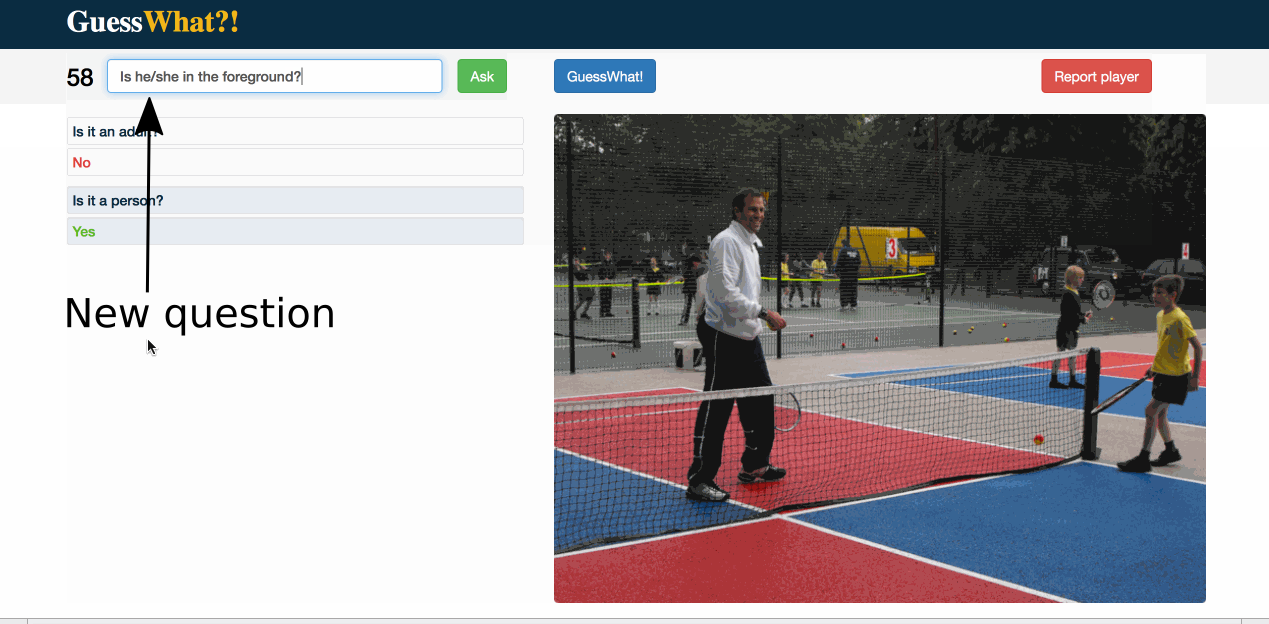}
\caption{}
\end{subfigure}
\begin{subfigure}{0.32\linewidth}
\includegraphics[width=\linewidth]{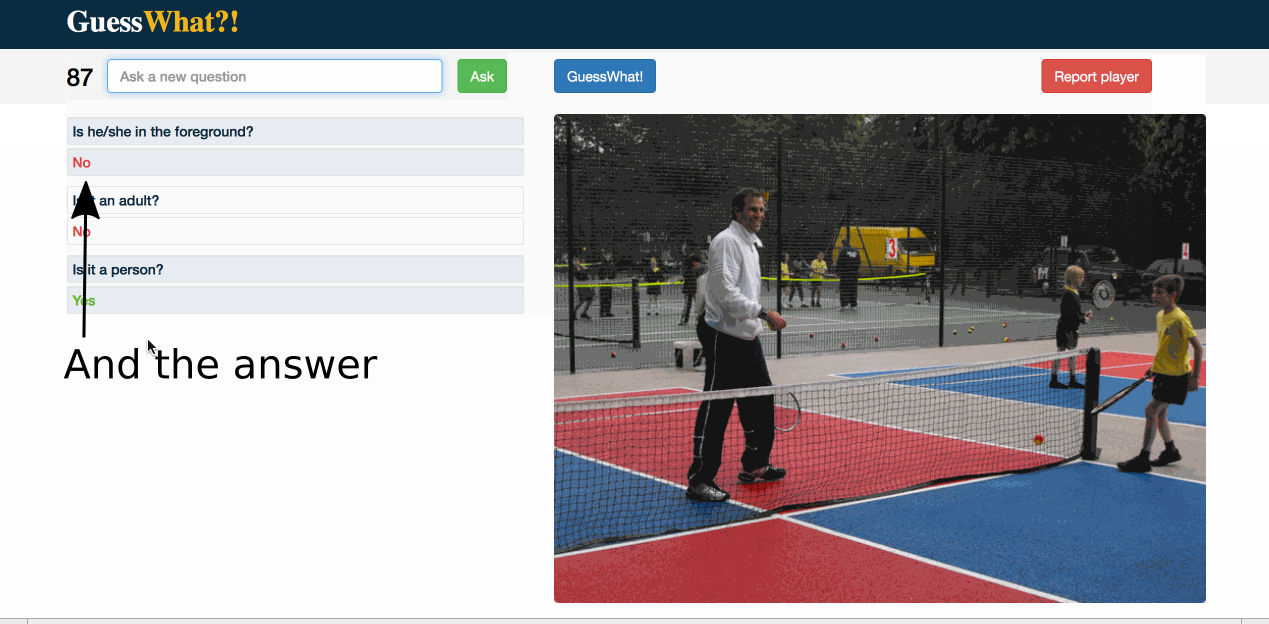}
\caption{}
\end{subfigure}
\begin{subfigure}{0.32\linewidth}
\includegraphics[width=\linewidth]{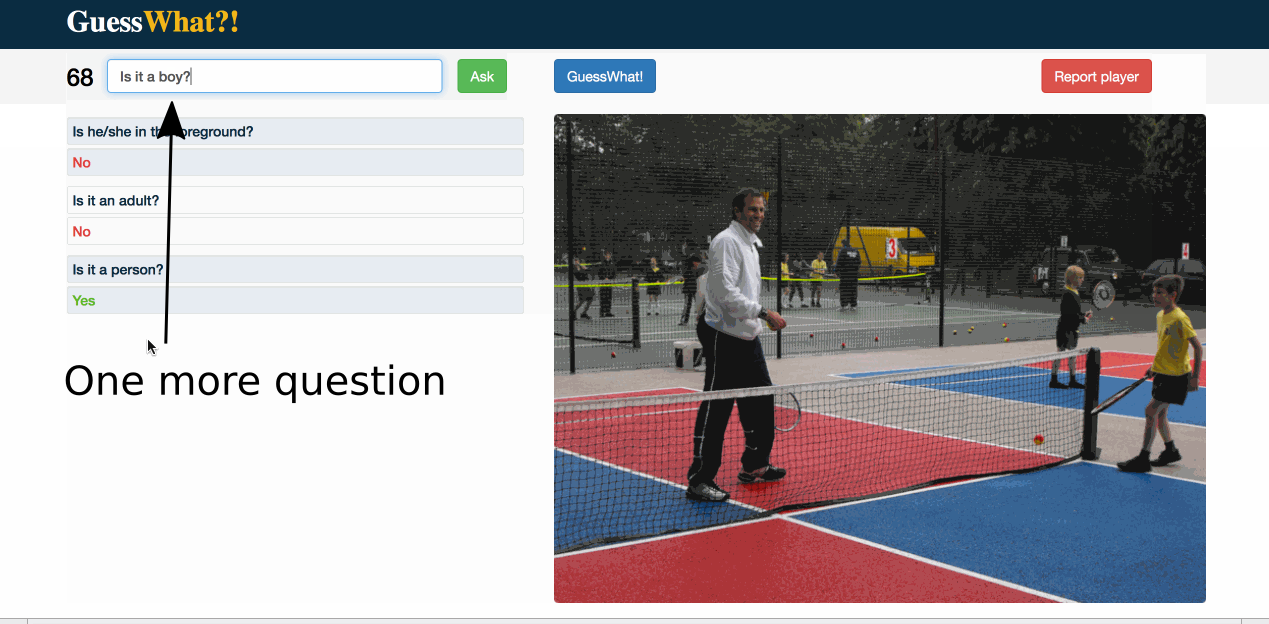}
\caption{}
\end{subfigure}
\begin{subfigure}{0.32\linewidth}
\includegraphics[width=\linewidth]{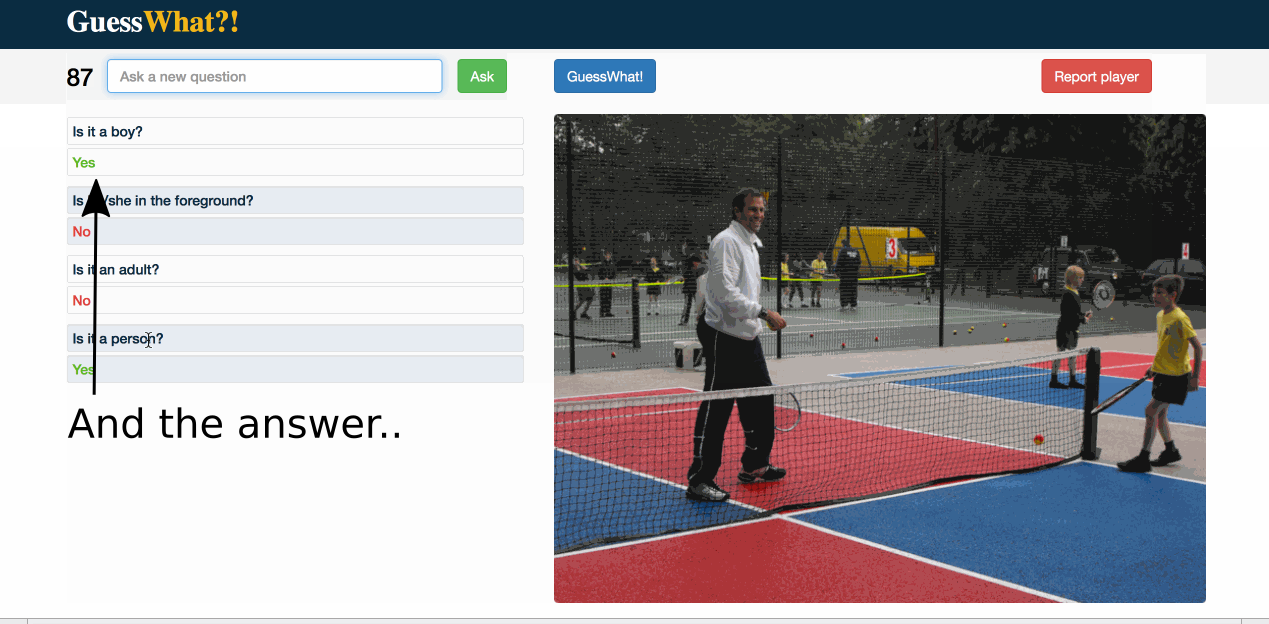}
\caption{}
\end{subfigure}
\begin{subfigure}{0.32\linewidth}
\includegraphics[width=\linewidth]{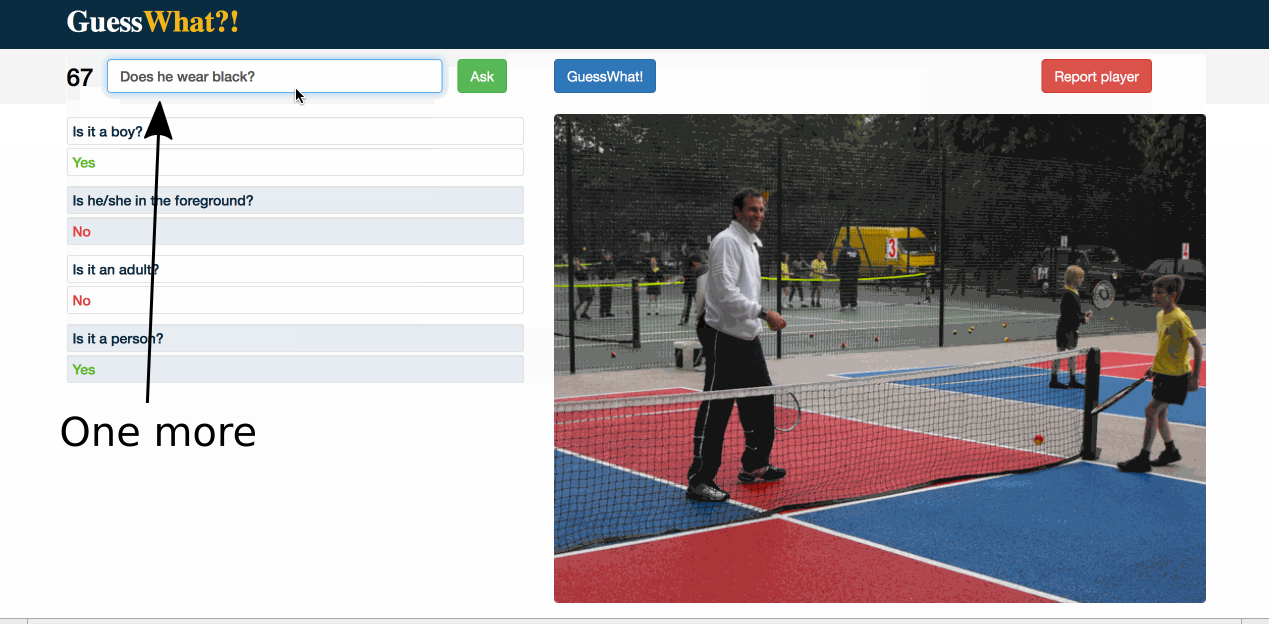}
\caption{}
\end{subfigure}
\begin{subfigure}{0.32\linewidth}
\includegraphics[width=\linewidth]{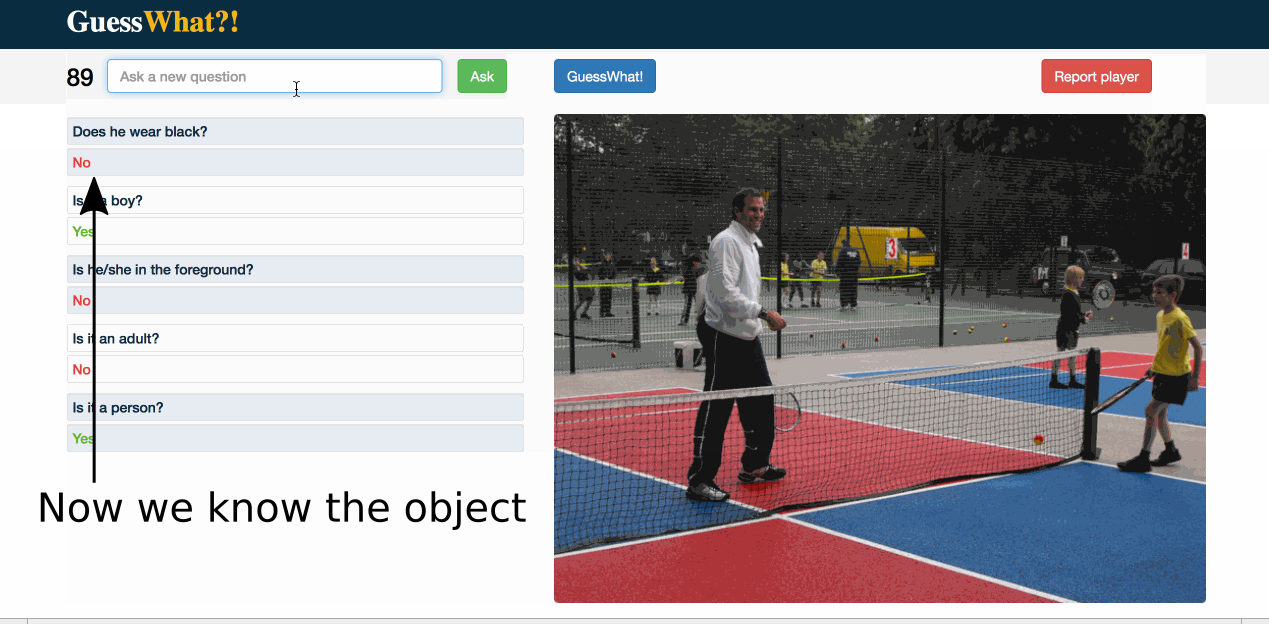}
\caption{}
\end{subfigure}
\begin{subfigure}{0.32\linewidth}
\includegraphics[width=\linewidth]{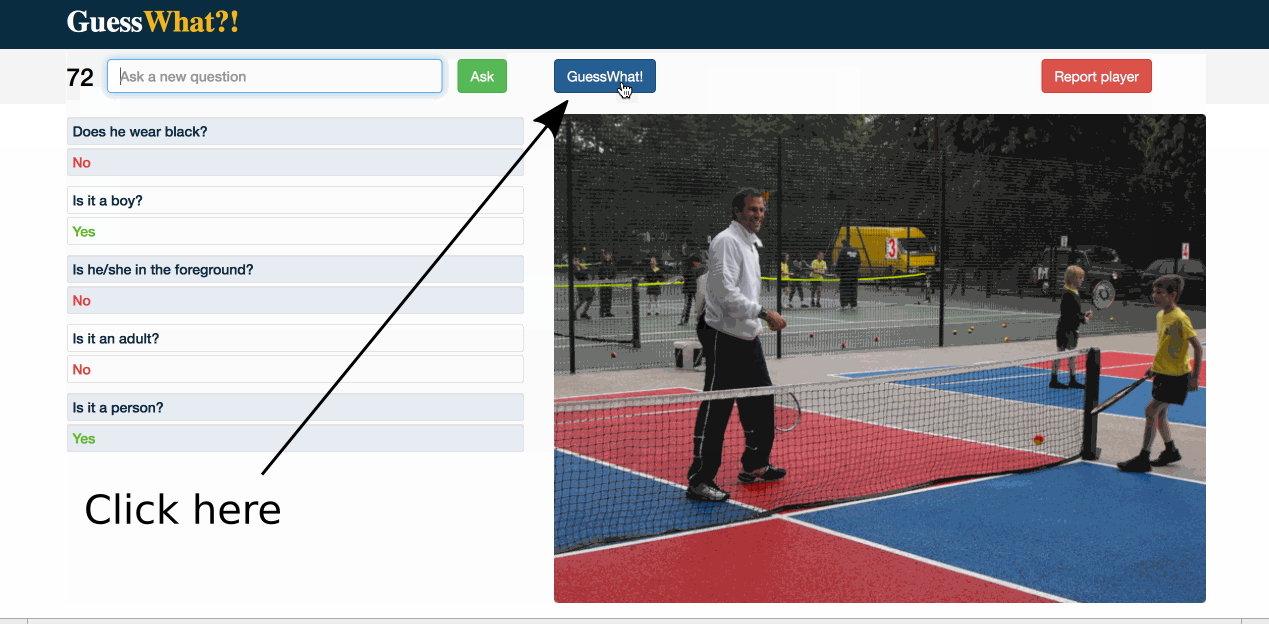}
\caption{}
\end{subfigure}
\begin{subfigure}{0.32\linewidth}
\includegraphics[width=\linewidth]{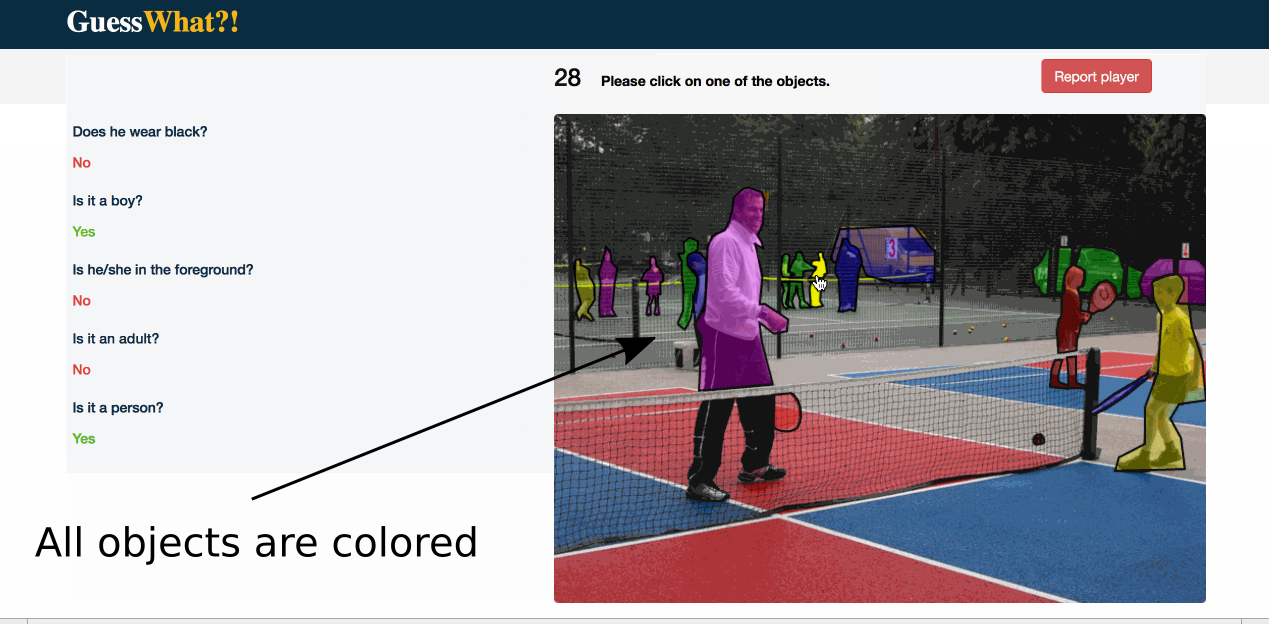}
\caption{}
\end{subfigure}\\
\begin{subfigure}{0.32\linewidth}
\includegraphics[width=\linewidth]{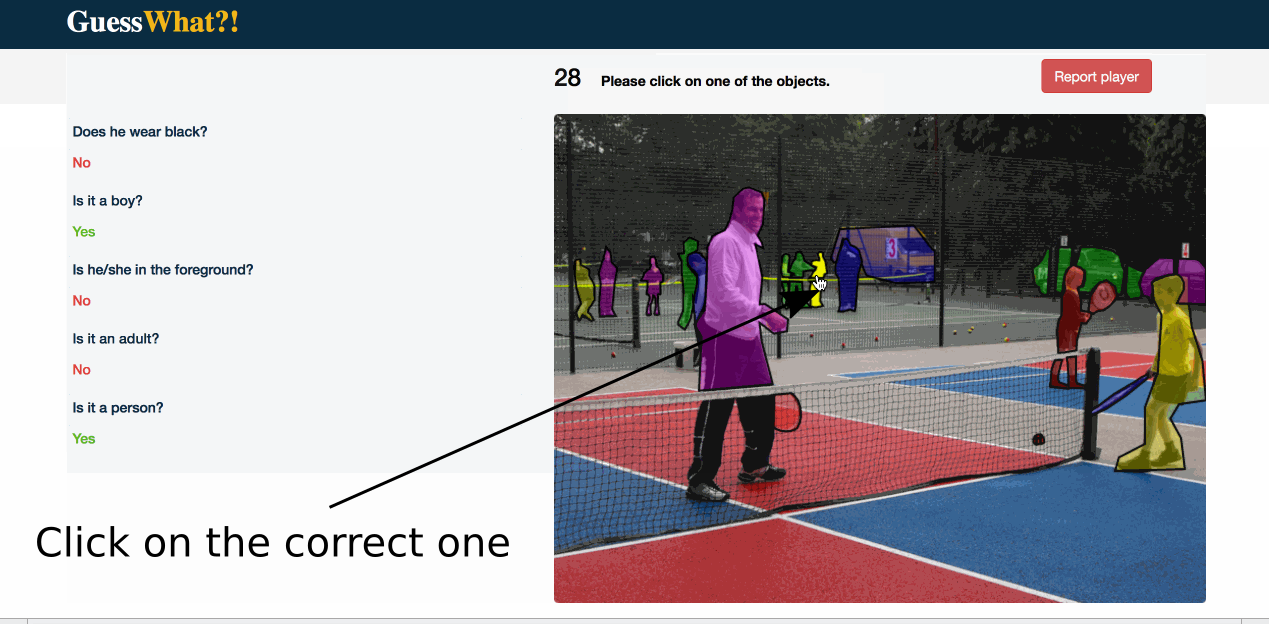}
\caption{}
\end{subfigure}
\begin{subfigure}{0.32\linewidth}
\includegraphics[width=\linewidth]{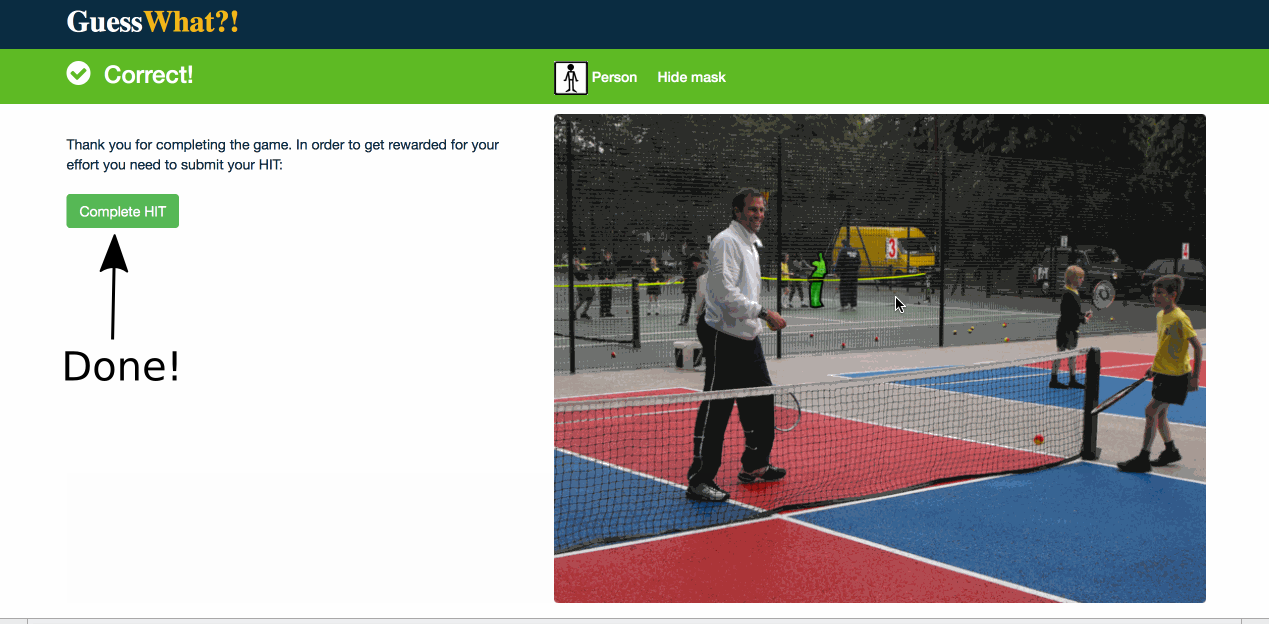}
\caption{}
\end{subfigure}
\caption{An example game from the perspective of the questioner. Shown from left to right and top to bottom.}
\label{fig:questioner}
\end{figure}

Figure \ref{fig:oracle}, \ref{fig:questioner} presents the instructions for the oracle and questioner before they started their first game. 

\begin{figure}[t!]
\centering
\begin{subfigure}{0.75\linewidth}
\includegraphics[width=\linewidth]{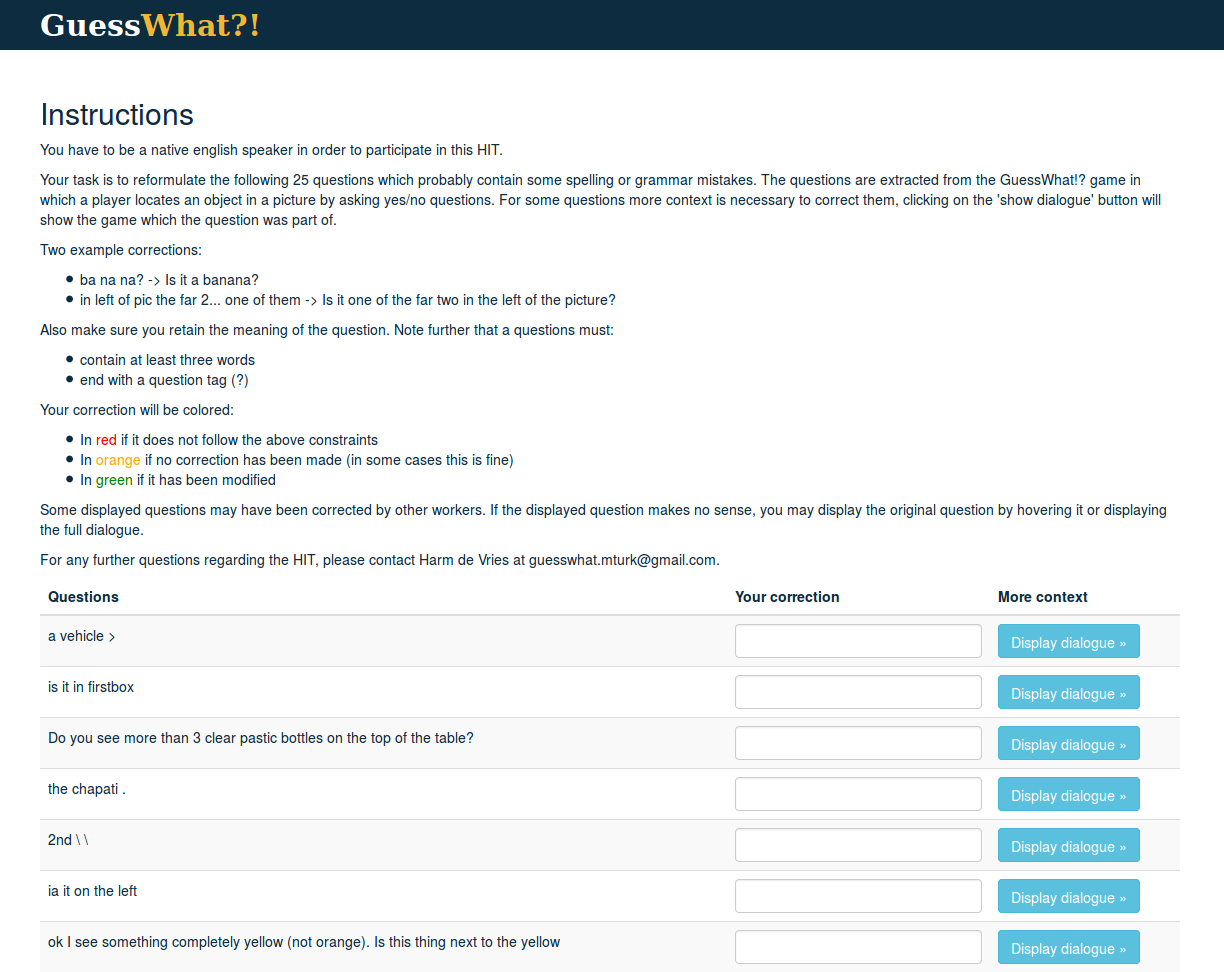}
\caption{Interface to fix ill-formatted questions}
\end{subfigure}
\begin{subfigure}{0.75\linewidth}
\includegraphics[width=\linewidth]{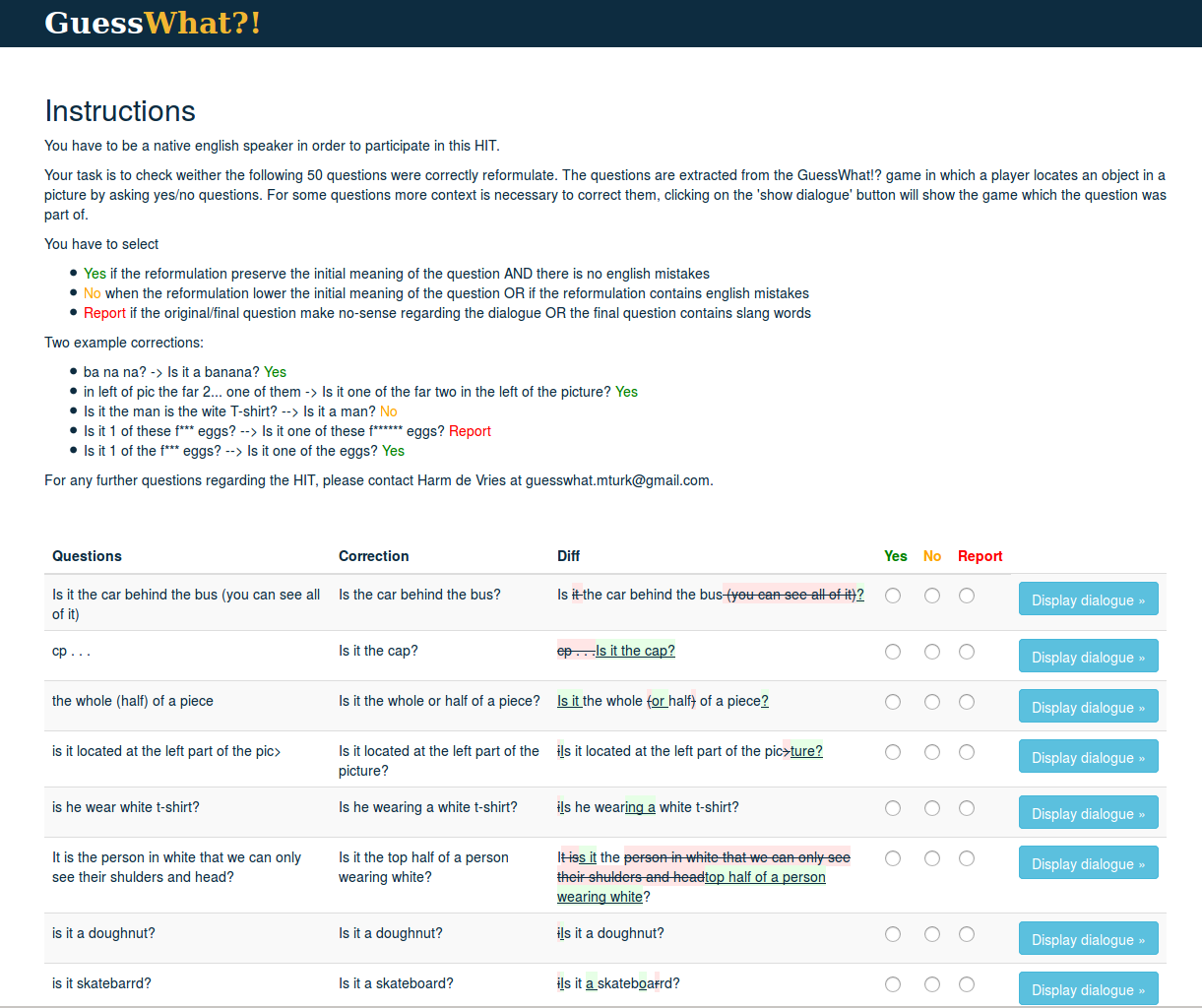}
\caption{Interface to validate the fix  ill-formatted questions}
\end{subfigure}
\caption{In the first task, we ask workers to correct mistakes in the questions. We then ask workers to validate the proposed correction by showing the difference between the original question and its correction. We alternate both tasks till all questions are corrected and validated.}
\label{fig:mistake_fixer}
\end{figure}

\clearpage

\section{\GW samples}
\label{ap:gw_samples}

\begin{figure}[h!]
\centering
\includegraphics[width=0.90\linewidth]{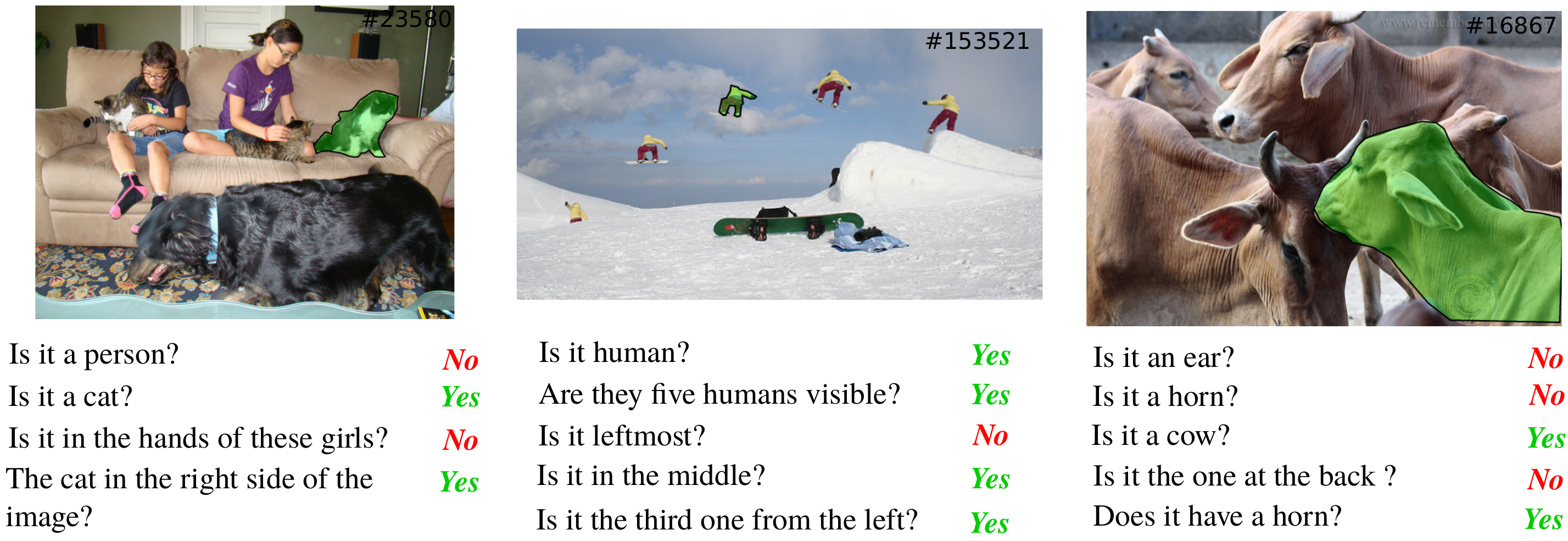}
  \caption{Three other examples of our dataset.}
\label{fig:example2}
\end{figure}

\begin{figure}[h!]
\centering
\includegraphics[width=0.90\linewidth]{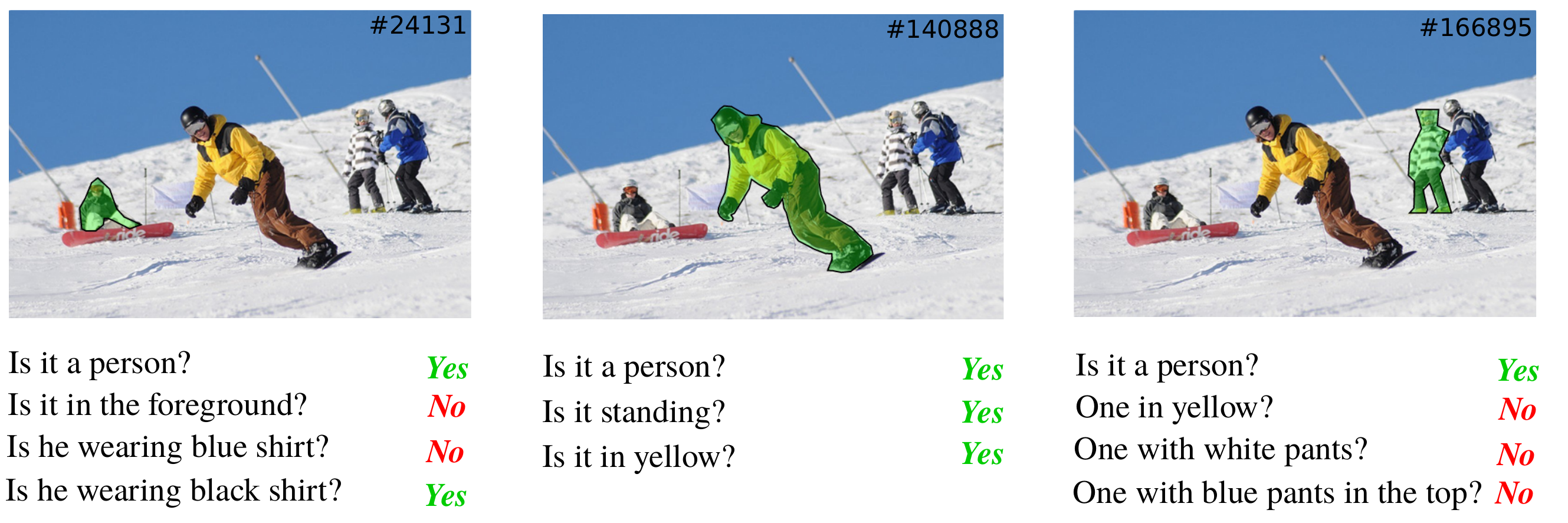}
  \caption{Same picture but different objects.}
\label{fig:same_picture}
\end{figure}

\begin{figure}[h!]
\centering
\includegraphics[width=0.90\linewidth]{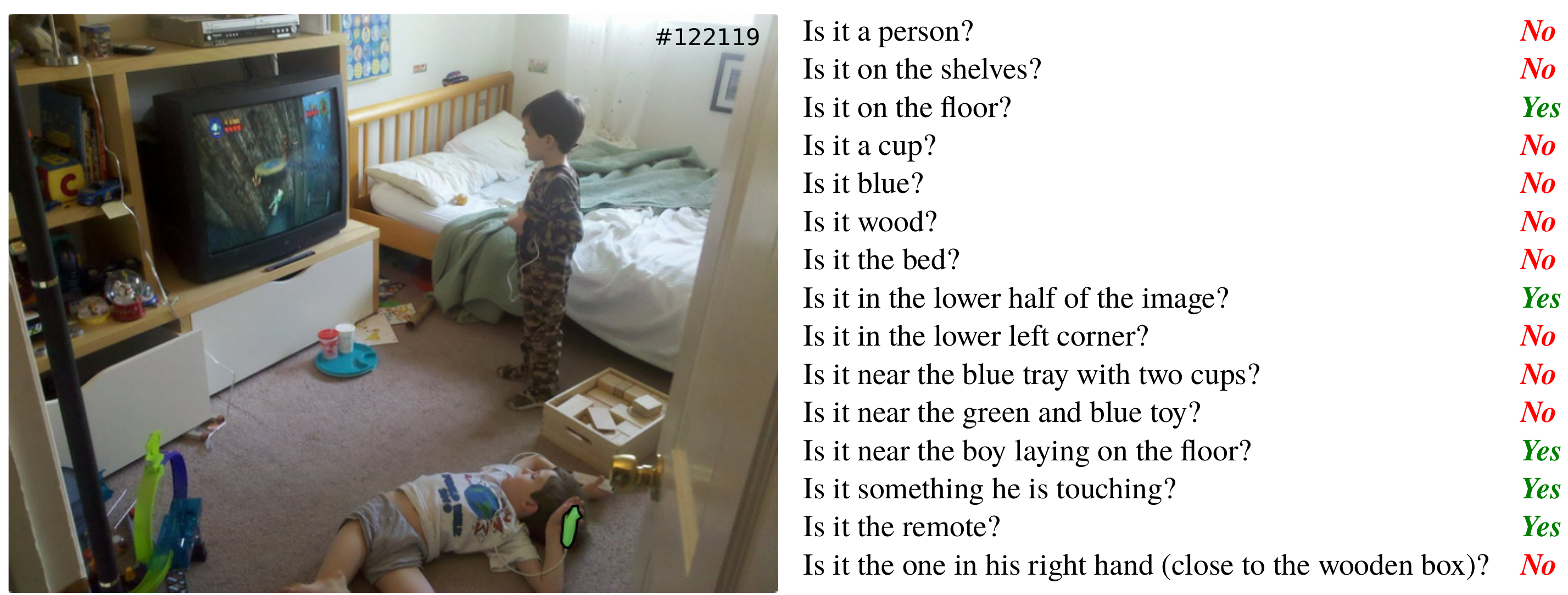}
  \caption{A long dialogue example in a very rich environment.}
\label{fig:long_example}
\end{figure}

\begin{figure}[h!]
\centering
\includegraphics[width=0.90\linewidth]{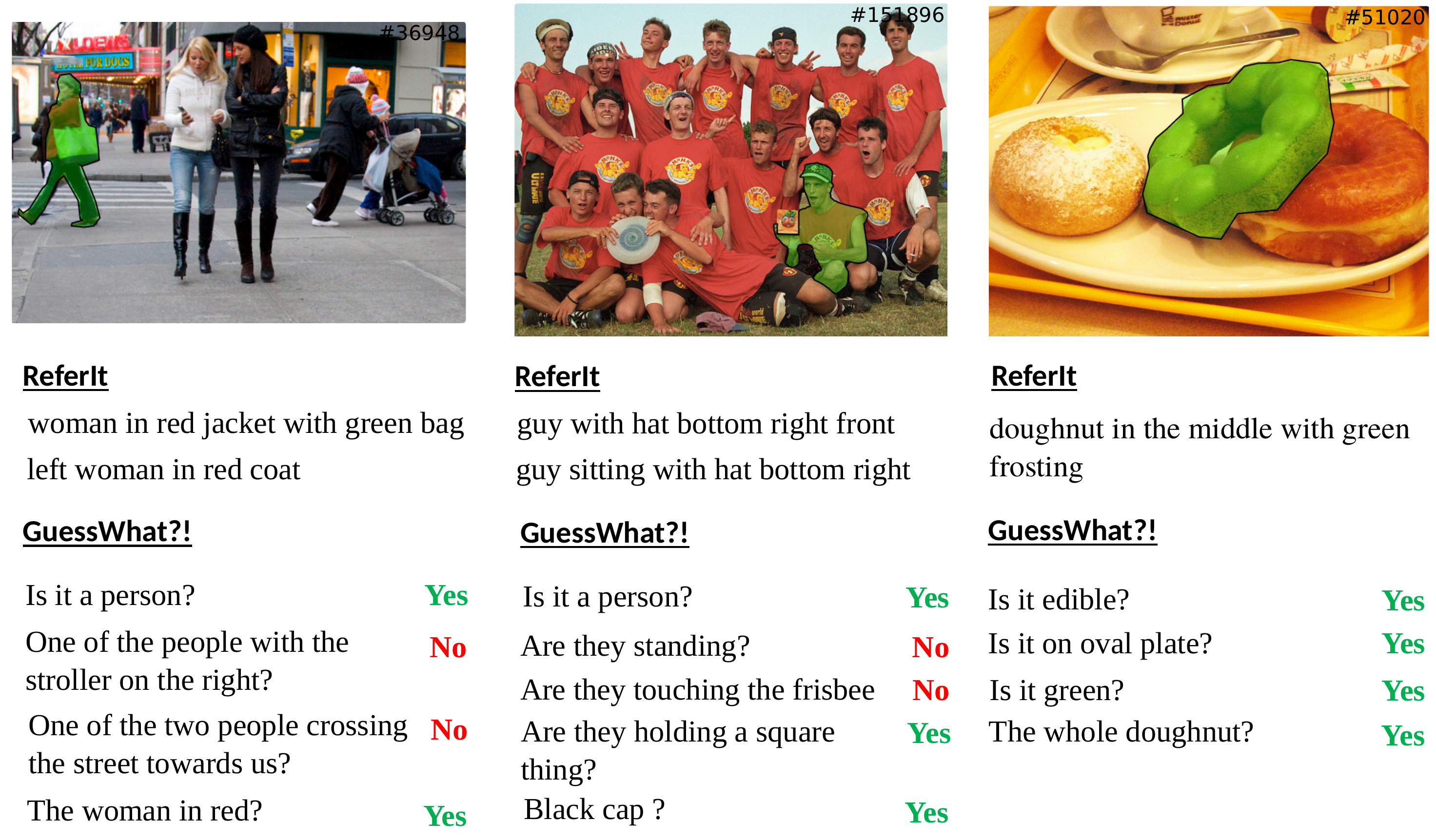}
  \caption{Samples illustrating the difference between \GW and \Referit games. As both dataset are constructed on top of MS COCO, we picked identical objects (and images).}
\label{fig:referit}
\end{figure}

\clearpage

\section{Additional database statistics}
\label{ap:statistics}
\begin{figure*}[h!]
    \centering
        \includegraphics[ width=\linewidth]{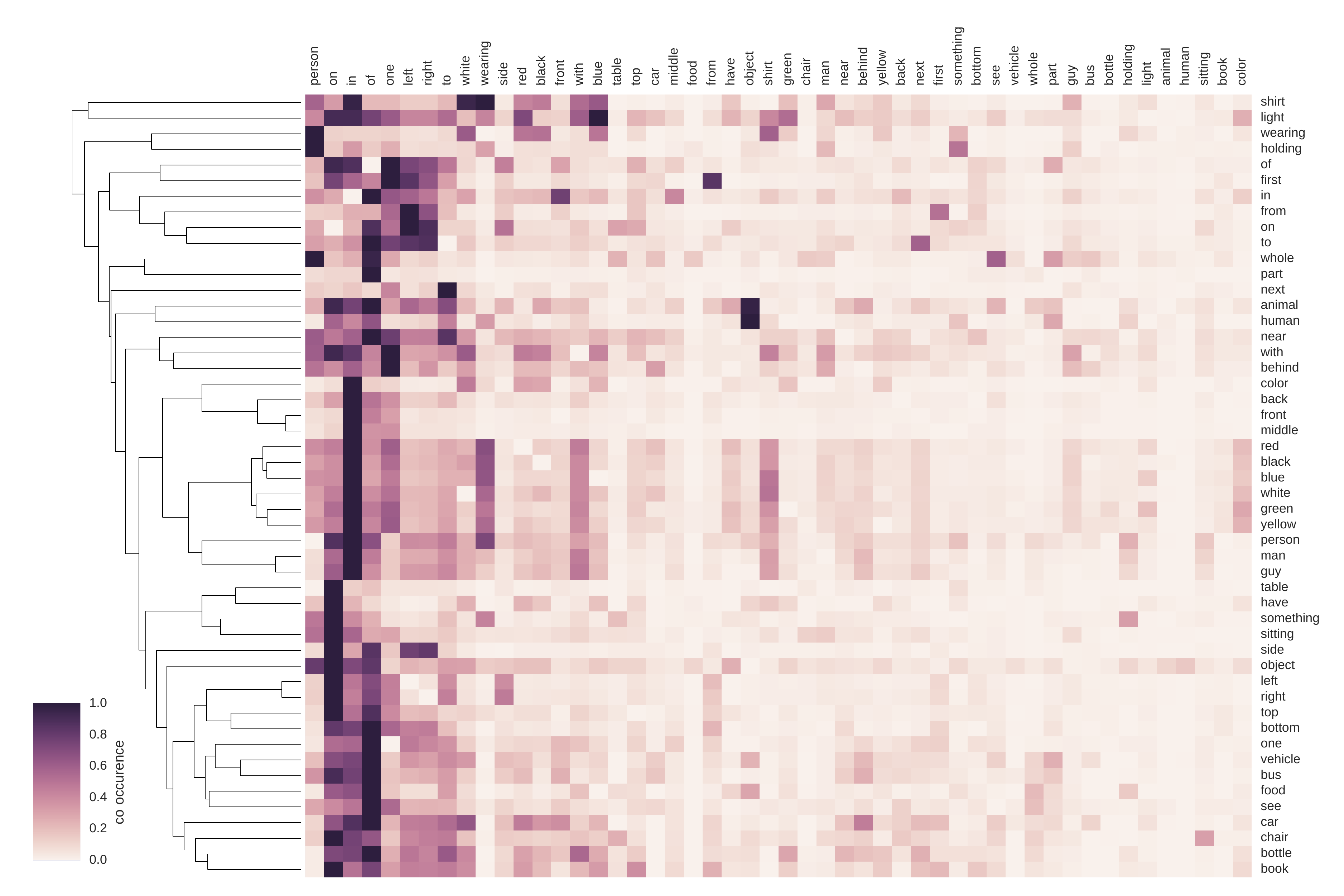}
    \caption{Co-occurrence matrix of words. Only the 50 most frequent words are kept. Rows are first normalized before being sorted thanks a hierarchical clustering with an euclidean distance.}
    \label{fig:word_cooccurence}
\end{figure*}

\begin{figure}
\centering
\includegraphics[width=\linewidth]{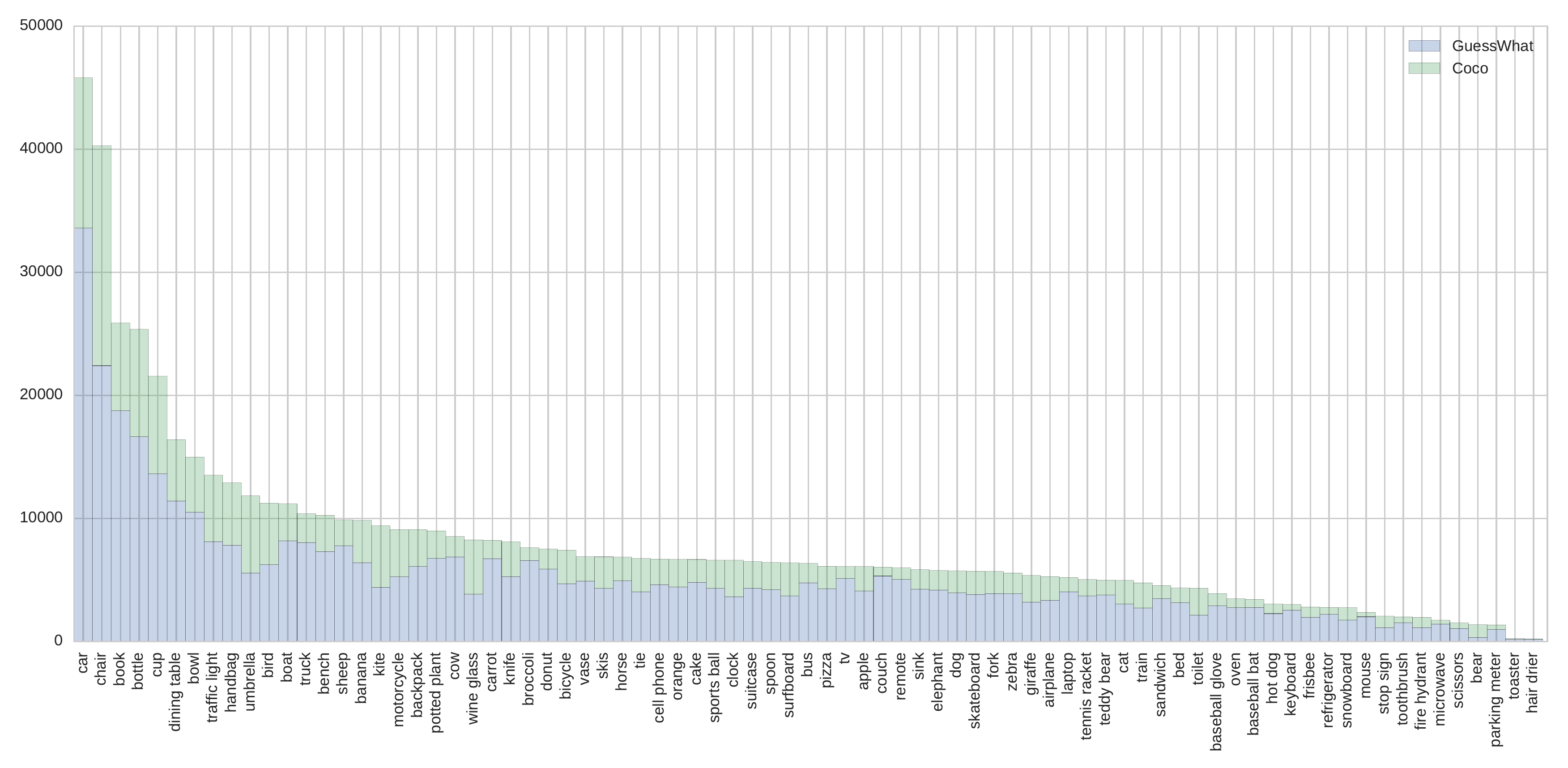}
\caption{Visualization of the object category distribution of MS COCO and \GW dataset. The person category was removed for clarity (resp. $273469$ and $188204$).}
\label{fig:coco_cat}
\end{figure}

\begin{figure}[t!]
\centering
\begin{subfigure}{0.45\linewidth}
\includegraphics[width=\linewidth]{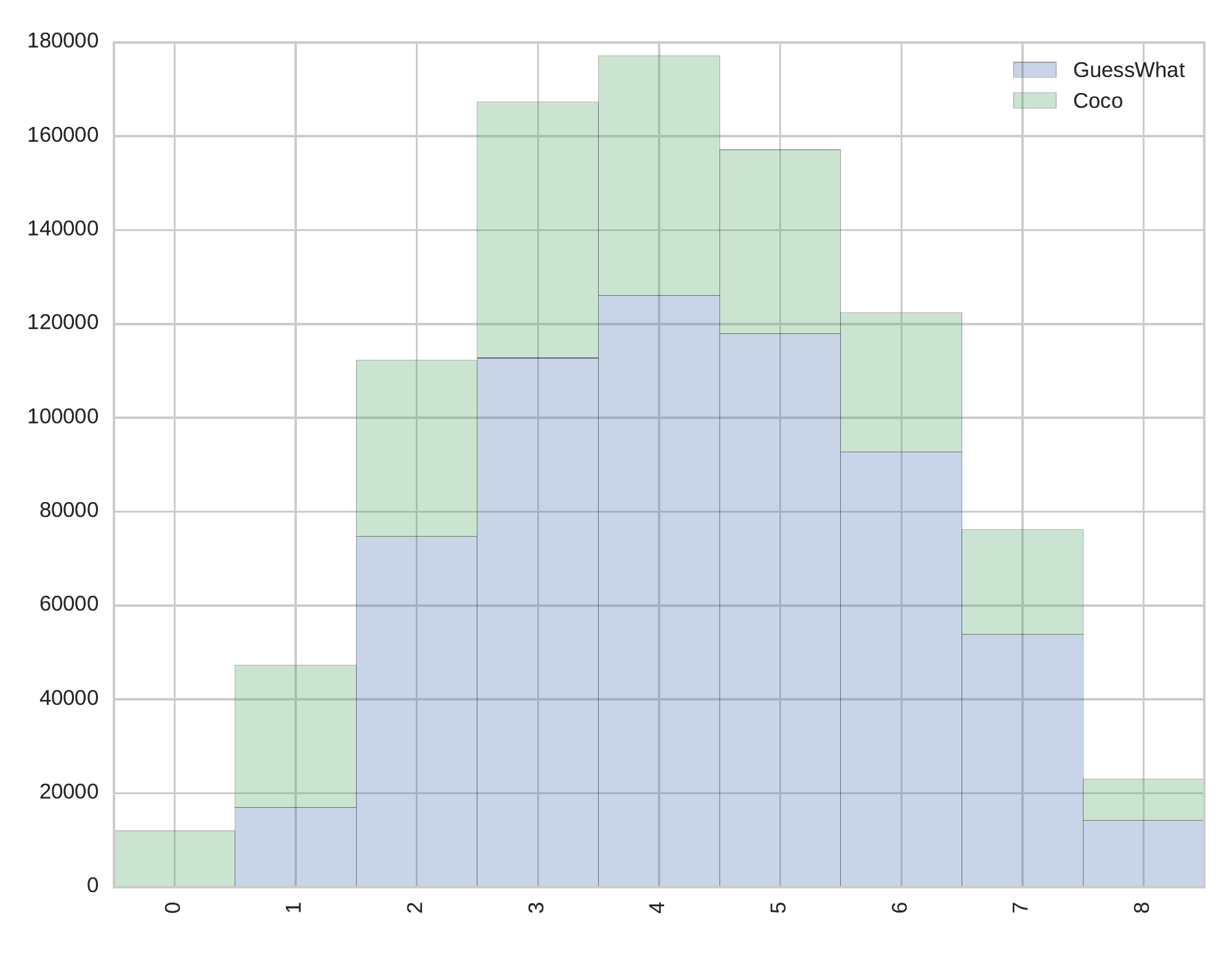}
\caption{}
\label{fig:coco_area}
\end{subfigure}
\begin{subfigure}{0.45\linewidth}
\includegraphics[width=\linewidth]{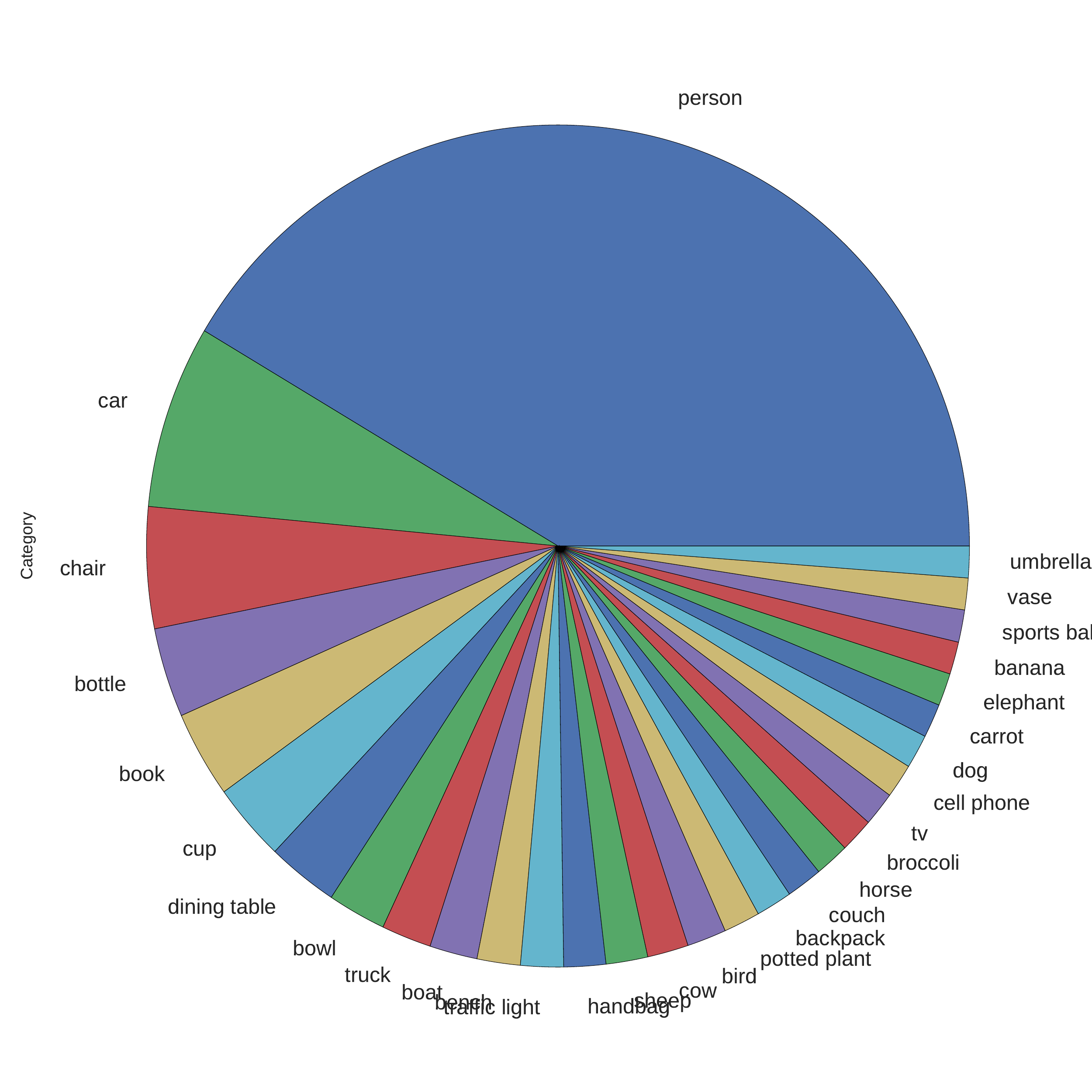}
\caption{}
\label{fig:top30}
\end{subfigure}
\caption{(a) Visualization of the object size distribution of MS COCO and \GW dataset. 
(b)  Distribution of the the 30 (out of 80) prominent object categories in the \GW which represent 71.3\% of the objects.
}
\vskip -1em
\end{figure}

Figure~\ref{fig:word_cooccurence} presents a word co-occurrence matrix of the \GW dataset. Figure~\ref{fig:coco_cat} and Figure~\ref{fig:coco_area} compares the object size and category distribution of \GW with MS Coco.

\newpage

\begin{figure}[t]
\centering
\begin{subfigure}{0.45\linewidth}
\includegraphics[width=\linewidth]{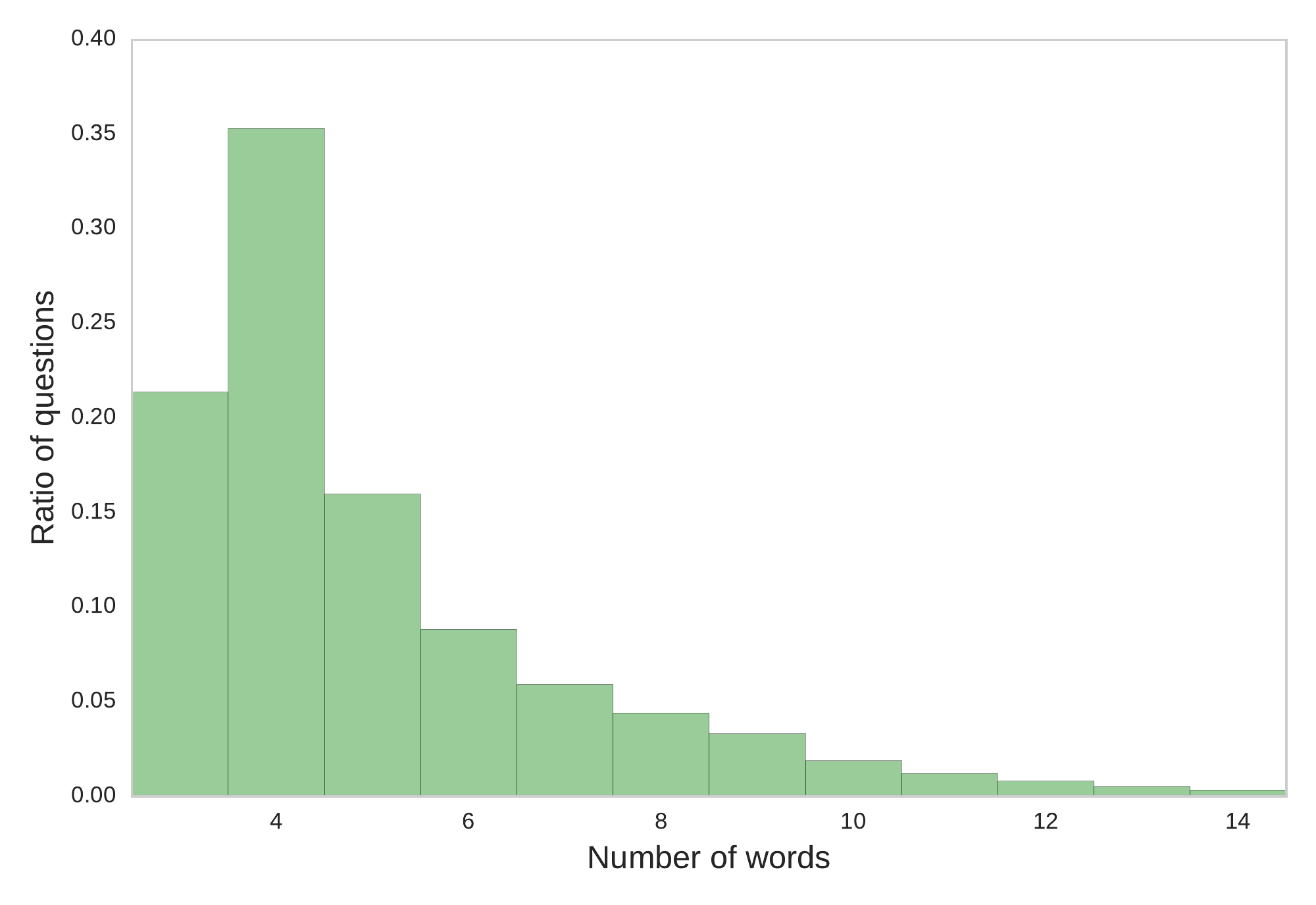}
\caption{}
\label{fig:new_words}
\end{subfigure}
\begin{subfigure}{0.45\linewidth}
\includegraphics[width=\linewidth]{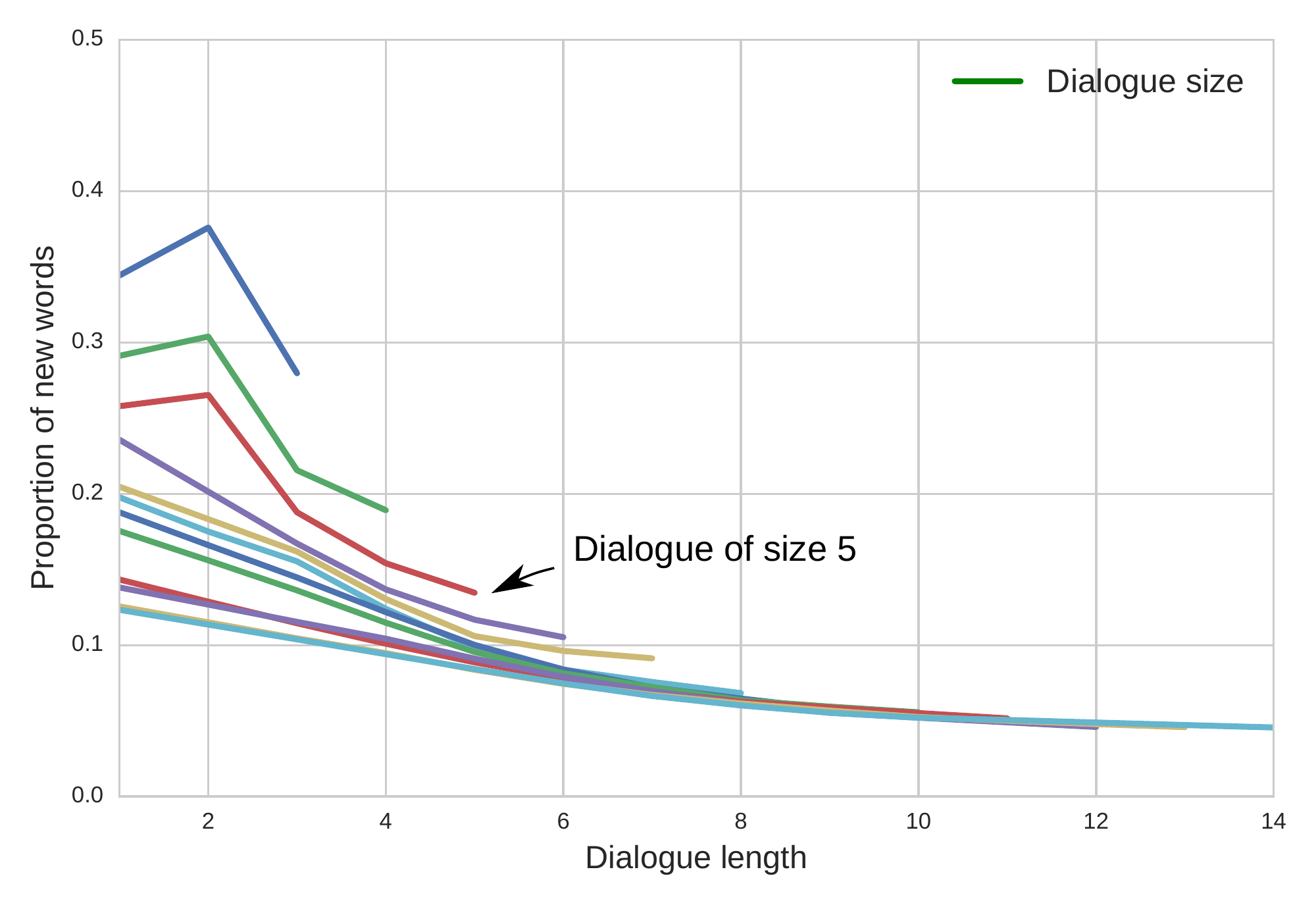}
\caption{}
\label{fig:w_q}
\end{subfigure}
\caption{(a) Number of words per question. The question length follows a Poisson-like distribution, a finding which is in line with other datasets~\cite{antol2015vqa}. (b) Percentage of apparition of new words along a dialogues. Questioner tends keep using the same words during the dialogues.}
\vskip -1em
\end{figure}

\begin{figure}[t]
  \begin{minipage}{\textwidth}
  \begin{minipage}{0.45\textwidth}
    \centering
    \begin{tabular}{|c|c|}
    \hline
    Topic 1 & Topic 2\\
    \hline
    Abstract words & Descriptive words\\
    \hline
    person & left \\
    food & one \\
    vehicle  & right \\
    human & wearing \\
    car & side\\
    one & white\\
    object & red\\
    animal & table\\
    \hline
    \end{tabular}
  \end{minipage}
  \hfill
  \begin{minipage}{0.45\textwidth}
    \centering
    \includegraphics[width=1\linewidth]{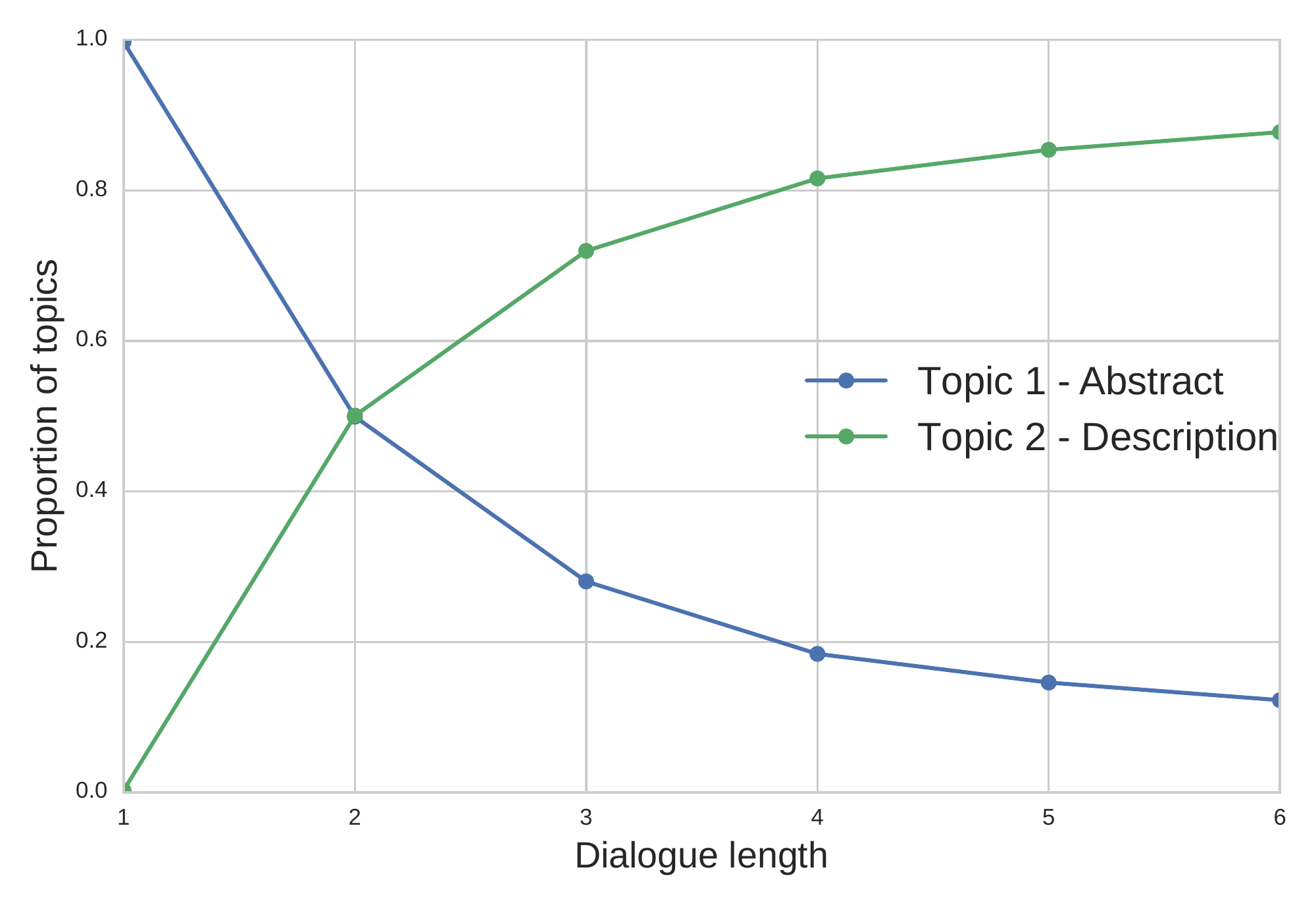}
    \end{minipage}
  \end{minipage}
\caption{Relative evolution of topics during a dialogue of size 6. We applied Data Topic Models (DTM) \cite{blei2006dynamic} with the python framework \cite{rehurek_lrec} on our dataset. The table reports the two prominent detected topics with their respective key words while the figure display their relative evolution during the dialogue. The topic titles are manually picked.}
\label{fig:DTM}
\end{figure}

\clearpage

\begin{figure}
\centering
\includegraphics[width=1\linewidth]{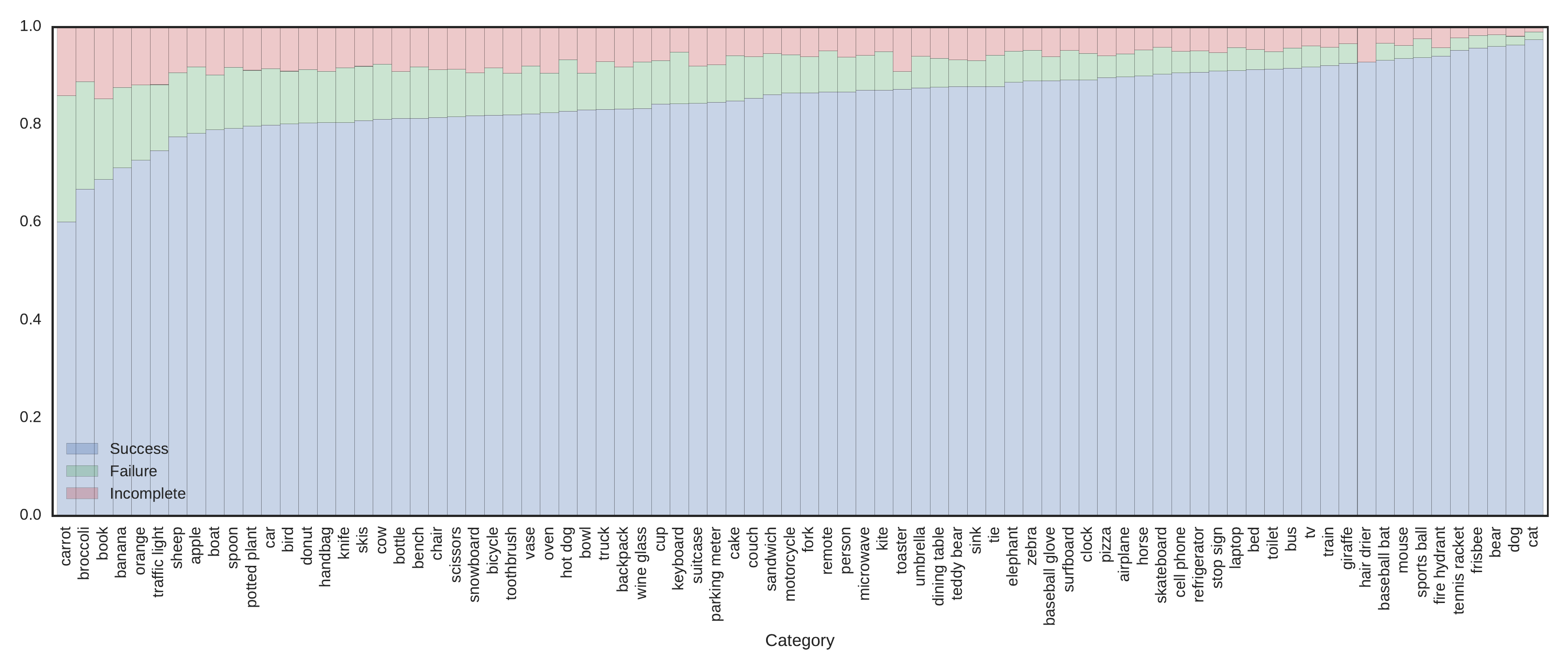}
\caption{Histogram of success ratio broken down per object category.}
\label{fig:success_category}
\end{figure}

\begin{figure}[t]
\centering
\begin{subfigure}{0.45\linewidth}
\includegraphics[width=\linewidth]{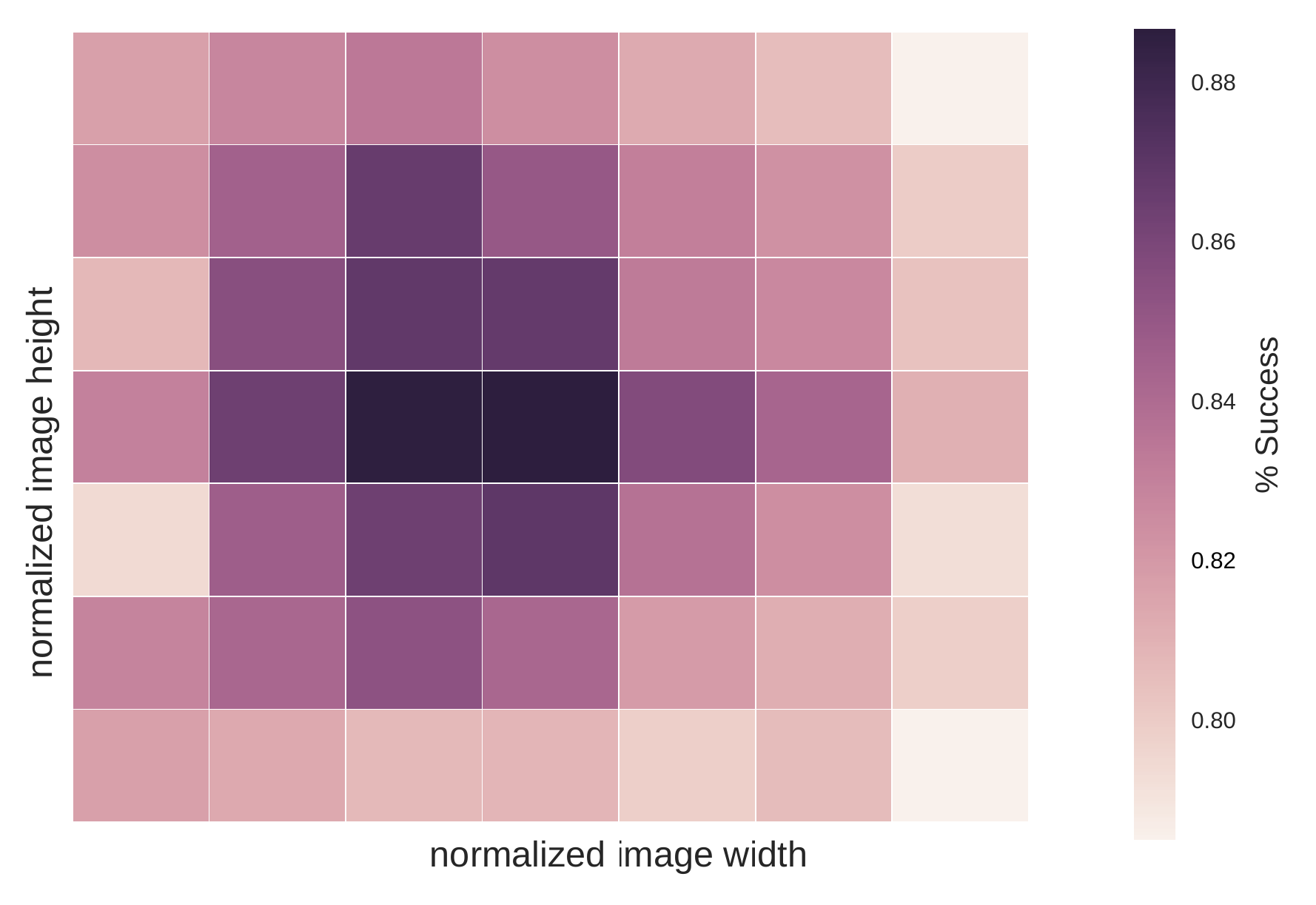}
\caption{}
\label{fig:success_spatial}
\end{subfigure}
\begin{subfigure}{0.45\linewidth}
\includegraphics[width=\linewidth]{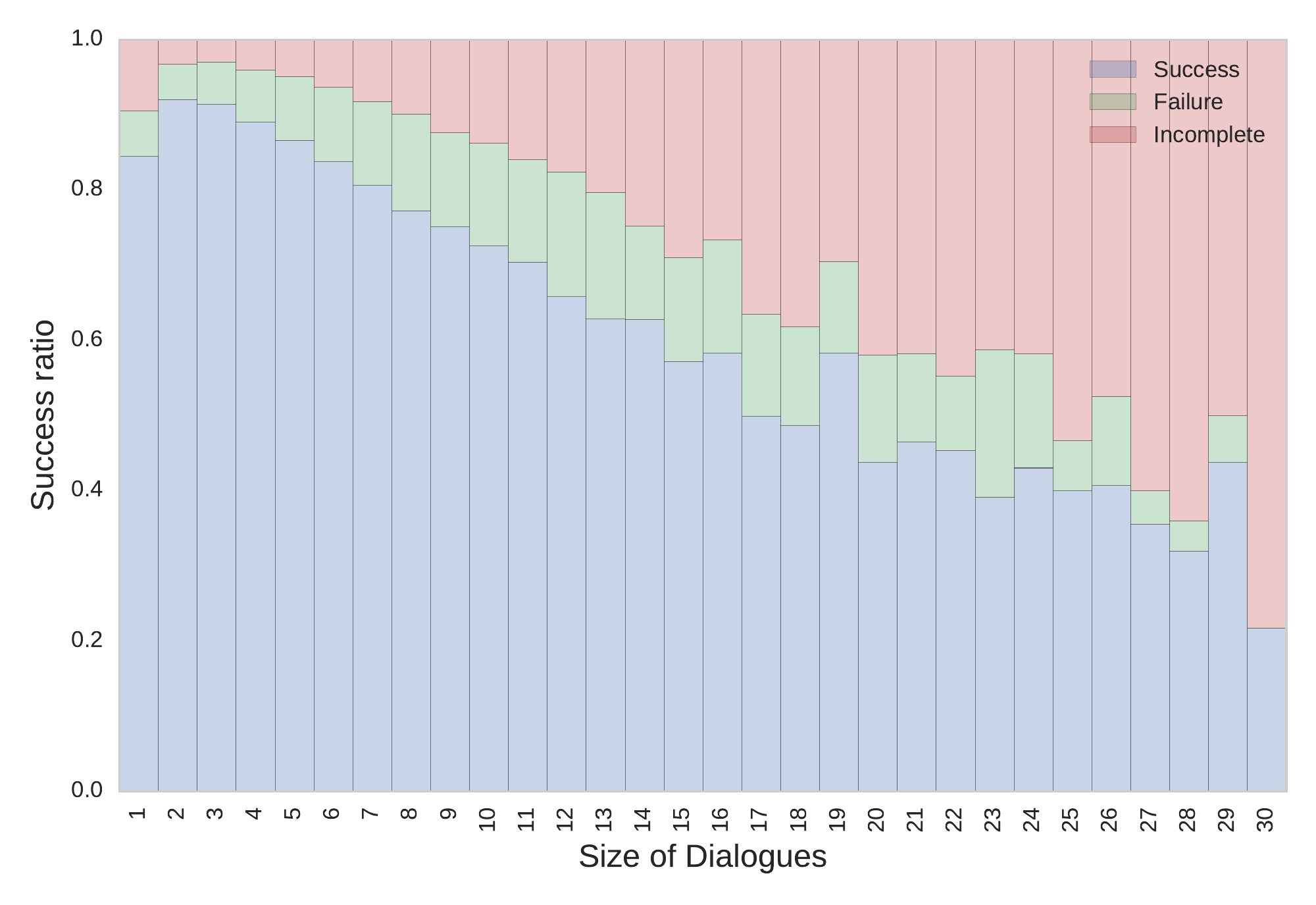}
\caption{}
\label{fig:success_length}
\end{subfigure}
\centering
\caption{(a) Heatmap of the success ratio with respect to the spatial location within the picture. (b) Histogram of the success ratio relative to the dialogue length.}
\end{figure}


\begin{landscape}
\begin{table}
\begin{subtable}[t]{\linewidth}
\small

\begin{tabular}{|l c|l c c |l c c |}
\hline
person & 14.48 &person & 3.20 &=& left & 2.95 &\textbf{\color{green}$\nearrow$}\\ 
food & 1.29 &left & 1.69 & & right & 2.32 &\textbf{\color{green}$\nearrow$}\\ 
animal & 1.16 &wearing & 1.20 &new& person & 2.28 &\textbf{\color{red}$\searrow$}\\ 
human & 1.03 &right & 1.02 &new& wearing & 1.66 &\textbf{\color{red}$\searrow$}\\ 
object & 0.77 &front & 0.97 &new& whole & 1.58 &new\\ 
car & 0.60 &white & 0.91 &new& white & 1.56 &=\\ 
vehicle & 0.57 &red & 0.77 &new& red & 1.26 &=\\ 
cat & 0.41 &car & 0.64 &\textbf{\color{red}$\searrow$}& black & 1.19 &\textbf{\color{green}$\nearrow$}\\ 
alive & 0.37 &black & 0.60 &new& front & 1.14 &\textbf{\color{red}$\searrow$}\\ 
dog & 0.35 &blue & 0.59 &new& blue & 1.10 &=\\ 
\hline
\end{tabular}\caption{Dialogues having 3 questions}
\end{subtable}

\begin{subtable}[t]{\linewidth}
\small
\begin{tabular}{|l c|l c c |l c c |l c c |l c c |}
\hline
person & 8.20 &person & 1.98 &=& left & 1.69 &\textbf{\color{green}$\nearrow$}& left & 1.92 &=& left & 2.13 &=\\ 
food & 1.03 &left & 1.03 &\textbf{\color{green}$\nearrow$}& person & 1.41 &\textbf{\color{red}$\searrow$}& right & 1.77 &\textbf{\color{green}$\nearrow$}& right & 2.04 &=\\ 
human & 0.56 &right & 0.66 &new& right & 1.26 &=& person & 1.20 &\textbf{\color{red}$\searrow$}& white & 1.28 &\textbf{\color{green}$\nearrow$}\\ 
animal & 0.46 &front & 0.59 &new& white & 0.84 &\textbf{\color{green}$\nearrow$}& white & 1.12 &=& person & 1.26 &\textbf{\color{red}$\searrow$}\\ 
vehicle & 0.45 &car & 0.51 &\textbf{\color{green}$\nearrow$}& wearing & 0.82 &\textbf{\color{green}$\nearrow$}& wearing & 0.93 &=& black & 0.90 &\textbf{\color{green}$\nearrow$}\\ 
object & 0.42 &white & 0.48 &new& side & 0.67 &\textbf{\color{green}$\nearrow$}& black & 0.79 &\textbf{\color{green}$\nearrow$}& wearing & 0.85 &\textbf{\color{red}$\searrow$}\\ 
car & 0.36 &wearing & 0.48 &new& red & 0.62 &\textbf{\color{green}$\nearrow$}& red & 0.72 &=& red & 0.80 &=\\ 
furniture & 0.24 &side & 0.43 &new& front & 0.58 &\textbf{\color{red}$\searrow$}& side & 0.69 &\textbf{\color{red}$\searrow$}& whole & 0.80 &new\\ 
left & 0.24 &red & 0.39 &new& black & 0.55 &new& blue & 0.65 &\textbf{\color{green}$\nearrow$}& blue & 0.75 &=\\ 
edible & 0.20 &vehicle & 0.39 &\textbf{\color{red}$\searrow$}& blue & 0.54 &new& front & 0.58 &\textbf{\color{red}$\searrow$}& front & 0.73 &=\\ 
\hline
\end{tabular}\caption{Dialogues having 5 questions}
\end{subtable}

\begin{subtable}[t]{\linewidth}
\small
\begin{tabular}{|l c|l c c |l c c |l c c |l c c |l c c |l c c |}
\hline
person & 5.89 &person & 1.44 &=& left & 1.08 &\textbf{\color{green}$\nearrow$}& left & 1.26 &=& left & 1.33 &=& left & 1.42 &=& left & 1.65 &=\\ 
food & 0.74 &left & 0.73 &\textbf{\color{green}$\nearrow$}& person & 0.96 &\textbf{\color{red}$\searrow$}& right & 1.08 &\textbf{\color{green}$\nearrow$}& right & 1.22 &=& right & 1.39 &=& right & 1.54 &=\\ 
human & 0.38 &right & 0.42 &new& right & 0.89 &=& person & 0.82 &\textbf{\color{red}$\searrow$}& white & 0.81 &\textbf{\color{green}$\nearrow$}& white & 0.88 &=& white & 0.96 &=\\ 
vehicle & 0.30 &table & 0.37 &\textbf{\color{green}$\nearrow$}& side & 0.57 &\textbf{\color{green}$\nearrow$}& white & 0.67 &\textbf{\color{green}$\nearrow$}& person & 0.80 &\textbf{\color{red}$\searrow$}& person & 0.84 &=& person & 0.90 &=\\ 
object & 0.28 &front & 0.36 &new& white & 0.50 &\textbf{\color{green}$\nearrow$}& side & 0.60 &\textbf{\color{red}$\searrow$}& wearing & 0.59 &\textbf{\color{green}$\nearrow$}& black & 0.63 &\textbf{\color{green}$\nearrow$}& red & 0.65 &\textbf{\color{green}$\nearrow$}\\ 
car & 0.26 &food & 0.35 &\textbf{\color{red}$\searrow$}& wearing & 0.48 &\textbf{\color{green}$\nearrow$}& wearing & 0.54 &=& side & 0.57 &\textbf{\color{red}$\searrow$}& red & 0.57 &\textbf{\color{green}$\nearrow$}& black & 0.63 &\textbf{\color{red}$\searrow$}\\ 
animal & 0.26 &side & 0.35 &new& red & 0.41 &new& red & 0.49 &=& red & 0.54 &=& wearing & 0.56 &\textbf{\color{red}$\searrow$}& blue & 0.57 &\textbf{\color{green}$\nearrow$}\\ 
furniture & 0.20 &car & 0.31 &\textbf{\color{red}$\searrow$}& table & 0.39 &\textbf{\color{red}$\searrow$}& table & 0.41 &=& black & 0.51 &\textbf{\color{green}$\nearrow$}& blue & 0.54 &\textbf{\color{green}$\nearrow$}& wearing & 0.52 &\textbf{\color{red}$\searrow$}\\ 
left & 0.14 &wearing & 0.28 &new& front & 0.38 &\textbf{\color{red}$\searrow$}& black & 0.41 &\textbf{\color{green}$\nearrow$}& blue & 0.49 &\textbf{\color{green}$\nearrow$}& side & 0.53 &\textbf{\color{red}$\searrow$}& next & 0.51 &new\\ 
boat & 0.14 &something & 0.28 &new& car & 0.37 &\textbf{\color{red}$\searrow$}& blue & 0.37 &new& front & 0.42 &\textbf{\color{green}$\nearrow$}& front & 0.45 &=& side & 0.51 &\textbf{\color{red}$\searrow$}\\ 
\hline
\end{tabular}\caption{Dialogues having 7 questions}
\end{subtable}
\caption{Proportions of the ten most common words for each depth of questions for sorted by the size of the dialogues}
\label{tab:word_occ_dialogues}
\end{table}

\end{landscape}

\clearpage
\section{All oracle baselines}
\label{ap:oracle_baselines}
\vspace{-0.5em}
\begin{table}[h!]
\begin{center}
\begin{tabular}{|l|l|l|l|}
\hline
Model & Train err & Valid err & Test err\\
\hline
Dominant class (no) & 47.4\% & 46.2\% & 50.9\%\\
Category & 43.0\% & 42.8\% & 43.1\%\\
Question & 40.2\% & 41.7\% & 41.2\%\\
Crop & 40.9\% & 42.7\% & 43.0\%\\
Image & 45.7\% & 46.7\% & 46.7\%\\
Spatial & 43.9\% & 44.1\% & 44.3\%\\
\hline
Category + Spatial & 41.6\% & 41.7\% & 42.1\%\\
Question + Crop & 22.3\% & 29.1\% & 29.2\%\\
Question + Image & 37.9\% & 40.2\% & 39.8\%\\
Question + Category & 23.1\% & 25.8\% & 25.7\%\\
Question + Spatial & 28.0\% & 31.2\% & 31.3\%\\
Spatial + Crop & 41.8\% & 42.4\% & 42.8\%\\
Crop + Image & 41.6\% & 42.1\% & 42.4\%\\
Spatial + Image & 42.2\% & 44.1\% & 44.2\%\\
Category + Crop & 41.0\% & 41.7\% & 42.3\%\\
Category + Image & 42.3\% & 42.7\% & 43.0\%\\
\hline
Category + Crop + Image & 40.6\% & 41.5\% & 41.8\%\\
Category + Spatial + Crop & 40.6\% & 41.6\% & 42.1\%\\
Question + Category + Spatial & 17.2\% & 21.1\% & \textbf{21.5}\%\\
Question + Crop + Image & 23.7\% & 29.9\% & 30.0\%\\
Category + Spatial + Image & 40.4\% & 42.0\% & 42.2\%\\
Question + Category + Image & 23.4\% & 27.1\% & 27.4\%\\
Question + Spatial + Image & 28.4\% & 32.5\% & 32.5\%\\
Spatial + Crop + Image & 41.6\% & 42.1\% & 42.5\%\\
Question + Category + Crop & 20.4\% & 24.4\% & 24.7\%\\
Question + Spatial + Crop & 19.4\% & 26.0\% & 26.2\%\\
\hline
Question + Category + Spatial + Crop & 16.1\% & 21.7\% & 22.1\%\\
Question + Spatial + Crop + Image & 20.7\% & 27.7\% & 27.9\%\\
Category + Spatial + Crop + Image & 40.3\% & 41.4\% & 41.8\%\\
Question + Category + Spatial + Image & 19.2\% & 23.2\% & 23.5\%\\
Question + Category + Crop + Image & 20.0\% & 25.3\% & 25.5\%\\
\hline
Question + Category + Spatial + Crop + Image & 17.8\% & 23.2\% & 23.3\%\\
\hline
\end{tabular}
\caption{Classification errors for all oracle baselines.}
\label{tab:all_oracle_baseline}
\end{center}
\vskip -1em
\end{table}

\clearpage
\section{Guesser generation model}
\label{ap:guesser}

\begin{figure*}[h]
    \centering
    \includegraphics[width=0.60\linewidth]{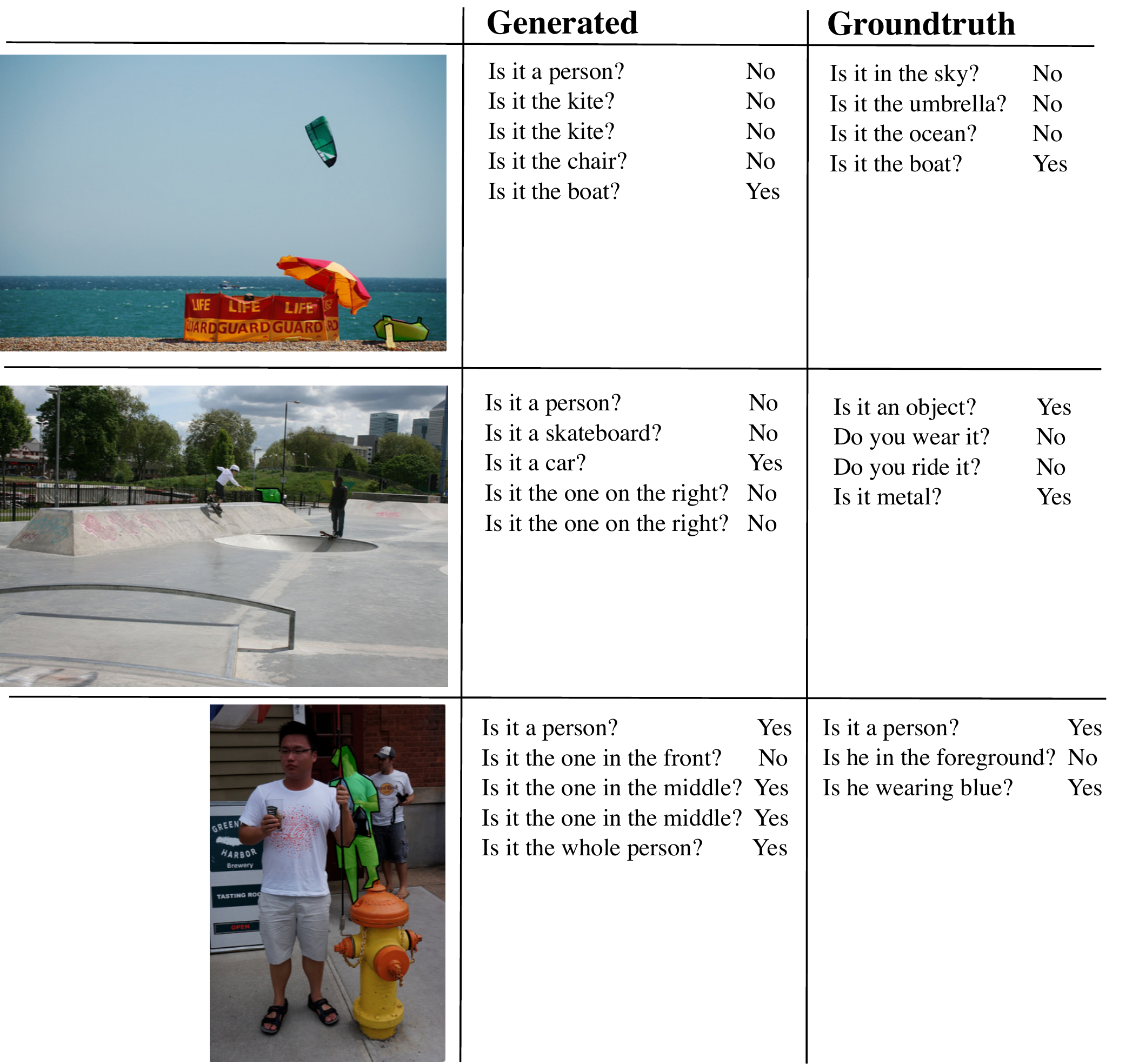}
    \caption{Three samples of QGen+GT model for which the correct object was predicted. }
    \label{fig:correct_samples}
\end{figure*}

\begin{figure*}[h]
    \centering
    \includegraphics[width=0.60\linewidth]{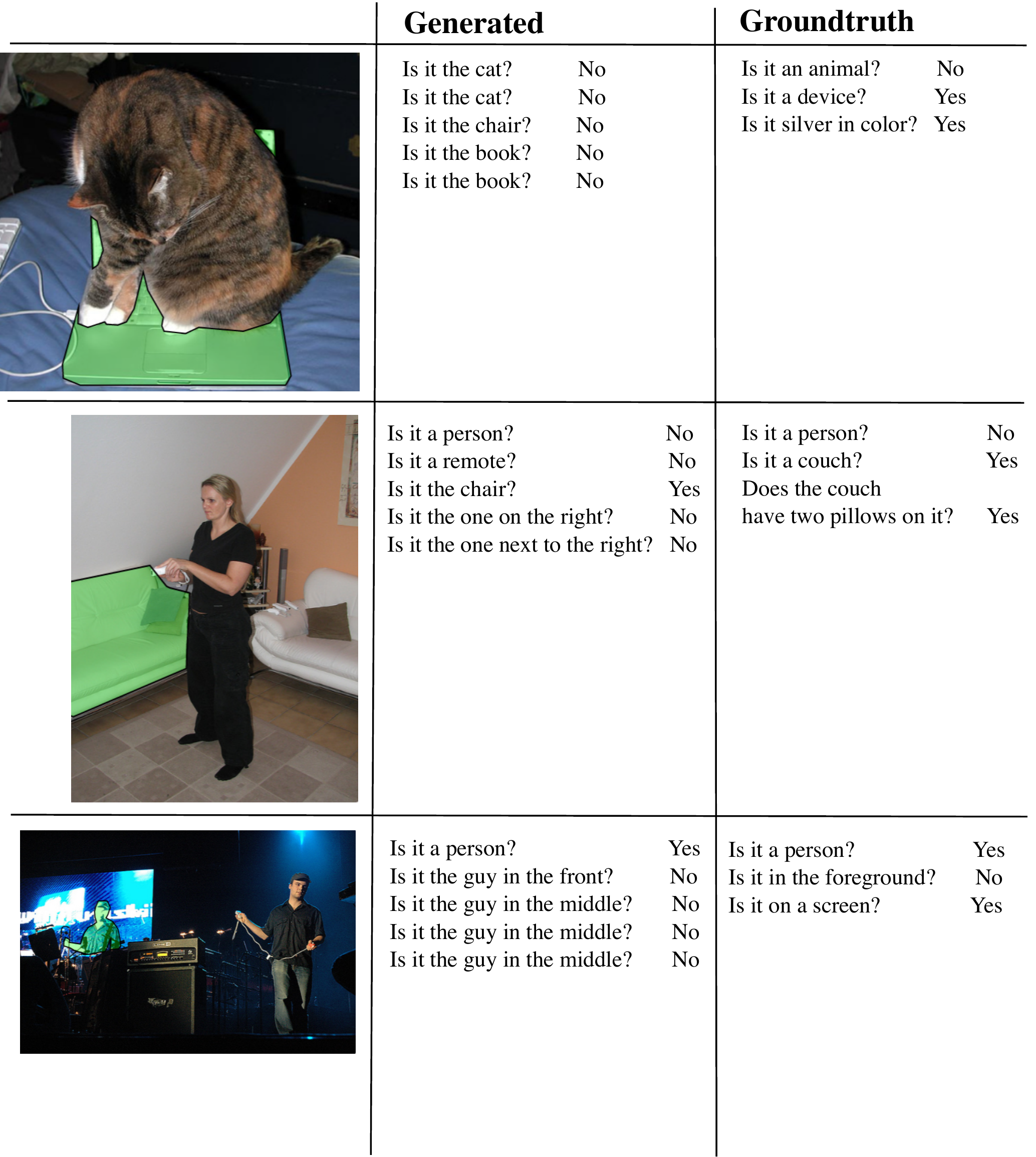}
    \caption{Three dialogue samples of QGen+GT model for which the wrong object was predicted.}
    \label{fig:incorrect_samples}
\end{figure*}



\end{document}